\documentclass{article}

\PassOptionsToPackage{numbers, compress}{natbib}
\usepackage[final]{neurips_2019}

\usepackage[utf8]{inputenc} % allow utf-8 input
\usepackage[T1]{fontenc}    % use 8-bit T1 fonts
\usepackage{hyperref}       % hyperlinks
\usepackage{url}            % simple URL typesetting
\usepackage{booktabs}       % professional-quality tables
\usepackage{amsfonts}       % blackboard math symbols
\usepackage{nicefrac}       % compact symbols for 1/2, etc.
\usepackage{microtype}      % microtypography

\usepackage{floatrow}
\newfloatcommand{capbtabbox}{table}[][\FBwidth]

\usepackage{blindtext}
\usepackage{wrapfig}

\usepackage[small,bf]{caption}
\usepackage{mathtools}
\usepackage{color}
\usepackage{lipsum}
\usepackage[textsize=scriptsize]{todonotes}

\definecolor{Sijia_color}{rgb}{0.858, 0.188, 0.478}
\definecolor{XC_color}{rgb}{0.1, 0.488, 0.478}

\definecolor{XL_color}{rgb}{0.0, 0.1, 0.9}
\usepackage{booktabs}       % professional-quality tables
\usepackage{amsfonts}       % blackboard math symbols
\usepackage{nicefrac}       % compact symbols for 1/2, etc.
\usepackage{microtype}      % microtypography
\usepackage{makecell}
\usepackage{adjustbox}
\usepackage[para,online,flushleft]{threeparttable}

\usepackage{bbding}
\usepackage{pifont}
\usepackage{wasysym}
\usepackage{amssymb}

\usepackage{multirow}
\usepackage{grffile}
\usepackage{url}

\usepackage{algorithm}
\usepackage{algorithmic}
\usepackage{setspace}
\AtBeginDocument{%
  \addtolength\abovedisplayskip{-0.4\baselineskip}%
  \addtolength\belowdisplayskip{-0.4\baselineskip}%
  \addtolength\abovedisplayshortskip{-0.4\baselineskip}%
  \addtolength\belowdisplayshortskip{-0.4\baselineskip}%
}

\newtheorem{myprop}{\bf{Proposition}}

\newtheorem{mythr}{\bf{Theorem}}
\newtheorem{mylemma}{\bf{Lemma}}

\newtheorem{lemma}{Lemma}[section]

\DeclareMathOperator{\diag}{diag}

\DeclareMathOperator*{\minimize}{\text{minimize}}

\newcommand{\argmin}{\operatornamewithlimits{arg\,min}}

\DeclarePairedDelimiterX{\inp}[2]{\langle}{\rangle}{#1, #2}

\DeclareMathOperator*{\st}{\text{subject to}}
\DeclareMathAlphabet\mathbfcal{OMS}{cmsy}{b}{n}
\newcommand{\Def}[0]{\mathrel{\mathop:}=}

\def\env@cases{%
  \let\@ifnextchar\new@ifnextchar
  \left.
  \def\arraystretch{1.2}%
  \array{@{}l@{\,\,}l@{}}%
}%
\makeatother

\title{ZO-AdaMM: Zeroth-Order Adaptive Momentum Method for Black-Box Optimization}

\author{%
Xiangyi Chen$^{1,*}$~~~Sijia Liu$^{2,}$\thanks{Equal contribution. 
}~~~~Kaidi Xu$^{3,*}$~~~Xingguo Li$^{4,*}$\\
\textbf{Xue Lin}$^{3}$~~~\textbf{Mingyi Hong}$^{1}$~~~\textbf{David Cox}$^{2}$\\
    $^1$University of Minnesota, USA \\
      $^2$MIT-IBM Watson AI Lab, IBM Research,  USA\\
    $^3$Northeastern University, USA  \\
    $^4$Princeton University, USA %\\
}

\begin{document}

\maketitle

\begin{abstract}
The adaptive momentum method (AdaMM), which uses past gradients to update descent directions and learning rates simultaneously, has become one of the most popular first-order optimization methods for solving machine learning  problems. However,  AdaMM is not suited for solving black-box optimization problems, where explicit gradient forms are difficult or infeasible to obtain. In this paper, we propose a zeroth-order  AdaMM (ZO-AdaMM) algorithm, that generalizes AdaMM to the gradient-free regime. We show that the convergence rate of ZO-AdaMM for  both  convex and nonconvex optimization is roughly a factor of $O(\sqrt{d})$ worse than that of the first-order AdaMM algorithm, where $d$ is problem size. In particular, we provide a deep understanding on why  Mahalanobis distance matters in convergence of ZO-AdaMM and other AdaMM-type methods. As a byproduct, our analysis   makes the first step toward understanding adaptive learning rate methods for nonconvex constrained optimization. Furthermore, we demonstrate two applications, designing  per-image and universal adversarial attacks from black-box neural networks, respectively. We perform extensive experiments on ImageNet and empirically show that  ZO-AdaMM converges much faster to a solution of high accuracy compared with  $6$ state-of-the-art ZO optimization methods.  
\end{abstract}

\vspace{-0.05in}
\section{Introduction} 
%\textcolor{Sijia_color}{[SL: Have not fully modified Introduction, will further revised XL's version this weekend.]}
\label{sec: intro}
\vspace{-0.05in}

The development of gradient-free optimization methods has become increasingly important to solve many machine learning problems in which explicit expressions of the gradients are expensive or infeasible to obtain \cite{liu2017zeroth,sahu2018towards,feurer2015efficient,autoweka2,chen2017zoo,ilyas2018blackbox,tu2018autozoom}.
\textit{Zeroth-Order (ZO)} optimization methods, one type of gradient-free optimization methods, mimic first-order (FO)  methods  but  approximate  the full gradient (or stochastic gradient) through   random gradient estimates, given by the difference of function values at random query points  \cite{nesterov2015random,ghadimi2013stochastic}.  
  Compared to Bayesian optimization, derivative-free trust region methods, genetic algorithms and other types of gradient-free methods \cite{shahriari2016taking,conn2009global,whitley1994genetic,conn2009introduction},   ZO optimization  has two main advantages: a) ease of implementation, via slight modification of  commonly-used gradient-based algorithms, and b) comparable convergence rates   to first-order algorithms.
  
%%% Related work 
%\textbf{Related work}. 

%\textcolor{Sijia_color}{[Review existing works here.]}

%%% Motivation of AdaMM

%\textbf{Motivation of AdaMM in the ZO setting}. 
Due to the stochastic nature of ZO optimization, which arises from both  data sampling and random gradient estimation, existing ZO methods suffer from large variance of the noisy gradient compared to FO stochastic
methods  \cite{liu2018signsgd}. In practice, this   causes   poor  convergence performance and/or   function query efficiency. To partially mitigate these issues, ZO sign-based SGD (ZO-signSGD) was proposed by \cite{liu2018signsgd} with the rationale that taking the sign of random gradient estimates (i.e., normalizing gradient estimates elementwise) as the descent direction improves the robustness of gradient estimators to stochastic noise. 
Although ZO-signSGD has faster convergence speed than many existing ZO algorithms, it is  only guaranteed to converge to  a neighborhood of a solution. In the FO setting, 
 taking the sign of a stochastic gradient as the descent direction gives rise to signSGD \cite{bernstein2018signsgd}. 
The use of sign of stochastic gradients also appears in  
adaptive momentum methods (AdaMM) such as Adam \cite{kingma2014adam}, RMSProp \cite{krizhevsky2012imagenet}, AMSGrad \cite{reddi2018convergence},
Padam
\cite{chen2018closing}, and AdaFom \cite{chen2018convergence}. Indeed, it has been suggested by  \cite{balles18a} that AdaMM enjoy dual advantages of  sign descent and  variance adaption.  
% {In theory, 
% it was  shown by \cite{reddi2018convergence, chen2018convergence,zhou2018convergence}  that AdaMM has $O(\sqrt{d}/\sqrt{T})$ worst-case convergence rate.  
% In practice, since AdaMM  carefully designs 
% descent direction and   learning rate using
%   {momentum}  and  {adaptive} strategies, it often leads to superior  convergence performance than SGD-type methods \cite{kingma2014adam,reddi2018convergence,chen2018convergence,zhou2018convergence}. 
% }

Considering the motivation of ZO-signSGD and the success of AdaMM in FO optimization, one   question arises: Can we generalize AdaMM to the ZO regime? 
To answer this question, we develop the \underline{z}eroth-\underline{o}rder \underline{ada}ptive \underline{m}omentum \underline{m}ethod (ZO-AdaMM) and analyze its convergence properties in both convex and nonconvex settings for constrained optimization. 

\paragraph{Contributions}  
%The {contributions} of this paper are both theoretical and practical.
{\textit{Theoretically},  for both convex and nonconvex optimization,  we show that ZO-AdaMM is roughly a factor of $O(\sqrt{d})$ worse than that of the FO AdaMM algorithm,  where $d$ is the number of optimization variables. 
We also 
 show that the   \textit{Euclidean} projection  based
 AdaMM-type methods could suffer non-convergence issues for constrained optimization. This highlights the necessity of  \textit{Mahalanobis distance} based projection. And we establish the 
 %Instead, our 
 Mahalanobis distance based convergence analysis, which
 %By leveraging Mahalanobis distance based convergence measure, we 
 makes  the first step toward understanding  adaptive learning rate   methods for  nonconvex constrained optimization.}
%  convergence measure fails to characterize the stationarity of AdaMM-type methods for constrained nonconvex optimization, and    propose a new Mahalanobis distance based convergence measure. 
 %to analyze the convergence  of ZO-AdaMM. 
%  Our analysis also makes the first step toward understanding  adaptive learning rate   methods for  nonconvex constrained optimization. For both convex and nonconvex optimization,  We show that ZO-AdaMM is roughly a factor of $O(\sqrt{d})$ worse than that of the FO AdaMM algorithm,  where $d$ is the number of optimization variables.
%  In the convex analysis of ZO-AdaMM, we reveal how the optimality gap of function values is associated with the variance of random gradient estimates.
% In the nonconvex analysis, we
%  show that the conventional Euclidean distance based convergence measure fails to characterize the stationarity of AdaMM-type methods, and    propose a new Mahalanobis distance based convergence measure to analyze the convergence  of ZO-AdaMM. {Our analysis also makes the first step toward understanding  adaptive learning rate   methods for  nonconvex constrained optimization.}
%To explore practical applications,

\textit{Practically},
 we formalize the experimental comparison of ZO-AdaMM with $6$  state-of-the-art ZO  algorithms in the application of black-box adversarial attacks to generate 
 both per-image and universal adversarial perturbations. Our proposal could provide an experimental benchmark for future studies on ZO optimization. 
 Code to reproduce experiments is released at the anonymous link \url{https://github.com/KaidiXu/ZO-AdaMM}.
%  \textcolor{XC_color}{[XC:We also gave the first analysis of ZO-AdaMM/AdaMM for nonconvex constrained optimization. And through the analysis, we found ZO-NES should use a different projection (this is shown in Proposition \eqref{prop: imp_distance}).] [XC: We mentioned ZO-AdaMM is $O(\sqrt{d})$ slower than ZO-SGD, but it is faster in practice, should we make some explanation? Like providing some insights why it is slower or stating our analysis may not be tight and the analysis is challenging?]}
%  \textcolor{Sijia_color}{SL: I handle your questions in the pink texts.}
%We summarize  our main contributions  as below. 
% \begin{itemize}
%     \item Compared to  first-order AdaMM \cite{reddi2018convergence, chen2018convergence, zhou2018convergence}, we show that ZO-AdaMM requires $O(\sqrt{d})$ times more iterations, leading to  $O(d/\sqrt{T})$ convergence rate.
%     \item In the analysis of ZO-AdaMM for convex stochastic optimization, we reveal how the optimally gap of function values is associated with the variance of random gradient estimates. 
%     \item In the analysis of ZO-AdaMM for nonconvex stochastic optimization, we
%     propose a new Mahalanobis distance based convergence measure and show that the conventional Euclidean distance based convergence measure fails to characterize the stationary condition of constrained optimization in the  AdaMM framework 
% \end{itemize}

\paragraph{Related work} Many types of ZO algorithms have been developed, and their convergence rates have been rigorously studied  under different problem settings. 
We highlight some recent works as below.
%\cite{ghadimi2013stochastic,lian2016comprehensive,duchi2015optimal,liu2018_NIPS,gu2016zeroth,liu2018stochastic,nesterov2015random,liu_globalsip18,ghadimi2016mini,chen2018frank,sahu2018towards,gao2014information,liu2017zeroth}.
%see a summary in  Table\,\ref{table: SZO_complexity_T}. %\cite{ghadimi2013stochastic,lian2016comprehensive,duchi2015optimal,liu2018_NIPS,gu2016zeroth,liu2018stochastic,nesterov2015random,liu_globalsip18,ghadimi2016mini,chen2018frank,sahu2018towards,gao2014information,liu2017zeroth}
For unconstrained stochastic optimization,  ZO stochastic gradient descent (ZO-SGD) \cite{ghadimi2013stochastic} and ZO stochastic coordinate descent (ZO-SCD) \cite{lian2016comprehensive}   were proposed, which have $O(\sqrt{d}/\sqrt{T})$ convergence rate, where 
%$d$ is the number of optimization variables, and 
$T$ is the number of iterations. Compared to   FO stochastic algorithms, ZO optimization suffers a slowdown dependent on   the variable dimension $d$, e.g., $O(\sqrt{d})$ for ZO-SGD and ZO-SCD.  In \cite{duchi2015optimal}, the tightness of the dimension-dependent factor $O(\sqrt{d})$ has been proved in the framework of ZO stochastic mirror descent (ZO-SMD). In order to further improve the iteration complexity of ZO algorithms, the  technique of variance reduction was applied to ZO-SGD and ZO-SCD, leading to ZO stochastic variance reduced  algorithms  with an improved convergence  rate in $T$, namely,  $O(d/T)$ \cite{liu2018_NIPS,gu2016zeroth,liu2018stochastic}. This improvement is aligned with ZO gradient descent (ZO-GD)   for deterministic nonconvex programming \cite{nesterov2015random}.
Moreover,   ZO versions of 
%projected SGD (PSGD) \cite{liu_globalsip18},
proximal SGD (ProxSGD) \cite{ghadimi2016mini}, Frank-Wolfe (FW) \cite{balasubramanian2018zeroth, sahu2018towards,chen2018frank}, and online alternating direction method of multipliers  (OADMM) \cite{liu2017zeroth,gao2014information} have been developed for   constrained  optimization. 
{Aside from the recent works on ZO algorithms mentioned before, %gradient-free optimization has a long history and 
there is rich literature in derivative-free optimization (DFO). 
Traditional DFO methods can be classified into direct search-based methods and model-based methods. Both the two type of methods are mostly iterative methods. The difference is that direct search-based methods refines its search direction based on the queried function values directly, while a model-based method builds a model that approximates the function to be optimized and updates the search direction based on the model. Representative methods of developed in DFO literature include NOMAD \citep{le2011algorithm,audet2006mesh}, PSWarm \citep{vaz2009pswarm}, Cobyla \citep{powell1994direct}, and BOBYQA \citep{powell2009bobyqa}. More comprehensive discussion on DFO methods can be found in  \cite{rios2013derivative,audet2017derivative}.}

\vspace{-0.05in}
\section{Preliminaries: Gradient Estimation via ZO Oracle}
\label{sec: ZO_grad}
\vspace{-0.05in}

% In this section,
% we provide a background on random gradient  estimation via the ZO oracle that  only returns function values given possible query inputs. 

%     ZO optimization methods that we focus on in this paper mimic first-order algorithms  but  approximate  the full gradient (or stochastic gradient) using the random gradient estimate    \cite{nesterov2015random,duchi2015optimal,ghadimi2013stochastic}. 
%   Compared to Bayesian optimization (Gaussian process), derivative-free trust region method, genetic algorithm and other existing gradient-free methods \cite{shahriari2016taking,conn2009global,whitley1994genetic,conn2009introduction},   ZO optimization  has two main advantages: a) ease of implementation by just slight modification to  commonly-used gradient-based algorithms, and b) comparable convergence rate   to first-order algorithms.
 
  The ZO gradient estimate of a function $f$ is constructed by the forward difference of two function values 
  at a random unit direction:
{\small \begin{align}\label{eq: grad_rand}
    \hat{\nabla}f  (\mathbf x ) =
    % \frac{d}{q\mu} \sum_{i=1}^q \left \{ [  f ( \mathbf x + \mu \mathbf u_i ) - f ( \mathbf x )]  \mathbf u_i \right \},
    { (d/\mu) [ f ( \mathbf x + \mu \mathbf u ) - f ( \mathbf x  ) ] }  \mathbf u,
\end{align}%
}where $\mathbf u$ is a  random vector drawn uniformly from the sphere of a unit ball, and $\mu > 0$ is a small step size, known as the smoothing parameter.
In many existing work such as \citep{nesterov2015random,ghadimi2013stochastic},
 the random direction vector $\mathbf u$   was drawn from the standard Gaussian distribution. Here
 the use of  uniform distribution ensures that the   ZO gradient estimate
  \eqref{eq: grad_rand} is defined in a bounded space rather than the whole real space required for Gaussian. 
%   Suppose that $f$ is $L_c$-Lipschitz continuous, then
%   a direct upper bound on the random gradient estimate is
% {\small   \begin{align}\label{eq: trivial_bd_ZOest}
% \hspace*{-0.1in}    \| \hat{\nabla}f  (\mathbf x ) \|_2 \leq & ({d}/{ \mu}) \|  f(\mathbf x + \mu \mathbf u) - f(\mathbf x) \|_2 %\nonumber \\
%     %\leq & \frac{d}{\mu} L_c \mu \| \mathbf u \|_2^2
%     \leq  d L_c, 
% \end{align} %  
% }where we have used the definition of Lipshitz continuity and  $\| \mathbf u \|_2 = 1$. 
As will be evident later, the boundedness of random gradient estimates is one of important conditions in the convergence analysis of ZO-AdaMM.
%on our proposed ZO optimization method.

The rationale behind  the ZO gradient estimate   \eqref{eq: grad_rand} is that although it is a biased approximation to the true gradient of $f$, it is \textit{unbiased} to  the gradient of the
randomized smoothing version of $f$ with parameter $\mu$ \cite{duchi2015optimal,liu2018_NIPS,gao2014information}, i.e.,
{\small  \begin{align}\label{eq: fmu_smooth}
 {f}_\mu(\mathbf x) = & \mathbb E_{\mathbf u \sim U_{\mathrm{B}}}[f(\mathbf x + \mu \mathbf u )],
 %= \frac{1}{\alpha(d)}\int_{B} f(\mathbf x + \mu \mathbf u) d  \mathbf u,
 \end{align}%
}where $\mathbf u \sim U_{\mathrm{B}}$ denotes 
 the uniform distribution over the unit Euclidean ball $\mathrm{B}$.
%  \textcolor{Sijia_color}{In spite of the unbiasedness of $\hat{\nabla }f$ with respect to (w.r.t.) $\nabla f_\mu$, the variance of the ZO gradient estimate is always proportional to   the dimension $d$  regardless of the value of the smoothing parameter $\mu$. \xl{(why mention this here? It seems not quite related. maybe better to mention this when we compare the first order and ZO gap or after Prop~\ref{prop: zo_adamm_nonconvex}.)}}
 We review  properties of the smoothing function \eqref{eq: fmu_smooth} and connections to the ZO gradient estimator  \eqref{eq: grad_rand} in Appendix\,\ref{app: grad_properties}.

\vspace{-0.05in}
\section{AdaMM from First to Zeroth Order}
\vspace{-0.05in}

% In this section, we first  briefly review the AMSGrad algorithm proposed in \cite{reddi2018convergence}, which specifies the first-order AdaMM of our interest. We then propose 
% ZO-AdaMM by leveraging random gradient estimation, and  provide some rationale for its possible advantages and technical challenges over the existing ZO optimization methods. 

Consider a stochastic optimization  problem of the generic form 
% the following generic problem where we are minimizing a function $f$, expressed in the expectation form as follows %{[In general, (1) is not a standard stochastic optimization problem. It should be $\mathbb E_{\boldsymbol{\xi}}[ f(\mathbf x; \boldsymbol{\xi}) ]$. Then finite-sum becomes a special case.]}
{\small \begin{align}\label{eq:problem}
\min_{\mathbf x \in \mathcal X} f(\mathbf x)=\mathbb{E}_{\boldsymbol \xi}[f(\mathbf x; \boldsymbol{\xi })],
\end{align}% 
}where $\mathbf x \in \mathbb R^d$ are optimization variables, $\mathcal X$ is a closed convex set, $f$ is a differentiable (possibly  nonconvex) objective function, 
and $\boldsymbol \xi$ is a certain random variable that captures
environmental uncertainties. 
%e.g., 
%randomly selected data sample or random noise in online optimization. 
In problem \eqref{eq:problem}, if $\boldsymbol \xi$ %\textcolor{XL_color}{[XL: $\boldsymbol \xi$?]} 
obeys a uniform distribution built on empirical   samples $\{ \boldsymbol \xi_i  \}_{i=1}^n$, then we recover a finite-sum  formulation with the objective function 
$f(\mathbf x) = \frac{1}{n}\sum_{i=1}^n f(\mathbf x; \boldsymbol{\xi}_i)$.

% the {first-order} adaptive momentum method (AdaMM) that we focus on in this work.
 %specified by the AMSGrad algorithm in \cite{reddi2018convergence}. 
%  With the aid of random gradient estimation, we propose ZO-AdaMM, and 
 %the \underline{z}eroth-\underline{o}rder \underline{ada}ptive \underline{m}omentum \underline{m}ethod (ZO-AdaMM). 
 %Lastly, we demonstrate its connection to some state-of-the-art ZO optimization algorithms, and 
%  provide some rationale for its possible advantages over the existing ZO optimization methods. 

%\vspace*{-0.3in}

\paragraph{First-order AdaMM in terms of AMSGrad  \cite{reddi2018convergence}.}
% Suppose that the gradient of  $f(\cdot; \boldsymbol{\xi})$ is available, problem \eqref{eq:problem} can then be solved by  first-order optimization methods that obey the   generic iteration 
% $
% \mathbf x_{t+1} = \mathbf x_t - r_t \boldsymbol{\Delta}_t 
% $, where $\mathbf x_t$ denotes the solution updated at the $t$th iteration for all $t \in [T] = \{ 1,2,\ldots, T\}$, $T$ is the number of iterations,
% $\boldsymbol{\Delta}_t$ is a certain  descent direction, and $r_t > 0$ is a learning rate. When $\boldsymbol{\Delta}_t$ represents the stochastic gradient $\nabla_{\mathbf x} f(\mathbf x_t; \boldsymbol{\xi}_t)$ at time $t$ and $r_t$ becomes the constant or decaying learning rate, the studied generic iteration simplifies to   stochastic gradient descent (SGD). 
% Compared to SGD, AdaMM  can achieve better empirical convergence performance by carefully designing 
% descent direction and   learning rate via
%   \textit{momentum}  and  \textit{adaptive} strategies, respectively \cite{kingma2014adam,reddi2018convergence,chen2018convergence,zhou2018convergence}. 
% From the application side,  
% AdaMM has become the most popular method to train deep neural networks. 
% %It also uses inverse of exponential moving average of squared past gradients to adjust the learning rate
We specify the algorithmic framework of AdaMM  by AMSGrad \cite{reddi2018convergence}, a modified version of Adam \cite{kingma2014adam} with convergence guarantees for both convex and nonconvex optimization. %\cite{reddi2018convergence,chen2018convergence,zhou2018convergence,phuong2019convergence}. 
In the algorithm, the descent direction $\mathbf m_t$ is given by an
{exponential} moving average of the past gradients.
The learning rate $r_t$ is adaptively penalized by a  square root of exponential moving averages of squared past gradients.
% \textcolor{Sijia_color}{[SL: please see something about Reddi's fix. For example, compared to Adam \cite{kingma2014adam}, xxx. We refer readers to xx for more details.]}
% The full algorithm of AdaMM is indicated from Algorithm\,\ref{alg:zoadam} by just replacing the ZO gradient estimate with   the first-order stochastic gradient. Very recently,
\textcolor{black}{It has  been proved in 
\cite{reddi2018convergence,chen2018convergence,zhou2018convergence,phuong2019convergence} that AdaMM can reach  $O({1}/{\sqrt{T}})$\footnote{In the paper, we could  omit  $\log(T)$ in Big $O$ notation.}
%\footnote{We omit its possible dependency on $d$ for simplicity, but more accurate analysis will be provided later.}
 convergence rate.   % in terms of the reduction \xl{(what is the reduction?)} that can be obtained after $T$ iterations. 
 Here we omit its possible dependency on $d$ for simplicity, but more accurate analysis will be provided later} {in Section~\ref{sec:nonconvex} and \ref{sec:extend}}. %Throughout the paper, we also omit the logarithmic dependency in Big $O$ notation.

%  We elaborate on ZO-AdaMM  in Algorithm\,\ref{alg:zoadam}, where the stochastic gradient is replaced by the ZO gradient estimate in AdaMM.

% \begin{algorithm}[tb]
% \caption{ZO-AdaMM}
% \label{alg:zoadam}
% \begin{algorithmic}
%   \STATE {\bfseries Input:} $\mathbf x_1 \in \mathcal X$, 
% step sizes $\{ \alpha_t \}_{t=1}^T$, $\beta_{1,t}, \beta_2 \in (0, 1]$, and set  $\mathbf m_0$, $\mathbf v_0 $  and $\hat{\mathbf v}_0 $ 
% \FOR{$t =  1,2,\ldots, T$}
% \STATE let $\hat {\mathbf g}_t = \hat {\nabla} f_t (\mathbf x_t)$ by \eqref{eq: grad_rand}, where $f_t(\mathbf x_t) \Def f(\mathbf x_t; \boldsymbol{\xi}_t)$
% \STATE $\mathbf m_t = \beta_{1,t} \mathbf m_{t-1} + (1-\beta_{1,t}) \hat {\mathbf g}_t$
% \STATE $\mathbf v_t = \beta_2 \mathbf v_{t-1} + (1-\beta_2) \hat {\mathbf g}_t^2$
% \STATE $\hat{\mathbf v}_t = \max (\hat{\mathbf v}_{t-1}, \mathbf v_t)$, and $\hat{\mathbf V}_t = \diag (\hat{\mathbf v}_t)$
% \STATE $\mathbf x_{t+1} = \Pi_{\mathcal X, \sqrt{\hat{\mathbf V}_t}} (\mathbf x_t - \alpha_t\hat{\mathbf V}_t^{-1/2}  \mathbf m_t ) $
%   \ENDFOR
% \end{algorithmic}
% \end{algorithm}

\begin{wrapfigure}{R}{0.505\textwidth}
\vspace{-8mm}
\begin{small}
% \resizebox{0.5\textwidth}{!}{
\begin{minipage}{1\textwidth}
\begin{algorithm}[H]
\caption{ZO-AdaMM}
\label{alg:zoadam}
\begin{algorithmic}
  \STATE {\bfseries Input:} $\mathbf x_1 \in \mathcal X$, 
step sizes $\{ \alpha_t \}_{t=1}^T$, $\beta_{1,t}, \beta_2 \in (0, 1]$, and set  $\mathbf m_0$, $\mathbf v_0 $  and $\hat{\mathbf v}_0 $ 
\FOR{$t =  1,2,\ldots, T$}
\STATE let $\hat {\mathbf g}_t = \hat {\nabla} f_t (\mathbf x_t)$ by \eqref{eq: grad_rand},  $f_t(\mathbf x_t) \Def f(\mathbf x_t; \boldsymbol{\xi}_t)$
\STATE $\mathbf m_t = \beta_{1,t} \mathbf m_{t-1} + (1-\beta_{1,t}) \hat {\mathbf g}_t$
\STATE $\mathbf v_t = \beta_2 \mathbf v_{t-1} + (1-\beta_2) \hat {\mathbf g}_t^2$
\STATE $\hat{\mathbf v}_t = \max (\hat{\mathbf v}_{t-1}, \mathbf v_t)$, and $\hat{\mathbf V}_t = \diag (\hat{\mathbf v}_t)$
\STATE $\mathbf x_{t+1} = \Pi_{\mathcal X, \sqrt{\hat{\mathbf V}_t}} (\mathbf x_t - \alpha_t\hat{\mathbf V}_t^{-1/2}  \mathbf m_t ) $
  \ENDFOR
\end{algorithmic}
\end{algorithm}
\end{minipage}
% }
\end{small}
\vspace{-2mm}
\end{wrapfigure}

\textbf{ZO-AdaMM}.
By integrating AdaMM with the random gradient estimator \eqref{eq: grad_rand},  we obtain ZO-AdaMM   in Algorithm\,\ref{alg:zoadam}. Here 
the square root, the square, the maximum, and the division operators are taken elementwise. Also, $\Pi_{\mathcal X, \mathbf H} (\mathbf a)$ denotes the projection operation under Mahalanobis distance with respect to $\mathbf H$, i.e., 
$\argmin_{\mathbf x \in \mathcal X}   \| \sqrt{\mathbf H}  (\mathbf x - \mathbf a)  \|_2^2$.
If $\mathcal X  = \mathbb R^d$,  the projection step 
%in Algorithm\,\ref{alg:zoadam} 
simplifies to 
$\mathbf x_{t+1} = \mathbf x_t - \alpha_t \hat{\mathbf V}_t^{-1/2} \mathbf m_t$. Clearly, $\alpha_t \hat{\mathbf V}_t^{-1/2}$ and $\mathbf m_t$ can be interpreted as the adaptive learning rate and the momentum-type descent direction, which adopt  exponential moving averages as follows,
{\small \begin{align}\label{eq: mt}
%     \mathbf m_t =  (1-\beta_{1})\sum_{j=1}^t ( 
%      \beta_1^{t-j} \hat {\mathbf g}_j 
%   ),  
    \mathbf m_t = \sum_{j=1}^t \left [ 
    \left ( 
    \prod_{k=1}^{t-j} \beta_{1,t-k+1}
    \right )  (1-\beta_{1,j}) \hat {\mathbf g}_j
    \right ],~ \mathbf v_t = (1-\beta_2)  \sum_{j=1}^t ( \beta_2^{t-j} \hat {\mathbf g}_j^2 ). %\label{eq: vt}
\end{align}%
}Here we assume that  $\mathbf m_0 = \mathbf 0$, $\mathbf v_0 = \mathbf 0$ and \textcolor{black}{$0^0 = 1$ by convention, and let $\hat {\mathbf g}_t = \hat {\nabla} f_t (\mathbf x_t)$ by \eqref{eq: grad_rand} with $f_t(\mathbf x_t) \Def f(\mathbf x_t; \boldsymbol{\xi}_t)$.
}

%{\color{red} [do we need to point out that this is not the original Adam algorithm, rather reddi's fix with the max operation? should cite reddi's paper to comment on where the max comes from.]}

\textbf{Motivation and rationale behind ZO-AdaMM}.
% It was recently shown in \cite{liu2018signsgd} that   
% the ZO gradient estimates could  suffer from having larger noise variance than (first-order) stochastic
% gradients, and thus taking the sign of these gradient estimates (i.e., normalizing gradient estimates elementwise) as the descent direction improves the convergence performance of ZO optimization. 
First, gradient normalization helps noise reduction in ZO optimization as shown by \cite{ilyas2018blackbox,liu2018signsgd}. In the similar spirit,  
%In the similar spirit of ZO-signSGD proposed by \cite{liu2018signsgd}, 
ZO-AdaMM also  normalizes the descent direction $\mathbf m_t$ by $\sqrt{\hat{\mathbf v}_t}$.
\textcolor{black}{Particularly, compared to AdaMM, ZO-AdaMM prefers a small value of $\beta_2$ in practice, implying a strong favor to normalize the current gradient estimate; see Fig\,\ref{fig: beta_per_uncons_cons_uni} in Appendix.}
In the extreme case of  $\beta_{1,t} = \beta_2 \to 0$ and $\hat{\mathbf v}_t = \mathbf v_t$, ZO-AdaMM could reduce to ZO-signSGD \cite{liu2018signsgd} since $  \hat{\mathbf V}_t^{-1/2} \mathbf m_t = \mathbf m_t / \sqrt{\mathbf v_t} = \hat{\mathbf g}_t / \sqrt{\hat{\mathbf g}_t^2} = \mathrm{sign}(\hat {\mathbf g}_t) $ known from \eqref{eq: mt}. However, 
the downside of ZO-signSGD  is its worse convergence accuracy  than ZO-SGD, i.e., it only converges  to a neighborhood of a stationary point \textcolor{black}{even for unconstrained optimization}. Compared to ZO-signSGD, ZO-AdaMM is able to cover ZO-SGD as a special case when $\beta_{1,t} =0$, $\beta_2 = 1$, $\mathbf v_0 = \mathbf 1$ and $\hat{\mathbf v}_0 \leq \mathbf 1$ from Algorithm\,1.
% where we abuse $\leq(\geq)$ for both scalars and entry-wise comparison of vectors. 
Thus, we hope that with appropriate choices of $\beta_{1,t}$ and $\beta_2$, ZO-AdaMM could enjoy dual advantages of ZO-signSGD and ZO-SGD.
Another motivation comes from the possible presence of time-dependent gradient priors \cite{ilyas2018prior}. Given this, the use of past gradients in momentum also helps noise reduction. 

%\sijia{add a motivation figure here.}

%  First, gradient normalization helps noise reduction in the ZO setting (Sec. 3, Liu et al., 2019). Thus, the use of adaptive learning rate is motivated. Indeed, as shown in Fig. A1 (supplement), ZO-AdaMM prefers a small value of \beta2, implying a strong favor to normalize the current gradient estimate. Another motivation comes from the presence of time-dependent gradient priors (Sec. 3.1, Ilyas, et al., arXiv:1807.07978v2). Given this, the use of past gradients in momentum also helps noise reduction. 
\vspace{-0.05in}
\paragraph{Why is ZO-AdaMM difficult to analyze?} The convergence analysis of ZO-AdaMM  becomes significantly more challenging than existing ZO methods due to the involved coupling 
 among stochastic sampling, ZO gradinet estimation, momentum, adaptive learning rate, and  projection operation. In particular, the use of Mahalanobis distance in projection step plays a key role on convergence guarantees. 
% the traditional convergence measure \cite{ghadimi2016mini}  for constrained nonconvex optimization fails to characterize the stationary condition  of  ZO-AdaMM. 
  And the conventional variance bound on ZO gradient estimates is insufficient to analyze the convergence  of ZO-AdaMM due to the use of adaptive learning rate. %\xl{(any reference on this?)}.
 %see Proposition~\ref{prop: imp_distance} for example.
In the next sections, we will carefully study the convergence of ZO-AdaMM under different  settings.

\vspace{-0.06in}
\section{Convergence Analysis of ZO-AdaMM for Nonconvex Optimization}\label{sec:nonconvex}
\vspace{-0.06in}

 \textcolor{black}{In this section, 
 we begin by providing a deep understanding on the importance of  Mahalanobis distance used in ZO-AdaMM (Algorithm\,\ref{alg:zoadam}), and then introduce the Mahalanobis distance based convergence  analysis for both unconstrained and constrained nonconvex optimization.
 %we first propose a new  Mahalanobis distance based convergence measure  and   provide   insights on why it  matters in   ZO-AdaMM. 
%Based on that, we analyze the convergence of ZO-AdaMM for both unconstrained and constrained nonconvex optimization. 
Our analysis  makes the first step toward understanding adaptive learning rate methods for {nonconvex constrained} optimization. 
Throughout the section, we make the following assumptions.
}

\textcolor{black}{
\textbf{A1}: $f_t(\cdot ) \Def f(\cdot; \boldsymbol{\xi}_t)$ %is differentiable and
has $L_g$-Lipschitz continuous gradient,
  where $L_g>0$.}

\textcolor{black}{
\textbf{A2}:  $f_t$ has $\eta$-bounded stochastic gradient  $\| \nabla f_t (\mathbf x)\|_\infty \leq \eta$.}

% \textcolor{Sijia_color}{
% \textbf{A3}: $f_t$ %is differentiable and
% is $L_c$-Lipschitz continuous,   where $L_c>0$. 
% }

% \textcolor{Sijia_color}{We note that \textbf{A1} and \textbf{A2} are standard assumptions used in analyzing the convergence of adaptive momentum-type methods \cite{reddi2018convergence,chen2018convergence, zhou2018convergence}. As will be evident later,  the introduction of \textbf{A3} enables us to bound the maximum $\ell_\infty$ norm of ZO gradient estimates over time.
% }

% \textcolor{Sijia_color}{
% \textbf{A\ref{sec:nonconvex}.2:}  The variance of stochastic gradients  $\nabla f_t(\mathbf x_t)$ is bounded as
% $\mathbb E_{\boldsymbol{\xi}}[ \| \nabla f_t(\mathbf x_t) -\nabla f(\mathbf x_t) \|_2^2 ] \leq \sigma^2$.
% }

\subsection{Importance of Mahalanobis distance based  projection operation}

\textcolor{black}{
%We begin by providing a deep understanding on the importance of  Mahalanobis distance used in ZO-AdaMM (Algorithm\,\ref{alg:zoadam}), and then introduce the Mahalanobis distance based convergence measured used in our analysis.
Recall from Algorithm\,\ref{alg:zoadam} that %for constrained optimization, 
ZO-AdaMM takes the projection operation $\Pi_{\mathcal X, \sqrt{\hat{\mathbf V}_t}}(\cdot)$ onto the constraint set $\mathcal X$ under Mahalanobis distance with respect to (w.r.t.) $\hat{\mathbf V}_t$. In some recent adversarial learning algorithms
%generating adversarial examples 
\cite{kurakin2016adversarial,ilyas2018black},   the Euclidean projection $\Pi_{\mathcal X}(\cdot)$    was used in both FO and ZO AdaMM-type methods rather than the Mahalanobis distance  based projection in Algorithm\,\ref{alg:zoadam}. However,  such an  implementation could lead to \textit{non-convergence}: Proposition\,\ref{prop: imp_distance} shows the non-convergence  issue of Algorithm\,\ref{alg:zoadam} using the Euclidean projection operation  when solving a simple linear program subject to $\ell_1$-norm constraint. This is an important point which is ignored in design of many algorithms on adversarial training \cite{madry2017towards}.
%designing process especially for robust machine learning.
}

% \textcolor{XC_color}{Existing formal analysis and implementation for Adam/AMSGrad are all targeting unconstrained optimization, since they are designed for neural network training. Since, ZO-AdaMM is also designed for applications involving constrained optimization such as adversarial attacks, we need to find a way to implement it in constrained optimization. The simplest way to do so is to project each iterate into the feasible set, which resembles projected SGD. However, such an implementation will lead to non-convergence in practice (see Proposition \ref{prop: imp_distance}).}

\begin{myprop}\label{prop: imp_distance}
%\textcolor{XL_color}{[XL: Restated. Check.]}\textcolor{XC_color}{[XC: Checked.]} 
Consider the following problem
{\small \begin{align}\label{eq: counter_example}
\begin{array}{ll}
 \displaystyle \minimize_{\mathbf x = [x_1, x_2]^T}      -2 x_1 - x_2 ;  & 
 \st     ~ |x_1 + x_2| \leq 1,
\end{array}
\end{align}%
}then Algorithm\,\ref{alg:zoadam}, initialized by  $\mathbf x = [0.5, 0.5]^T$, using the Euclidean projection $\Pi_{\mathcal X}(\cdot)$  converges to a fixed point $[0.5, 0.5]^T$ rather than a stationary point of \eqref{eq: counter_example}.
%In Algorithm\,\ref{alg:zoadam}, if the projection step is defined under Euclidean distance with $\mathbf V_t = \mathbf I$, then the converged solution is not necessarily a stationary point of problem \eqref{eq:problem}. A specific example is given by 
%\begin{align}\label{eq: counter_example}
%\begin{array}{ll}
% \displaystyle \minimize_{\mathbf x = [x_1, x_2]^T}    & -2 x_1 - x_2   \\
% \st     &  |x_1 + x_2| \leq 1,
%\end{array}
%\end{align}
%where \textcolor{XC_color}{if we initialize $x_1 = [0.5, 0.5]^T$,} the converged point $[0.5, 0.5]^T$ is not a stationary point of problem {\color{red}[mention initialization]} \eqref{eq: counter_example}.
\end{myprop}
\noindent 
\textit{\textbf{Proof}:  The proof investigates a special case of Algorithm\,\ref{alg:zoadam}, projected  signSGD; See Appendix \ref{app: prop_divergence}.} %\hfill $\square$

 Proposition\,\ref{prop: imp_distance} indicates that replacing the Mahalanobis distance based projection in Algorithm \ref{alg:zoadam} with Euclidean projection will lead to a divergent algorithm, highlighting the importance of using Mahalanobis distance. However, the use of Mahalanobis distance based projection  complicates the convergence analysis, especially in constrained optimization. Accordingly, we define a Mahalanobis based convergence measure that can simplify the analysis and can be converted into the traditional convergence measure.

\textcolor{black}{
%We next present the Mahalanobis distance based convergence measure.
Let $\mathbf x^+ = \mathbf x_{t+1}$, $\mathbf x^{-} = \mathbf x_t$,
$\mathbf g = \mathbf m_t$,   $\omega  = \alpha_t$ and $\mathbf H = \hat{\mathbf V}_t^{1/2}$, the  projection step of  Algorithm\,\ref{alg:zoadam} can be written in the generic form
{\small \begin{equation} \label{eq: project}
     \mathbf x^+ =\argmin_{\mathbf x \in  \mathcal X} \{  %\langle g,u \rangle 
     \inp{\mathbf g}{ \mathbf x}
    + (1/\omega ) D_{\mathbf H}(\mathbf x,\mathbf x^{-}) \},
\end{equation}%
}where $D_{\mathbf H}(\mathbf x,\mathbf x^{-}) = \|\mathbf H^{1/2} (\mathbf x-\mathbf x^{-})\|^2/2$ gives the {Mahalanobis distance} w.r.t. $\mathbf H$, and  $\| \cdot \|$ denotes   $\ell_2$ norm. Based on \eqref{eq: project}, the concept of \textit{gradient mapping} \cite{ghadimi2016mini} is  given by
{\small \begin{equation}\label{eq: prox}
P_{\mathcal X,\mathbf H}(\mathbf x^{-}, \mathbf g,\omega ) \Def   (\mathbf x^{-} - \mathbf x^{+})/\omega.
\end{equation}%
}The gradient mapping $P_{\mathcal X, \mathbf H}(\mathbf x^{-}, \mathbf g,\omega ) $  yields a natural interpretation: %good alternative of the original gradient $\mathbf g$: 
a  projected version of $\mathbf g$ at the point $\mathbf x^{-}$ given the learning rate $\omega$, yielding
$\mathbf x^+ = \mathbf x^{-} - \omega   P_{\mathcal X,\mathbf H}(\mathbf x^{-1}, \mathbf g,\omega ) $. We note that different from \cite{ghadimi2016mini,reddi2016proximal}, the gradient mapping in \eqref{eq: prox}  is defined on  the projection under the  Mahalanobis distance $D_{\mathbf H} (\cdot, \cdot)$ rather than the  Euclidean distance.
}

With the aid of \eqref{eq: prox}, we propose the Mahalanobis distance based convergence measure for ZO-AdaMM: %  general settings as 
{\small \begin{align}\label{eq: conv_measure_general}
\|   \mathcal  G(\mathbf x_t)\|^2 \Def   \|\hat {\mathbf V}_t^{1/4} P_{\mathcal X,{\hat {\mathbf V}_t^{1/2}}}( \mathbf x_t, \nabla f(\mathbf x_t),\alpha_t) \|^2.
\end{align}%
% Clearly, if $\hat{\mathbf V}_t = \mathbf I$, namely, applying Euclidean distance to \eqref{eq: project}, then
% the convergence measure $ \mathcal G(\mathbf x_t)$ reduces to the definition of the conventional gradient mapping in \cite{ghadimi2016mini,reddi2016proximal}. 
%Moreover, 
}If $\mathcal X = \mathbb R^d$, then the convergence measure \eqref{eq: conv_measure_general}  reduces to 
{\small \begin{align}\label{eq: conv_measure_uncons}
    \|\hat {\mathbf V}_t^{-1/4}  \nabla f(\mathbf x_t) \|^2 , 
\end{align}%
}which corresponds to the squared Euclidean norm of gradient in a linearly transformed coordinate system $\mathbf y_t= \hat {\mathbf V}_t^{1/4} \mathbf x_t$. As will be evident later, the measure \eqref{eq: conv_measure_uncons} can be  transformed to the conventional measure $\| \nabla f(\mathbf x_t) \|^2 $ for unconstrained optimization.
  We remark that {Mahalanobis (M-) distance facilitates our convergence analysis in an equivalently transformed space, over which the analysis can be generalized from the conventional projected gradient descent framework.}
    % It is motivated by the fact that multiplying gradients by a positive definite matrix in gradient descent is equivalent to running gradient descent in a linear transformed coordinate system. 
%  Recall the ZO-AdaMM update rule using true gradients with $\beta_{1,t} = 0$ and $\mathcal X = R^d$ is $\mathbf x_{t+1} = \mathbf x_t - \alpha \hat  {\mathbf V}_t^{-1/2}  \nabla f(\hat {\mathbf x}_t)$. 
{\color{black}To get intuition, let us consider a simpler first-order case with the $\mathbf x$-descent step  given by Algorithm\,\ref{alg:zoadam} as    $\beta_{1,t} = 0$ and $\mathcal X = \mathbb R^d$: $\mathbf x_{t+1} = \mathbf x_t - \alpha \hat  {\mathbf V}_t^{-1/2}  \nabla f( {\mathbf x}_t)$. Note that the ZO case is more involved but follows the same intuition. 
Upon defining $\mathbf y_t \triangleq \hat {\mathbf V}_t^{1/4} \mathbf x_t$, the $\mathbf x$-update can then be rewritten as the update rule in $\mathbf y$:  $\mathbf y_{t+1} = \mathbf y_t -
 \alpha \hat {\mathbf V}_t^{-1/4}  \nabla_{} f( {\mathbf x}_t)$. 
 Since 
%  We also have the gradient of $f(\mathbf x_t)= f(\hat {\mathbf V}_t^{-1/4} \mathbf y_t)$ w.r.t $\mathbf y_t$  is 
 $\nabla_{\mathbf y_t} f(\mathbf x_t)
 %= \frac{\partial }{\partial {\mathbf y}_t} f(\hat {\mathbf V}_t^{-1/4} \mathbf y_t) 
 = ( \frac{\partial  {\mathbf  x}_t
}{\partial  {\mathbf  y}_t} )^T   \nabla f( {\mathbf x}_t)
% \frac{\partial f
% %(\mathbf x_t)
% }{\partial {\mathbf  x}_t} 
=  \hat {\mathbf  V}_t^{-1/4}  \nabla f( {\mathbf x}_t)$, the $\mathbf y$-update,
$\mathbf y_{t+1} = \mathbf y_t - \alpha
\nabla_{\mathbf y} f( {\mathbf x}_t)$,
   obeys the gradient descent framework.
%This means the gradient descent update rule on $\mathbf y$ is
%$\mathbf y_{t+1} = \mathbf y_t -
% \alpha \hat {\mathbf V}_t^{-1/4}  \nabla_{} f( {\mathbf x}_t)$ (multiplying  $ \hat {\mathbf V}_t^{1/4}$ at its both sides),   reduces to the original update with respect to $\mathbf x$ 
 %by  multiplying  $ \hat {\mathbf V}_t^{1/4}$ at its both sides.
 {In the constrained case, a similar but more involved analysis can be made,  showing that  the \textit{M-projection in the $\mathbf x$-coordinate system} is \textit{equivalent} to the  \textit{Euclidean projection in the $\mathbf y$-coordinate system} which makes projected gradient descent applicable to the update in $\mathbf y$. By contrast, the direct use of  \textit{Euclidean projection in the $\mathbf x$-coordinate system} leads to \textit{divergence} in  ZO-AdaMM (Proposition\,\ref{prop: imp_distance}).}}

\vspace{-0.05in}
\subsection{Unconstrained nonconvex optimization}
\vspace{-0.05in}

% We discuss the convergence rate of ZO-AdaMM for unconstrained optimization in this section. As we have seen in the convergence guarantee for stochastic constrained optimization, the convergence measure will converge to constant defined by variance of the gradient estimator, to make the algorithm converge to a stationary point, a increasing batch size is required. However, in unconstrained optimization, we do not need to increase batch size to make the algorithm converge.  The analysis is significantly different from constrained optimization and a tighter dependency on $d$ can be achieved in unconstrained optimization.

We next demonstrate the convergence analysis of ZO-AdaMM for unconstrained nonconvex optimization. 
%where the convergence measure is given by \eqref{eq: conv_measure_uncons}, which will also be related to  
% $\|  \nabla f(\mathbf x_t) \|^2$ later. 
% Unlike Proposition\,\ref{prop: optgap2var_convex} in the convex analysis, we are unable to 
% it becomes much more involved to   link the optimality gap with the variance of ZO gradient estimates. 
In  Proposition\,\ref{prop: zo_adamm_nonconvex},
we begin by exploring the relationship between the convergence measure \eqref{eq: conv_measure_uncons} and ZO gradient estimates; \textit{See Appendix \ref{app: prop_main_uncons} for proof.}

% \begin{align}
% \|\hat V_t^{1/4} P_{X,{\hat V_t^{1/2}}}(x_t \nabla f(x_t),\alpha_t) \|^2
% \end{align}
% is a natural convergence measure for ZO-AdaMM, this measure  reduces to 
% \begin{align}
%   \|\hat V_t^{-1/4}  \nabla f(x_t) \|^2
% \end{align}
% when $X = \mathbb{R}^d$. Thus, we first state a main theorem for this convergence measure and then discuss its relationship to  $\|  \nabla f(x_t) \|^2$.

% \begin{myprop}\label{prop: zo_adamm_nonconvex}
% Suppose that  $\mathcal X = \mathbb R^d$, $\hat V_{1,ii}^{1/2} \geq c, \,  \forall i \in [d]$ and $\|\hat g_t\|_\infty \leq G_{zo}, \, \forall t \in [T]$. Let $\mu = 1/\sqrt{Td}$, $\alpha_t = 1/\sqrt{Td}$ and $\gamma \Def \frac{\beta_1}{\beta_2} < 1$.  Pick $R$ uniformly randomly from 1 to $T$, ZO-AdaMM yields
% \begin{align} \label{eq: final_bound}
%     \mathbb E \left[ \| \hat V_t^{-1/4} \nabla f(x_R)\|^2 \right]
%     \leq & \frac{  L^2}{2c}\frac{d}{T} + 2\frac{\sqrt{d}}{\sqrt{T}}D_f \nonumber \\
%     &+ 2\frac{d}{T}E \left[\left(\frac{\eta G _{zo} }{1-\beta_1} + 2\eta^2\right)   \frac{1}{c}   \right]  \nonumber\\
%     &+ \frac{\sqrt{d}}{\sqrt{T}} \frac{4L +  5L\beta_1^2}{2(1-\beta_1)^2} \frac{1-\beta_1}{1-\beta_2}\frac{1}{1-\gamma} 
% \end{align}
% \end{myprop}

\begin{myprop}\label{prop: zo_adamm_nonconvex}
\textcolor{black}{
Suppose that \textbf{A1}-\textbf{A2} hold and let
$\mathcal X = \mathbb R^d$, $\hat {\mathbf v}_{0}^{1/2} \geq c \mathbf 1$, 
$f_{\mu}(\mathbf x_1) - \min_{\mathbf x} f_{\mu}(\mathbf x) \leq D_f$, $\beta_{1,t} = \beta_1$, $\gamma \Def {\beta_1}/{\beta_2} < 1$, 
$\mu = 1/\sqrt{Td}$, and $\alpha_t = 1/\sqrt{Td}$ in Algorithm\,\ref{alg:zoadam}, then ZO-AdaMM yields
{\small \begin{align} \label{eq: final_bound}
  \mathbb  E\left[ \left \| \hat {\mathbf V}_R^{-1/4} \nabla  f(\mathbf x_R) \right \|^2 \right]
    \leq  &   \frac{  L_g^2}{2c}\frac{d}{T} + 2 D_f \frac{\sqrt{d}}{\sqrt{T}}  +  \frac{L_g(4+5\beta_1^2) (1-\beta_1)}{2(1-\beta_1)^2(1-\beta_2)(1-\gamma)} %\frac{1-\beta_1}{1-\beta_2}\frac{1}{1-\gamma} 
    \frac{\sqrt{d}}{\sqrt{T}} \nonumber \\
    &  + \frac{2}{c}  \mathbb E  \left [    2\eta^2 + \frac{\eta   \max_{t \in [T]} \{  \| \hat{\mathbf g}_t \|_\infty \} }{1-\beta_1}      \right ]  \frac{d}{T}, %\nonumber\\
    % &\quad \quad + \frac{\sqrt{d}}{\sqrt{T}} \frac{4L_g+  5L_g\beta_1^2}{2(1-\beta_1)^2} \frac{1-\beta_1}{1-\beta_2}\frac{1}{1-\gamma},
\end{align}%
}where $\mathbf x_R$ is picked uniformly randomly from  $\{  \mathbf x_t \}_{t=1}^T$, and $\hat {\mathbf g}_t = \hat {\nabla} f_t (\mathbf x_t)$ by \eqref{eq: grad_rand}.
%$\hat{\mathbf g}_t$ is gradient estimate at time $t$.% defined in Algorithm\,\ref{alg:zoadam}. %\sijia{$L_c$ or $\eta$?}
% \sijia{Is $\eta$ representing $ \| \nabla f_t(\mathbf x_t) \|_\infty \leq \eta $ }
}
\end{myprop}
%\textbf{Proof}:  See Appendix \ref{app: prop_main_uncons}
%\textcolor{Sijia_color}{[SL: please check the proof when you denote  the starting point, which is $\mathbf m_0$, $\mathbf v_0$ but $\mathbf x_1$ in algorithm\,1.]}
%\hfill $\square$

Proposition \ref{prop: zo_adamm_nonconvex} implies that the convergence  rate of ZO-AdaMM has a dependency on    ZO gradient estimates in terms of   $ G_\mathrm{zo} \Def \max_{t\in [T]}\{  \| \hat{\mathbf g}_t \|_\infty \} $.
% This is different from Proposition\,\ref{prop: regret_mid} for  convex optimization, where we can directly associate the optimality gap of function values with the variance of ZO gradient estimates. Such a discrepancy is introduced by the difference in convergence measures.
Moreover, if we consider the FO AdaMM \cite{chen2018convergence,zhou2018convergence} in which the ZO gradient estimate $\hat{\mathbf g}_t$ is replaced with the stochastic gradient, then
one can 
simply assume $\max_{t\in [T]}\{  \| {\mathbf g}_t \|_\infty \}$ to be a {dimension-independent  constant} under \textbf{A2}. 
%where $\hat{\mathbf g}_t = \nabla f_t$.
However, in the ZO setting,  $G_\mathrm{zo}$ is no longer independent of $d$. For example, it could  be directly  bounded by
 %in the order of $O(d)$
%using the naive bound 
$ \| \hat{\nabla}f  (\mathbf x ) \|_2 \leq ({d}/{ \mu}) \|  f(\mathbf x + \mu \mathbf u) - f(\mathbf x) \|_2 
    \leq  d L_c$ 
% % when $f_t$ is $L_c$-Lipschitz continuous,
% % $
% %   \| \hat{\nabla}f_t  (\mathbf x ) \|_2 \leq  ({d}/{ \mu}) \|  f_t(\mathbf x + \mu \mathbf u) - f_t(\mathbf x) \|_2
% %     \leq  d L_c$,
% % where we have used the definition of Lipshitz continuity and  $\| \mathbf u \|_2 = 1$.
%  using the naive bound  \eqref{eq: trivial_bd_ZOest} \xl{(Move \eqref{eq: trivial_bd_ZOest} to here seems more natural.)} 
under the following assumption:

\textbf{A3}: $f_t$  is $L_c$-Lipschitz continuous.

%by imposing     Lipschitz continuity of $f_t$.
 %Thus, a tighter bound on $G_{\mathrm{zo}}$ is required.
\textcolor{black}{
% It becomes highly nontrivial to 
% derive a tighter bound on $G_{\mathrm{zo}}$ due to the presence of $\max$ operation.
In Proposition\,\ref{prop: g_zo_bound}, 
we show that  the dimension-dependency of $G_\mathrm{zo}$ can be further improved by using sphere concentration results; \textit{See Appendix \ref{app: prop_concentration} for proof.}
%Thus, we introduce  Lipschitz continuity of $f_t$ to simply bound the function difference 
%can be implied by Lipschitz continuity and thus we introduce Lipschitz condition on the objective functions for simplicity.
}

\begin{myprop} \label{prop: g_zo_bound}
%Define $G_{zo} \Def \max_{t\in [T]}$, 
%Upon defining 
\textcolor{black}{
Under \textbf{A3},     $\max \{ d, T\} \geq  3$, and given $\delta \in (0,1)$,
% $f_t$ is $L_{c}$-Lipschitz  continuous,
}
 %and define
%$\xi \Def  4 \log (dT/\delta)$ for any $\delta$ such that $\xi \geq 2 + 2\log \xi$,
%Given $\delta \in (0,1)$, 
then with probability at least $1-\delta$, %we have
{\small \begin{align}\label{eq: single_bound}
    %G_{zo} \Def
    \max_{t\in [T]} \{  \| \hat{\mathbf g}_t \|_\infty \} \leq 2 L_c \sqrt{ d\log ({dT}/{\delta}) } .
    %2\eta\sqrt{ d\log ({dT}/{\delta}) } .
\end{align}%
}\end{myprop}
%\textbf{Proof:} See Appendix \ref{app: prop_concentration}. 
%\sijia{Give a remark on bounding this is not trivial.}
% \textcolor{Sijia_color}{[Please modify the proof, the assumption has changed, and the constant is changed from $\eta$ to $L_c$?]} \textcolor{XC_color}{[updated]}
%\hfill $\square $
% \textcolor{Sijia_color}{[SL: Needs extra insights to say the above bound is tight. Need to discuss even we naively using $\ell_2$, what rate will obtain?]}
\textcolor{black}{Here we provide some insights on Proposition \ref{prop: g_zo_bound}. Since
% Recall that the gradient estimate $\hat{\mathbf g}_t$ is a unit vector on a spere times $d$ and the function difference, the $L_{\infty}$ norm of his could be in the order of $O(d)$ in worst case. However, observing that
the unit random  vector used to define $\hat{\mathbf g}_t$ is uniformly sampled on a sphere, $\| \hat{\mathbf g}_t \|_\infty$ can be improved to $O(\sqrt{d})$ with high probability. This is a tight bound since when the function difference is a constant, the lower bound satisfies $\| \hat{\mathbf g}_t \|_\infty = \Omega(\sqrt{d})$ by sphere concentration.  It is also not surprising that our  bound \eqref{eq: single_bound} grows with $T$ since we bound the maximum $\| \hat{\mathbf g}_t \|_\infty$ over $T$ realizations with high probability. The time-dependence is required to compensate the growth of the probability that there exists an estimate with the extreme $\ell_{\infty}$ value versus time.}
\textcolor{black}{Note that as long as $T$ has  polynomial rather than exponential  dependency on $d$, we then always have $    \max_{t\in [T]} \{  \| \hat{\mathbf g}_t \|_\infty \} = O(\sqrt{d\log{(d)}})$.}
%then $\log T$ always grows slow.
% \xl{(also mention that we only need polynomial time $T$ on dimensions rather than exponential, so $\log T$ grows slow.)}
% {In this case, the probability that there exists an estimate with the extreme $\ell_{\infty}$ norm grows with $T$. Thus, an increase in upperbound need to be introduced to compensate the growth of this probability.}\sijia{The last sentence is not clear. an increase in upper bound? Do you want to say that the dependence on $T$ is introduced to compensate the growth of the probability there exists an estimate with the extreme $\ell_{\infty}$ norm versus $T$.}
Based on  Proposition\,\ref{prop: zo_adamm_nonconvex} and Proposition\,\ref{prop: g_zo_bound},
%Using Proposition \ref{prop: g_zo_bound}, 
the convergence rate of ZO-AdaMM is provided by Theorem \ref{thm: zo_adamm_high_prob}; \textit{See Appendix \ref{app: proof_adamm_high_prob} for proof.}
\begin{mythr} \label{thm: zo_adamm_high_prob}
{Suppose that \textbf{A1} and \textbf{A3} hold. 
% conditions  in Proposition\,\ref{prop: zo_adamm_nonconvex} and \ref{prop: g_zo_bound} hold.
%in particular assume that $\mathcal{X}\equiv \mathbb{R}^{d}$. %define $\xi = 4 \log (dT/\delta)$, for any $\delta$ such that $\xi \geq 2 + 2\log \xi$, with probability at least $1- \delta$, 
Given parameter settings  in Proposition\,\ref{prop: zo_adamm_nonconvex} and \ref{prop: g_zo_bound},
%$\delta \in (0,1)$, 
then with probability at least $1-1/(T\sqrt{d})$, 
ZO-AdaMM yields
}
% \begin{align} \label{eq: rate_adap_measure}
%   & \mathbb  E\left[ \left \| \hat {\mathbf V}_R^{-1/4} \nabla  f(\mathbf x_R) \right \|^2 \right]
%     \leq  \frac{  L_g^2}{2c}\frac{d}{T} + 2 D_f \frac{\sqrt{d}}{\sqrt{T}} \nonumber \\
%     &\quad \quad +  \frac{L_g(4+5\beta_1^2) (1-\beta_1)}{2(1-\beta_1)^2(1-\beta_2)(1-\gamma)} %\frac{1-\beta_1}{1-\beta_2}\frac{1}{1-\gamma} 
%     \frac{\sqrt{d}}{\sqrt{T}} \nonumber \\
%     &\quad \quad + \frac{2}{c}  \mathbb E \left[   2\eta^2 + \frac{2\eta^2  \sqrt{ d\log ({dT}/{\delta})} }{1-\beta_1}      \right]  \frac{d}{T}. %\nonumber\\
%     % &\quad \quad + \frac{\sqrt{d}}{\sqrt{T}} \frac{4L_g+  5L_g\beta_1^2}{2(1-\beta_1)^2} \frac{1-\beta_1}{1-\beta_2}\frac{1}{1-\gamma}
% \end{align}
{\small \begin{align} \label{eq: rate_adap_measure}
   & \mathbb  E\left[ \left \| \hat {\mathbf V}_R^{-1/4} \nabla  f(\mathbf x_R) \right \|^2 \right]
     = O \left ( \sqrt{d}/\sqrt{T} + d^{1.5}/T \right ).
\end{align}%
}\end{mythr}
%\textbf{Proof}: See Appendix \ref{app: proof_adamm_high_prob}. 
%The result can be obtained immediately after substituting \eqref{eq: single_bound} into \eqref{eq: final_bound} \textcolor{XC_color}{and use the fact that $\eta \leq L_{c}$}.
% \textcolor{Sijia_color}{[Please update the proof since the previous proposition changed.]}\textcolor{XC_color}{[updated]}
%\hfill $\square$

%  \textcolor{black}{Theorem \ref{thm: zo_adamm_high_prob} is a counterpart of Proposition \ref{prop: zo_adamm_nonconvex} conditioned on the event of Proposition \ref{prop: g_zo_bound} with $\delta = 1/T\sqrt{d}$. %It seems it is just substituting \eqref{eq: single_bound} into \eqref{eq: final_bound} but one cannot do so due to the bias introduced by conditioning on the event in Proposition \ref{prop: g_zo_bound}. More efforts need to be taken dealing with a new bias introduced by conditioning.  
%  }
 {We can also extend the convergence rate of ZO-AdaMM in 
 Theorem\,\ref{thm: zo_adamm_high_prob} using the measure $\mathbb E [\|\nabla f(\mathbf x_R)\|^2]$.
%  that ZO-AdaMM achieves $O(\sqrt{d}/\sqrt{T} + d^{1.5}/T)$ convergence rate for nonconvex unconstrained optimization under the proposed convergence measure in \eqref{eq: conv_measure_uncons}, which can be further transferred in terms of $\mathbb E [\|\nabla f(\mathbf x_R)\|^2]$. 
%\textcolor{XC_color}{ Since  $ \hat V_{t,ii}^{-1/2} \geq {1}/{\max_{t\in [T]} \{  \| \hat{\mathbf g}_t \|_\infty \} }$ (by the update rule)}, 
%where we recall that $\hat V_{t,ii} = \hat g_{t,i}^2$ for any $i$  and $G_\mathrm{zo} =    \max_{t\in [T]} \{  \| \hat{\mathbf g}_t \|_\infty \} $, 
%  one can obtain directly from \eqref{eq: single_bound} that
Since  $ \hat V_{t,ii}^{-1/2} \geq {1}/{\max_{t\in [T]} \{  \| \hat{\mathbf g}_t \|_\infty \} }$ (by the update rule), 
 we obtain from \eqref{eq: single_bound} that
% we have 
% $
%     \mathbb  E\left[ \left \| \hat {\mathbf V}_t^{-1/4} \nabla  f(\mathbf x_R) \right \|^2 \right] \geq \frac{1}{2 \eta \sqrt{d \log (dT/\delta )}} E\left[ \| \nabla f(x_R)\|^2 \right]$.
% That is, 
{\small \begin{align} \label{eq: high_prob_bound}
     \mathbb E\left[ \| \nabla f(\mathbf x_R)\|^2 \right]   
    \leq & 2 L_c \sqrt{d \log (dT/\delta )} \mathbb  E\left[ \left \| \hat {\mathbf V}_R^{-1/4} \nabla  f(\mathbf x_R) \right \|^2 \right].
\end{align}
}%\textcolor{XC_color}{conditioned on the event of Proposition \ref{prop: g_zo_bound} with $\delta = 1/Td^{0.5}$}. 
Theorem\,\ref{thm: zo_adamm_high_prob}, together with \eqref{eq: high_prob_bound}, implies $O(d/\sqrt{T} + d^2/T)$ convergence rate of ZO-AdaMM under the conventional measure. We remark that compared to the FO rate ${O}(\sqrt{d}/\sqrt{T} + d/T)$ \cite{zhou2018convergence} of   AdaMM for   unconstrained nonconvex optimization under \textbf{A1}-\textbf{A2}, %\xl{(what are \textbf{A\,4.1}-\textbf{A\,4.2}?)}, 
ZO-AdaMM suffers $O(\sqrt{d})$ and  $O(d)$ slowdown on
 the rate term   $O(1/\sqrt{T})$ and $O(1/T)$, respectively. 
% A similar  dimension-dependent slowdown was found for 
This dimension-dependent slowdown is similar to 
ZO-SGD versus SGD shown by \citep{ghadimi2013stochastic}. %see Table\,\ref{table: SZO_complexity_T}. 
We also remark that compared to FO-AdaMM, ZO-AdaMM requires additional \textbf{A3} to bound the $\ell_\infty$ norm of ZO gradient estimates. 
%  We also note that the rate of existing  ZO algorithms often   focuses only on the  sub-linear term of $T$.
%  {\color{red}[is the last sentence precise? I thought that most ZO algorithm cares about dimension $d$.]}
 %\xl{(Do we want to give a table for the first-order and ZO gaps on $d$ for SGD and Adam?)}
%It has been shown in \cite{duchi2012randomized}that for zeroth-order optimization, the complexity lower bound is $\Omega(d/\sqrt{T})$ 
}

\vspace{-0.05in}
\subsection{Constrained nonconvex optimization }
\vspace{-0.05in}

{
To analyze ZO-AdaMM in a  general constrained case, one needs to  handle the coupling effects from all three factors: momentum,  adaptive learning rate, and  projection operation. 
%In Theorem\,\ref{thm: nonconvex_cons}, 
Here we  focus on addressing the coupling issue in the last two factors, which  yields our results on ZO-AdaMM   at $\beta_{1,t} = 0$. This is  equivalent to the ZO version of RMSProp \cite{krizhevsky2012imagenet}   with Reddi's convergence fix in \cite{reddi2018convergence}.
When the  momentum factor comes into play, the scenario becomes much more complicated.  We leave the answer to the  general case  $\beta_{1,t} \neq 0$ for future research. 
Even for  SGD  with momentum, 
we are  not aware of any successful convergence analysis
for stochastic {constrained} nonconvex optimization. 
}

\textcolor{black}{It is known from  SGD 
\cite{ghadimi2016mini} that 
 the presence of projection  induces a stochastic bias  (independent of iteration number $T$) for constrained nonconvex optimization.
In Theorem\,\ref{thm: nonconvex_cons}, we show that the same challenge holds for ZO-AdaMM. Thus, one has to adopt the variance reduced   gradient estimator, which induces higher  querying complexity than the estimator \eqref{eq: grad_rand}; \textit{See Appendix\,\ref{app: proof_uncons} for proof}.
%   over multiple random direction vectors $\{ \mathbf u_i \}_{i=1}^q$ and $b$ i.i.d. random mini-batch samples $\{ \boldsymbol {\xi}_j \}_{j=1}^b$,
}

\begin{mythr} \label{thm: nonconvex_cons}
Suppose that \textbf{A1}-\textbf{A2} hold, $\hat {\mathbf v}_{0}^{1/2} \geq c \mathbf 1$, $f_{\mu}(\mathbf x_1) - \min_{\mathbf x} f_{\mu}(\mathbf x) \leq D_f$, $\alpha_t = \alpha \leq \frac{c}{L_g}$, {$\mu = \frac{1}{\sqrt{Td}}$}, and $\beta_{1,t} = 0$  in Algorithm\,\ref{alg:zoadam},  then the convergence rate of ZO-AdaMM under \eqref{eq: conv_measure_general} satisfies 
%Assume $\min_i{{V_{1,ii}^{1/2}}} \geq c$, set $\alpha_t = \alpha \leq \frac{c}{L_g}$ and $\beta_1 = 0$, 
%By uniformly randomly picking $R$ from $[T]$, we have
{\small \begin{align}
\mathbb E [ \|  \mathcal G(\mathbf x_R) \|^2 ]    \leq &  \frac{6 D_f }{\alpha T} + \frac{3 L_g^2 d}{4 c T} +\frac{6 \eta^2 }{c^4 T}    (\max_{t\in [T]} \mathbb E[\|\hat {\mathbf g}_t - f_{\mu}(\mathbf x_t)\|^2 ]+ d \eta^2)    +  \frac{3c+9}{c}  \max_{t\in [T]}\mathbb E[\|\hat {\mathbf g}_t - f_{\mu}(\mathbf x_t)\|^2 ], \nonumber
\end{align}%
}where  $\mathbf x_R$ is picked uniformly randomly from  $\{  \mathbf x_t \}_{t=1}^T$, $\mathcal G(\mathbf x)$ has been defined in \eqref{eq: conv_measure_general},  
%$f_{\mu}(\mathbf x) = \mathbb E_{\boldsymbol \xi}[f_{\mu}(\mathbf x; \boldsymbol \xi)]$, 
and $f_{\mu}$ is the smoothing function of $f$ defined in \eqref{eq: fmu_smooth}. %\sijia{Is the same $c$ for $\hat{\mathbf v}_0$ and $\alpha$?}
%\sigma_{\hat g}^2 \geq E[\|\hat g_t - f_{\mu}(x_t)\|^2$ is the upper bound on the variance of $\hat g_t$.

%\textcolor{XC_color}{[XC: changed a little bit, the variance is max variance over T iterations.]}
\end{mythr}
%\textbf{Proof}: {See Appendix\,\ref{app: proof_uncons}.} 
%\textcolor{Sijia_color}{SL: what is $\mu$?}
%\hfill $\square$

{Theorem\,\ref{thm: nonconvex_cons} implies that regardless of the number of iterations $T$, ZO-AdaMM only  converges to a solution's neighborhood  whose size
 is determined by    the variance of ZO gradient estimates $ \max_{t\in [T]}\mathbb E[\|\hat {\mathbf g}_t - f_{\mu}(\mathbf x_t)\|^2 ]$.
 %, which is introduced by the coupling between the noise on gradient estimates and the projection operator.  
 To make this term diminishing, we consider the following variance reduced gradient estimator  built on multiple stochastic samples  and random direction vectors \cite{liu2018signsgd}, 
%  the variance reduced gradient estimator \eqref{eq: grad_rand_ave} is needed, which yields   $\sigma_{\hat g}^2 = O \left ( {1}/{b} + {d}/{(bq)} \right)$. Thus, by choosing $b \geq  \sqrt{T} $ and $q \geq 1$, the rate of ZO-AdaMM in Theorem\,\ref{thm: nonconvex_cons} becomes $\mathbb E [ \|  \mathcal G(\mathbf x_R) \|^2 ]  = O(d/\sqrt{T})$. 
}
\textcolor{black}{
{\small \begin{align}\label{eq: grad_rand_ave}
  \hspace*{-0.08in}  \hat{\mathbf g}_t = \frac{1}{bq} \sum_{j \in \mathcal I_t} \sum_{i=1}^q \hat{\nabla} f(\mathbf x_t; \mathbf u_{i,t}, \boldsymbol{\xi}_j), \quad 
  \hat{\nabla} f(\mathbf x_t; \mathbf u_{i,t}, \boldsymbol{\xi}_j) \Def 
  \frac{d[  f ( \mathbf x_t + \mu \mathbf u_{i,t}; \boldsymbol{\xi}_j ) - f ( \mathbf x_t; \boldsymbol{\xi}_j )]  }{\mu} \mathbf u_{i,t},
    %\frac{ d [ f ( \mathbf x + \mu \mathbf u ) - f ( \mathbf x  ) ] }{\mu} \mathbf u,
\end{align}%
}where $\mathcal I_t$ is a mini-batch containing $b$  stocahstic samples at time $t$, 
and $\{ \mathbf u_{i,t} \}_{i=1}^q$ are $q$   random direction vectors  at time $t$. 
% We remark that different the average random gradient estimator used in \cite{liu2018signsgd}, the estimator \eqref{eq: grad_rand} only requires $(b+q)$ times   random sampling since it is free of re-sampling $q$ i.i.d. random direction vectors  at each stocahstic sample $\boldsymbol{\xi}_i$. 
{We present the variance of \eqref{eq: grad_rand_ave} in Lemma\,\ref{lemma: smooth_f_random_stochastic}, whose proof is induced from \textnormal{\cite[Proposition\,2]{liu2018signsgd}} by using $\| \nabla f_t \|_2^2 \leq d \| \nabla f_t \|_\infty^2 = d  \eta^2 $ in \textbf{A2}.}
}

\begin{mylemma}
\label{lemma: smooth_f_random_stochastic}
Suppose that  \textbf{A1}-\textcolor{black}{\textbf{A2} hold}, 
%and $f_t$ is $L_c$-Lipschtiz continuous, 
%with bounded variance 
% $\mathbb E_{\boldsymbol{\xi}}[ \| \nabla f_t(\mathbf x_t) -\nabla f(\mathbf x_t) \|_2^2 ] \leq \sigma^2$, 
then for $\mu \leq 1/\sqrt{d}$, the variance of \eqref{eq: grad_rand_ave} yields
{ \small 
\begin{align}
%\small
& \mathbb E \left [
\| 
\hat {\mathbf g}_t - \nabla f_\mu (\mathbf x_t)
\|_2^2
\right ] 
= O \left  ( {d}/{b} + {d^2}/{q} \right )  . 
\label{eq: second_moment_grad_random}
\end{align}% 
}
% where the expectation is taken over all randomness including both random direction sampling and stochastic sampling, and $f_{\mu}(\cdot ; \boldsymbol{\xi}) $ denotes the smoothing function of $f(\cdot; \boldsymbol{\xi})$. 
\end{mylemma}
%\textbf{Proof:} \textcolor{black}{
%The results can be directly obtained from 
%\textnormal{\cite[Proposition\,2]{liu2018signsgd}} by using %$\| \nabla f_t \|_2^2 \leq d \| \nabla f_t \|_\infty^2 = d  \eta^2 $ by \textbf{A2}.} \hfill $\square$

\textcolor{black}{
% It is also known from \eqref{eq: second_moment_grad_random} that for  $\mu \leq 1/\sqrt{d}$, the
% variance of the ZO gradient estimate is present, leading to  $O(d/q)$  dimension-dependency    caused by random gradient estimation  and $O(1/b)$ dependency on the  variance of stochastic samples. By letting $\hat{\mathbf g}_t = \hat{\nabla } f(\mathbf x_t)$ given by \eqref{eq: grad_rand}, 
Based on Lemma\,\ref{lemma: smooth_f_random_stochastic}, the rate of ZO-AdaMM in    Theorem\,\ref{thm: nonconvex_cons} becomes
$
    \mathbb E [ \|  \mathcal G(\mathbf x_R) \|^2 ]  = O ( {d}/{T} + {d}/{b} + {d^2}/{q}  )
$. Note that if \textbf{A3} holds, then   the dimension-dependency can be improved by $O(d)$ factor based on Lemma\,\ref{lemma: smooth_f_random_stochastic}.
% Thus, by choosing $b =   O(  \sqrt{T} ) $ and $q = O(T)$, we then obtain $O(d/\sqrt{T} + d^2/T)$ convergence rate. 
To the best of our knowledge, even in  the FO case we are not aware of existing   convergence rate analysis on     adaptive learning rate methods  for nonconvex contrained optimization. 
%under the Mahalanobis distance based convergence measure \eqref{eq: conv_measure_general}.
}

\vspace{-0.05in}
\section{Extended Analysis of ZO-AdaMM}\label{sec:extend}
\vspace{-0.05in}
% In this section, we provide more thorough discussion on ZO-AdaMM including rate analysis for convex  optimization and comparison of iteration/query complexity   with existing ZO methods.   

\paragraph{ZO-AdaMM for constrained convex optimization}
Different from the nonconvex case, 
 the convergence of ZO-AdaMM for convex optimization is commonly measured by  the average regret 
 %(in terms of the optimality gap of function values  at each iteration)
$
    R_T = \mathbb E \left [ \frac{1}{T} \sum_{t=1}^T f_t(\mathbf x_t) - % \min_{\mathbf x \in \mathcal X} 
    \frac{1}{T} \sum_{t=1}^T f_t(\mathbf x^*) \right ]
$ \cite{reddi2018convergence,chen2018closing}, where recall that  $f_t(\mathbf x_t) = f(\mathbf x_t; \boldsymbol{\xi}_t)$, and $\mathbf x^*$ is the optimal solution. 
We provide the average regret with the ZO gradient estimates by leveraging its connection to the smoothing function  of $f_{t}$ in Proposition\,\ref{prop: regret_mid}; \textit{see Appendix \ref{app: prf_prop_regret_mid} for proof.}
\begin{myprop}\label{prop: regret_mid}
Suppose that $\alpha_t = \alpha/\sqrt{t}$, $\beta_{1,t} =\beta_{1}/t$ with 
$\beta_{1,1} = \beta_1$, 
$\beta_1, \beta_2 \in [0, 1)$,  $\gamma \Def \beta_1/\sqrt{\beta_2} < 1$ and $\mathcal X$ has bounded diameter $D_{\infty}$, then  ZO-AdaMM for convex optimization yields
{\small \begin{align}\label{eq: regret_mid}
& R_{T,\mu} \Def \mathbb E \left [ \frac{1}{T} \sum_{t=1}^T f_{t,\mu}(\mathbf x_t) - % \min_{\mathbf x \in \mathcal X} 
    \frac{1}{T} \sum_{t=1}^T f_{t,\mu}(\mathbf x^*) \right ] \nonumber \\
    \leq &       \frac{D_\infty^2 \sum_{i=1}^d \mathbb E  [\hat{v}_{T,i}^{1/2} ] }{\alpha  (1-\beta_1) \sqrt{T}} +  \frac{ D_\infty^2}{2(1-\beta_1) T} \sum_{t=1}^T \sum_{i=1}^d \frac{ \beta_{1} \mathbb E [ \hat{v}_{t,i}^{1/2} ] }{\alpha \sqrt{t}} + \frac{\alpha \sqrt{1+\log{T}}      \sum_{i=1}^d \mathbb E \| \hat {\mathbf g}_{1:T,i} \|   }{(1-\beta_1)^2(1-\gamma)\sqrt{1-\beta_2}T}  .
\end{align}%
}where $f_{t,\mu}$ denotes the smoothing function of $f$ defined by \eqref{eq: fmu_smooth}, 
$\hat v_{t,i}$ denotes the $i$th element of the vector $\hat {\mathbf v}_t$ defined in Algorithm\,\ref{alg:zoadam}, and  $\hat{\mathbf g}_{1:T,i} \Def [ \hat{g}_{1,i}, \ldots, \hat{g}_{T,i} ]^\top$.
\end{myprop}
%\textbf{Proof}: {See Appendix \ref{app: prf_prop_regret_mid}.} \hfill $\square$

We remark that Proposition\,\ref{prop: regret_mid} would reduce to  \cite[Theorem 4]{reddi2018convergence} by replacing ZO gradient estimates $\hat{\mathbf g}_{1:T,i}$ and $\hat v_{t,i} $ with FO gradients
${\mathbf g}_{1:T}$ and $ v_t $. However, it was recently shown by \cite{phuong2019convergence} that the proof of  \cite[Theorem 4]{reddi2018convergence} is problematic. To address   the proof issue, in Proposition\,\ref{prop: regret_mid} we present a simpler fix   than \cite[Theorem\,4.1]{phuong2019convergence} and show  that the conclusion of \cite[Theorem 4]{reddi2018convergence} keeps correct. 
%even if its proof is problematic. 
In the FO setting, the  rate of AdaMM under   \textbf{A2} for constrained convex optimization  is given by  $O(d/\sqrt{T})$ \cite[Corollary\,4.4]{chen2018closing}. Here \textbf{A2} provides the direct $\eta$-upper bound on $|g_{t,i}|$ and $\hat{v}_{t,i}^{1/2}$, and we consider worst-case rate analysis without imposing extra assumptions like sparse gradients\footnote{The work \cite{ilyas2018prior} showed the lack of sparsity in gradients while generating adversarial examples.}. 
%suppose that $\hat {\mathbf v}_0 = \mathbf 0$. 
In the ZO setting, we need  further 
bound $   | \hat{ g}_{t,i} | $ and $\hat v_{t,i}$  and link $R_{T,\mu}$ to $R_T$, where the former is achieved by Proposition\,\ref{prop: g_zo_bound}  and the latter is achieved by the relationship between $f_t$ and its smoothing function $f_{t,\mu}$ shown in Lemma\,\ref{lemma: smooth_f_random}-(a), yielding
$   f_{t}(\mathbf x_t)  -  f_{t}(\mathbf x^*)  \leq f_{t,\mu}(\mathbf x_t) - f_{t,\mu}(\mathbf x^*) + 2 \mu  L_c$. Thus, given $\mu \leq d/ \sqrt{T}$ and assuming %that \textbf{A1}-\textbf{A2}  and 
conditions in Proposition\,\ref{prop: g_zo_bound} hold, then the rate of ZO-AdaMM %for convex optimization 
becomes 
$
    R_{T} \leq  2 \mu L_c + R_{T, \mu} = O(d^{1.5}/\sqrt{T})
$,
 which is $O(\sqrt{d})$ worse than the  AdaMM. 

% where we have used the facts that $\hat{v}_{t,i}^{1/2} \leq \eta$ and 
% $\frac{1}{d} \sum_{i=1}^d \sqrt{\| {\mathbf g}_{1:T,i} \|_2^2} \leq \sqrt{\frac{1}{d} \sum_{i=1}^d \| {\mathbf g}_{1:T,i} \|^2 } = \sqrt{\frac{1}{d} \sum_{t=1}^T \| {\mathbf g}_{t} \|^2 }$
\vspace{-0.05in}
\paragraph{Comparison with other ZO methods}
%%% ZO-AdaMM versus others
Since the existing  convergence analysis for different ZO methods is built on different problem settings and assumptions. The direct comparison over the convergence rates might not be fair enough. Thus, in Table\,\ref{table: SZO_complexity_T} we compare ZO-AdaMM with  others ZO methods from $4$ perspectives: a) the type of gradient estimator, b) the setting of smoothing parameter $\mu$, c) convergence rate, and d) function query complexity. 

Table\,\ref{table: SZO_complexity_T} shows that for unconstrained nonconvex optimization, the convergence of ZO-AdaMM achieves worse dependency  on  $d$ than ZO-SGD \cite{ghadimi2013stochastic}, ZO-SCD \cite{lian2016comprehensive} and ZO-signSGD \cite{liu2018signsgd}. However, it has milder  choice of $\mu$ than ZO-SGD,    less query complexity than ZO-SCD, and   no $T$-independent convergence bias    compared to ZO-signSGD.
Also,
for constrained nonconvex optimization, ZO-AdaMM yields the similar rate to ZO-ProxSGD \cite{ghadimi2016mini}, which also implies ZO projected SGD (ZO-PSGD). 
% requires a large $q$ to mitigate the convergence bias, however, it
% has milder 
% condition on $\mu $  and  weaker assumption \textbf{A2} (versus \textbf{A3}) 
% than ZO-ProxSGD \cite{ghadimi2016mini}. 
%To  
%achieves the    similar rate, ours requires $O(q)$ more function queries. 
 For constrained convex optimization, the rate of ZO-AdaMM is $O(d)$ worse than ZO-SMD \cite{duchi2015optimal} but ours has the significantly improved dimension-dependency in $\mu$. 
 {We also highlight that  at the first glance, ZO-AdaMM has a worse $d$-dependency (regardless of choice of $\mu$) than ZO-SGD. However, even in the FO setting, AdaMM has an extra $O(\sqrt{d})$ dependency in the worst case due to the effect of (coordinate-wise) gradient normalization  when bounding the distance of two consecutive updates. Thus, in addition to comparing with different ZO methods, Table\,\ref{table: SZO_complexity_T} also summarizes the convergence performance of FO AdaMM.  Note that our rate yields $O(\sqrt{d})$ slowdown compared to  FO AdaMM though bounding ZO gradient estimate norm requires  stricter assumption. 
 %Such a slowdown factor could be tight based on the optimal ZO rate analysis in  \cite{duchi2015optimal}.
 }
 %Note that  
% the rate of  ZO-SMD is optimal, however,   the tightness of
% ours has not been proven. 

\iffalse
in addition to comparing with different ZO methods, Table\,\ref{table: SZO_complexity_T} also summarizes the convergence performance of FO AdaMM.  As we can see,  our rate yields $O(\sqrt{d})$ slowdown compared to  FO AdaMM. Such a slowdown factor could be tight based on the optimal ZO rate analysis in  \cite{duchi2015optimal}. However, unlike $O(\sqrt{d}/\sqrt{T})$ rate for SGD under  \textbf{A2}, AdaMM has an extra $O(\sqrt{d})$ dependency in the worst case due to the effect of (coordinate-wise) gradient normalization  when bounding the distance of two consecutive updates.
\fi
%\cite[Lemma\,A.2]{zhou2018convergence}. 
% \sijia{Is the reference correct?  when bounding the distance of two consecutive updates \cite[Lemma\,A.2]{zhou2018convergence}.}

\vspace{-0.05in}
\begin{table*}[htb]
\centering
\caption{Summary of convergence rate and  query complexity of various ZO algorithms given $T$ iterations. %{\red[Sahu et al uses small ``t"]}
}
\label{table: SZO_complexity_T}
\begin{adjustbox}{max width=1\textwidth }
\begin{threeparttable}
\begin{tabular}{|c|c|c|c|c|c|}
\hline
Method        & \begin{tabular}[c]{@{}c@{}} Assumptions %\\ setting
\end{tabular}       & \begin{tabular}[c]{@{}c@{}}Gradient\\estimator \end{tabular} 
&     \begin{tabular}[c]{@{}c@{}}Smoothing\\parameter $\mu$   \end{tabular} 
&  \begin{tabular}[c]{@{}c@{}}Rate \end{tabular}       
& \begin{tabular}[c]{@{}c@{}}Query \end{tabular} 
\\ \hline  
% ZO-GD  \cite{nesterov2015random}
% & nonconvex, unconstrained 
% &  {GauGE$^{1}$}
% &  $ O\left (\frac{1}{\sqrt{dT}} \right )$ 
% & $\mathbb E [ \| \nabla f(\mathbf x_T) \|_2^2] = O\left (\frac{d}{T} \right )$  & $O\left (T \right ) $ 
% %$O\left (|\mathcal B|  T \right ) $ 
% \\ \hline
ZO-SGD  \cite{ghadimi2013stochastic} & 
\begin{tabular}[c]{@{}c@{}}NC$^1$, UCons$^1$, \textbf{A1}, \textbf{A3}$^2$ \end{tabular} 
&  GauGE$^1$
& $O \left (\frac{1}{d\sqrt{T}}\right )$   &
$O\left ( \frac{\sqrt{d}}{\sqrt{T}}  + \frac{d}{T} 
\right ) $
&   $O\left (T \right ) $  %$O\left (|\mathcal B|  T \right ) $ 
\\ \hline
ZO-SCD \cite{lian2016comprehensive} & NC, UCons, \textbf{A1}, \textbf{A3}$^2$  &  CooGE$^1$
%&$T$ 
& $O \left (\frac{1}{\sqrt{T}} + \frac{1}{\sqrt{d}}
%\min \{ \frac{1}{(dT)^{-1/4}}, \frac{1}{\sqrt{d}} \} 
\right 
)$   & 
$ O \left ( \frac{\sqrt{d}}{\sqrt{T}}
+ \frac{d}{T}
\right )$  & 
\begin{tabular}[c]{@{}c@{}}$O\left ( d T \right )$ %\\$|\mathcal S|$ is \# of coordiantes 
\end{tabular} 
% $O\left (|\mathcal B| |\mathcal S| T \right ) $
\\ \hline
ZO-signSGD \cite{liu2018signsgd}  & NC, UCons, \textbf{A1}, \textbf{A3}  &  sign-UniGE$^1$
& $O  \left ( \frac{1}{\sqrt{dT}}  \right  )$ &   $O (\frac{\sqrt{d}}{\sqrt{T}} + \frac{\sqrt{d}}{\sqrt{b}} + \frac{d}{\sqrt{bq}} 
%\frac{\sqrt{d}}{\sqrt{|\mathbf B|}} + \frac{d}{\sqrt{q |\mathcal B|}} 
 %\frac{d}{\sqrt{ |\mathcal B|}}
 )^3$ &
\begin{tabular}[c]{@{}c@{}} $O\left (bq  T \right ) $  %\\$|\mathcal B|$ is minibatch size
\end{tabular} 
%$O\left (|\mathcal B| q T \right ) $ 
\\ \hline 
% ZO-SVRG \cite{liu2018_NIPS} &
% nonconvex, unconstrained
% &
% UniGE$^{1}$   
% &
% $O\left ( \frac{1}{\sqrt{dT}}  \right )$ &
% \begin{tabular}[c]{@{}c@{}} $\mathbb E [ \| \nabla f(\mathbf x_T) \|_2^2] = O\left (\frac{{d}}{{T}} + \frac{\delta ({\mathcal B} )}{\sqrt{|\mathbf B|}}  \right )$  
%  \\ $\delta(\mathcal B) \in \{ 0,1\}$  
% \end{tabular} 
% & 
% \begin{tabular}[c]{@{}c@{}} $O \left (  n P  +  |\mathcal B| Pm \right )^2$
%   % \\ $n$ is full batch size, $T = Pm$ 
%   \end{tabular}
% %$O \left (  q n S + q |\mathcal B| S m  \right )$, $T = Sm^{**}$
% \\ \hline
% ZO-PSGD \cite{liu_globalsip18} &
% NC, Cons, \textbf{A1}, \textbf{A3} 
% &
% central-UniGE$^1$  
% &
% $O\left ( \frac{ 1}{\sqrt{d  bq}}\right )$ & $ O \left ( \frac{1}{\sqrt{T}}  + \frac{1}{b} + \frac{d}{bq} \right )$
% & $O\left (bq  T \right ) $ 
% \\ \hline
\begin{tabular}[c]{@{}c@{}} ZO-ProxSGD /  \\
ZO-PSGD
\cite{ghadimi2016mini} %\\$|\mathcal B|$ is minibatch size
\end{tabular} 
&
NC, Cons$^4$, \textbf{A1}, \textbf{A3}
%composite$^{***}$
&
GauGE 
&
$ O\left ( \frac{1}{\sqrt{dT}}\right )$ &
$ O \left ( %\frac{d}{|\mathcal B| q T} + 
\frac{d^2}{q  T} + \frac{d}{q } 
\right )
$ &  
$O\left (q T \right ) $ 
\\ \hline
ZO-SMD \cite{duchi2015optimal}&
C, Cons, \textbf{A3}
&
GauGE/UniGE 
& $O\left ( \frac{ 1}{ d t }\right )$
 & $ O\left ( \frac{ \sqrt{d}}{\sqrt{T} }\right )$ & $O(T)$
\\ \hline
% ZO-OADMM \cite{liu2017zeroth} &
% C, Cons
% &
% GauGE/UniGE   
% %&  $T$ 
% &
% $O\left ( \frac{1}{d^{1.5}t}\right )$ &
% $   O \left ( \frac{\sqrt{d}}{\sqrt{T}} \right )$ &  $O\left (  T \right ) $ 
% \\ \hline
\textit{AdaMM} \cite{chen2018convergence,zhou2018convergence} &
\begin{tabular}[c]{@{}c@{}} NC, UCons, \textbf{A1}, \textbf{A2} \end{tabular}
&
SGE$^1$   
& n/a
% \begin{tabular}[c]{@{}c@{}} 
% $O\left ( \frac{1}{\sqrt{dT}}\right )$ 
% \end{tabular}
&
$O \left ( \frac{\sqrt{d}}{\sqrt{T}} + \frac{d}{T} \right )$ 
&  n/a 
\\ \hline
\textit{AdaMM} \cite{reddi2018convergence,chen2018closing,phuong2019convergence} &
\begin{tabular}[c]{@{}c@{}} C, Cons,   \textbf{A2} \end{tabular}
&
SGE   
& n/a
% \begin{tabular}[c]{@{}c@{}} 
% $O\left ( \frac{1}{\sqrt{dT}}\right )$ 
% \end{tabular}
&
$O \left ( \frac{{d}}{\sqrt{T}} \right )$ 
&  n/a 
\\ \hline
\textbf{ZO-AdaMM} &
\begin{tabular}[c]{@{}c@{}} NC, UCons, \textbf{A1}, \textbf{A3} \end{tabular}
&
UniGE   
&
\begin{tabular}[c]{@{}c@{}} 
$O\left ( \frac{1}{\sqrt{dT}}\right )$  \end{tabular}
&
$O \left ( \frac{d}{\sqrt{T}} + \frac{d^2}{T} \right )$ 
&  $O\left (   T \right ) $  
\\ \hline
\textbf{ZO-AdaMM} &
\begin{tabular}[c]{@{}c@{}} NC, Cons, \textbf{A1}, \textbf{A3}  %\textbf{A2} 
\\ $\beta_{1,t} = 0$ \end{tabular}
&
UniGE   
%&  $T$ 
&
\begin{tabular}[c]{@{}c@{}} 
$O\left ( \frac{1}{\sqrt{dT}} \right )$ \end{tabular}
&
% $O \left ( \frac{d}{T} + \frac{d}{b} + \frac{d^2}{q}  \right )$ 
$O \left ( \frac{d}{T} + \frac{1}{b} + \frac{d}{q}  \right )$ 
&  $O\left (  bq T \right ) $  \\ \hline 
\textbf{ZO-AdaMM} &
\begin{tabular}[c]{@{}c@{}}  C, Cons, \textbf{A3} \end{tabular}
&
UniGE   
%&  $T$ 
&
\begin{tabular}[c]{@{}c@{}} 
$O\left ( \frac{d}{\sqrt{T}} \right )$ \end{tabular}
&
% $\mathbb E [ f(\mathbf x_T) - f(\mathbf x^*) ] =   O \left ( \sqrt{1+\frac{d}{q}} \frac{\sqrt{d}}{\sqrt{T}}  \right )$ 
$O \left ( \frac{d^{1.5}}{\sqrt{T}}  \right )$ 
&  $O\left ( T \right ) $  \\ \hline 
\end{tabular}
\begin{tablenotes}
    \small
     \item[1] \textit{Abbreviations}. NC: 
     \underline{N}on\underline{c}onvex; UCons: \underline{U}n\underline{c}onstrained;  GauGE: \underline{Gau}ssian random vector based      \underline{g}radient \underline{e}stimate;  
     UniGE: \underline{Uni}form random vector based      \underline{g}radient \underline{e}stimate;  
      CooGE: \underline{Coo}rdinate-wise \underline{g}radient \underline{e}stimate; SGE: \underline{s}tochastic (first-order) \underline{g}radient \underline{e}stimate
    \\
    \item[2] Assumption of bounded variance of stochastic gradients is implied from \textbf{A3}.
    % Notations: $\mathcal S$ is the index set of coordinates, and $|\mathcal S|$ is the set size. 
    % $\mathcal B$ is  minibatch with batch size $|\mathcal B| \leq n$.  In ZO-SVRG, $\delta(\mathcal B) = 0$ if $\mathcal B$ is the full batch, and $0$ otherwise. And $T = Pm$, where $P$ and $m$ are   lengths of outer and inner loops, respectively.\\
    \item[3] Convergence of ZO-signSGD  is measured by $\mathbb E [ \| \nabla f(\mathbf x_T) \|_2 ]$ rather than its square used in other algorithms for nonconvex optimization.
    % $\phi(\mathbf x_t)$ characterizes the dual gap, namely \cite{sahu2018towards}, $\phi(\mathbf x_t) = \{ \max_{\mathbf x \in \mathcal X} \inp{\nabla f(\mathbf x_t)}{\mathbf x_t -\mathbf x} \}$.
    % \item[4] ZO-ProxSGD implies ZO-PSGD \cite{liu_globalsip18} when the proximal operator becomes   projection onto constraint sets. 
\end{tablenotes}
\end{threeparttable}
\end{adjustbox}
\vspace*{-0.1in}
\end{table*}

\vspace{-0.1in}
\section{Applications to Black-Box  Adversarial Attacks}
\vspace{-0.05in}

In this section, we demonstrate the effectiveness of ZO-AdaMM by experiments on generating black-box adversarial examples. 
Our experiments will be performed on Inception V3 \cite{szegedy2016rethinking} using  %CIFAR-10 and 
ImageNet \cite{deng2009imagenet}. Here we focus on two types of black-box adversarial attacks: \textit{per-image} adversarial perturbation \cite{xu2018structured} and \textit{universal} adversarial perturbation against multiple images \cite{chen2017zoo,ilyas2018blackbox,suya2017query,cheng2018query}. For each type of attack, we allow both constrained and unconstrained optimization problem settings.
We compare our propos ed ZO-AdaMM method with $6$ existing ZO algorithms:
 ZO-SGD, ZO-SCD and ZO-signSGD for unconstrained optimization, and ZO-PSGD, ZO-SMD  and ZO-NES  for constrained optimization.
The first $5$ methods have been summarized in Table\,\ref{table: SZO_complexity_T}, and
  ZO-NES refers to the black-box attack generation method 
 in \cite{ilyas2018blackbox}, which applies  a projected version of ZO-signSGD using
 natural evolution strategy (NES) based random gradient estimator.
 In our experiments,  every method takes the same number of queries   per iteration. Accordingly, the total query complexity is consistent with the number of iterations.
{We refer to Appendix\,\ref{app: exp_para} for  details on experiment setups.}

\begin{figure*}[htb]
\centerline{
\begin{tabular}{cc}
% \includegraphics[width=.21\textwidth,height=!]{figures/id_11_ObjectValPlot_per_uncons.pdf}  & \hspace*{-0.3in}
% \includegraphics[width=.21\textwidth,height=!]{figures/id_11_DistortionPlot_per_uncons.pdf} 
% & \includegraphics[width=.21\textwidth,height=!]{figures/id_11_ObjectValPlot_per_cons.pdf}  & \hspace*{-0.2in}
%  \includegraphics[width=.21\textwidth,height=!]{figures/id_11_DistortionPlot_per_cons.pdf} \\
\begin{tabular}{cc}
 \includegraphics[width=.23\textwidth,height=!]{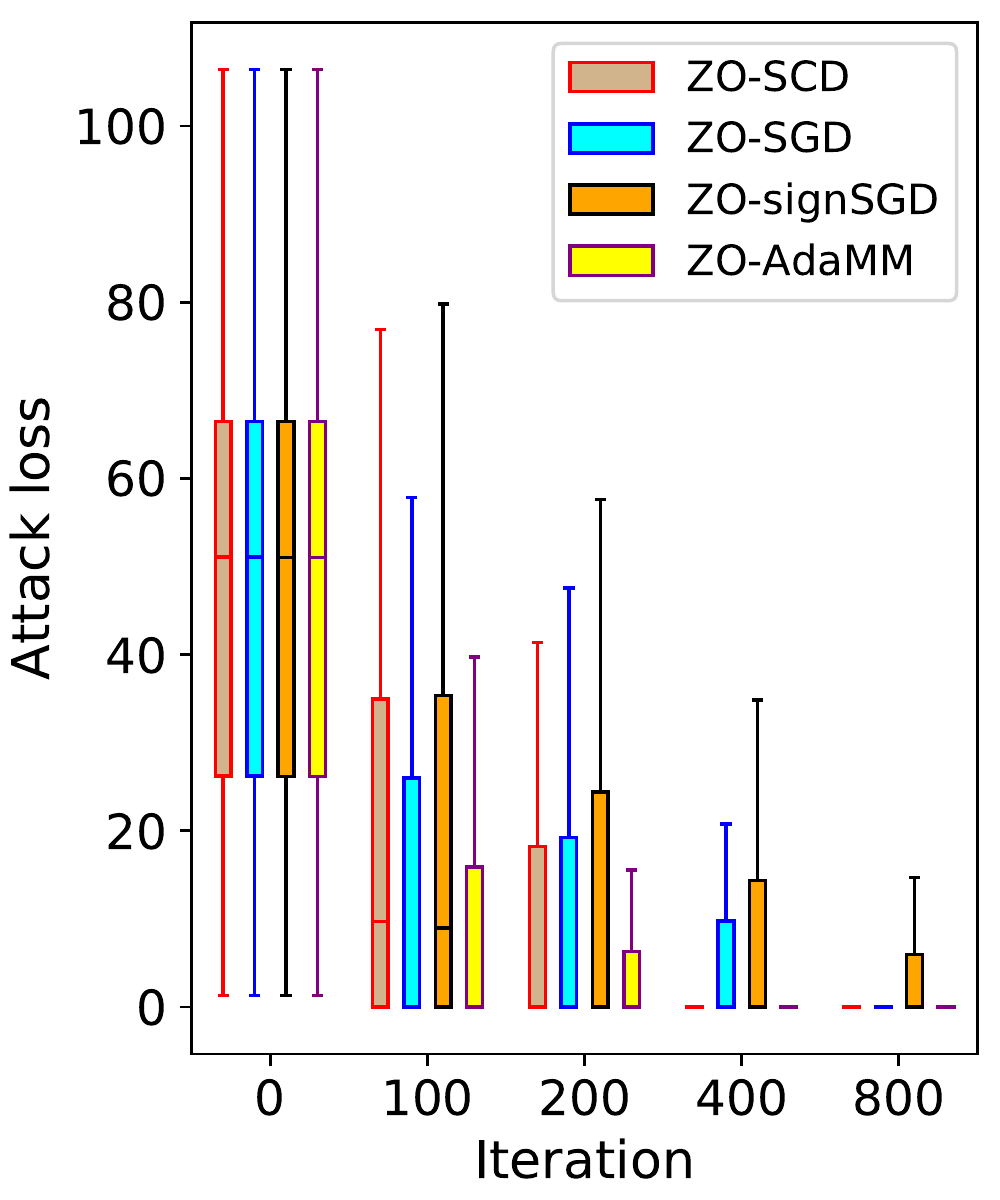}  & \hspace*{-0.15in}
\includegraphics[width=.23\textwidth,height=!]{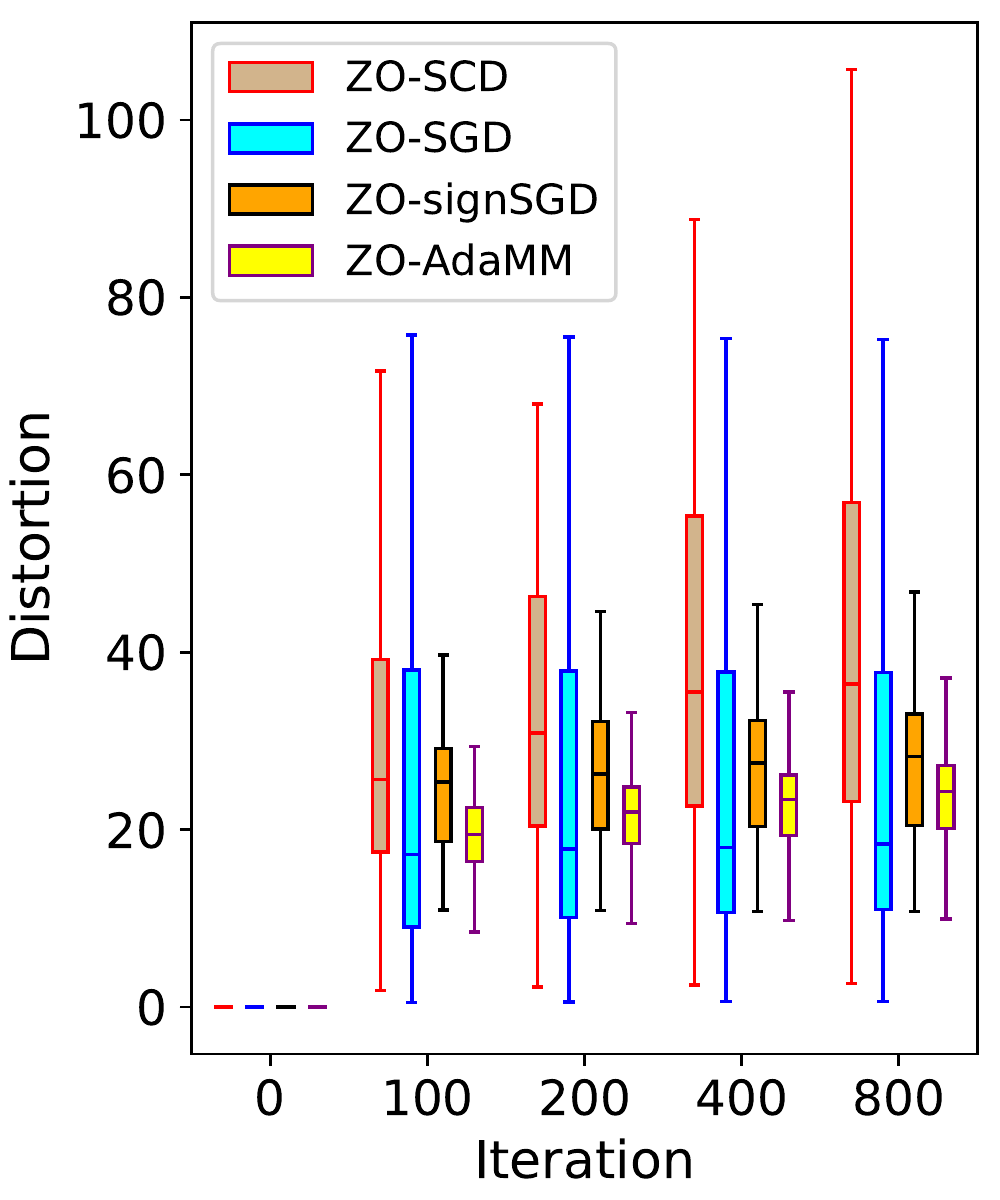} 
\end{tabular}
& 
\begin{tabular}{cc}
\includegraphics[width=.23\textwidth,height=!]{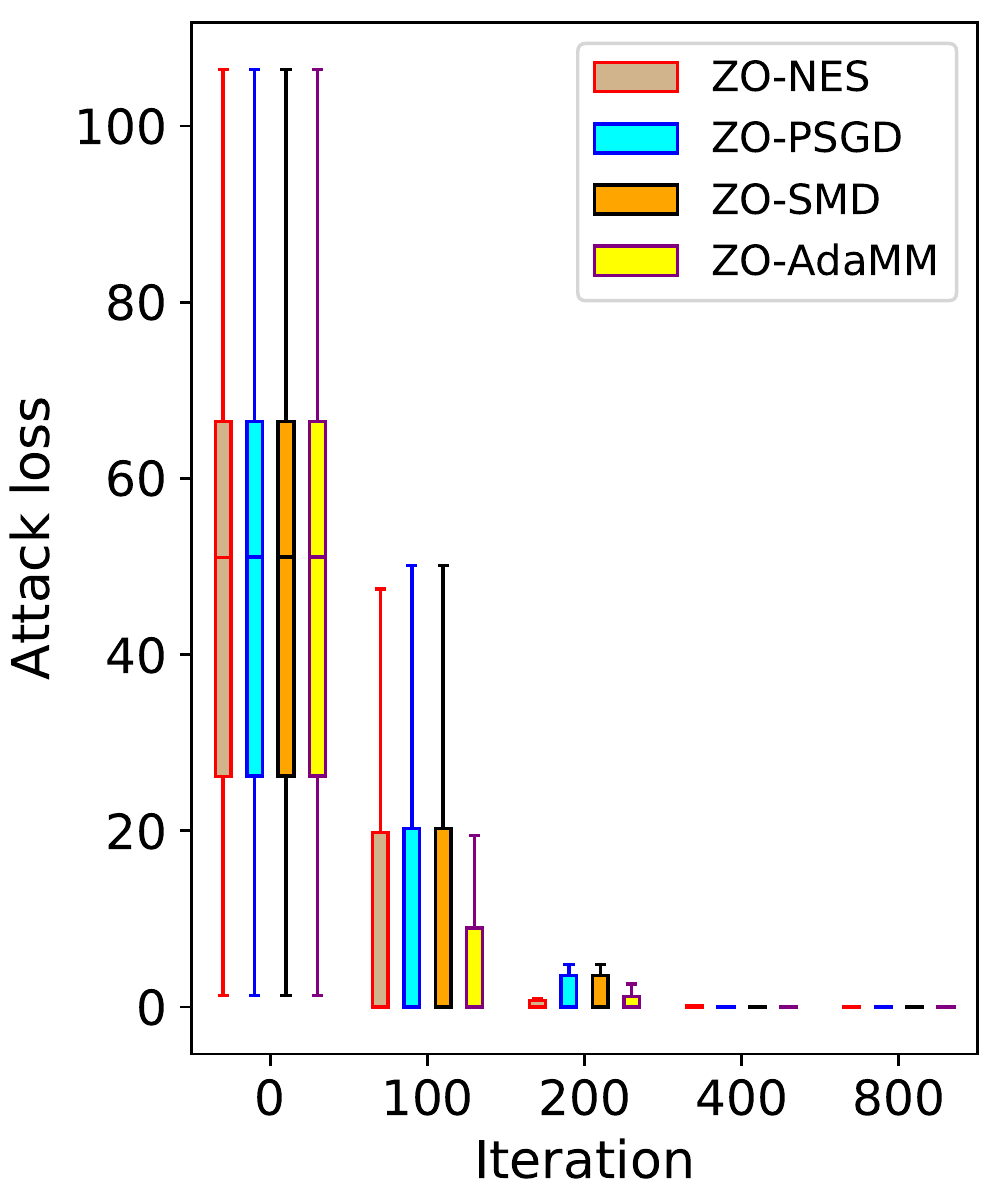}  & \hspace*{-0.15in}
 \includegraphics[width=.23\textwidth,height=!]{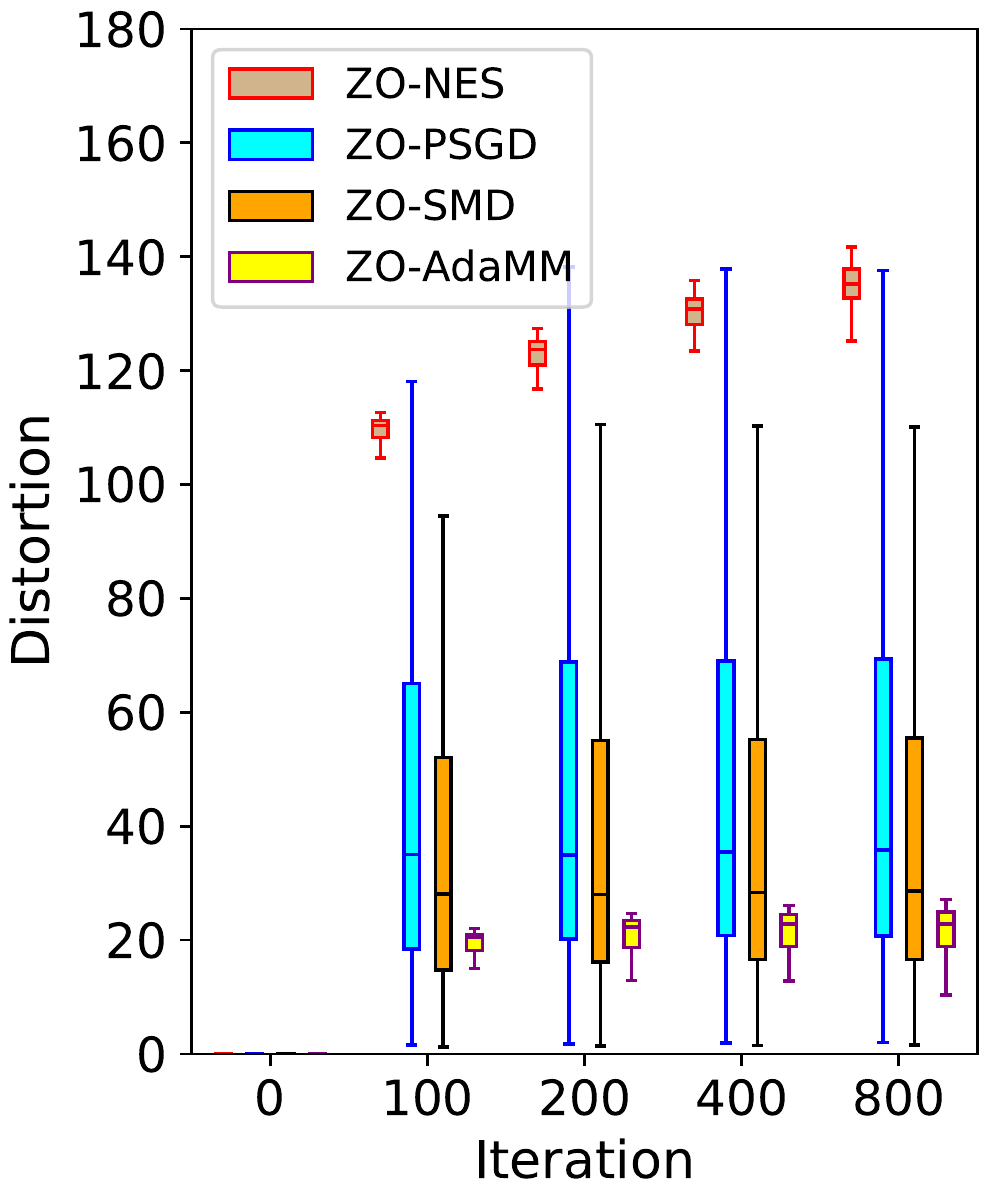}
\end{tabular}
 \\
%\\
%\includegraphics[width=.25\textwidth,height=!]{Plots_SL/id_11_LossPlot.pdf}\\
%{(a) unconstrained setting (objective value \& distortion)} 
(a) unconstrained setting
 &  %{(b) constrained setting (objective value \& distortion)} 
 (b) constrained setting
\end{tabular} \vspace*{-0.1in}
}
\caption{\footnotesize{ The
attack loss  and  adversarial   distortion   
%$\| \boldsymbol{\delta} \|_2^2$ 
v.s. iterations. Each box represents   results from $100$ images.
%to 
 %design black-box adversarial examples of   images with ID $502$ (left) and $836$ (right). 
% for  attacking 10 images with the true class label `brambling'.
}}\label{fig: loss_dist_per_uncons_cons}
%\vspace*{-0.1in}
\end{figure*}

\vspace*{-0.1in}
\paragraph{Per-image adversarial perturbation}
In Fig.\,\ref{fig: loss_dist_per_uncons_cons},
%-\ref{fig: loss_dist_single_image_supp2}, 
we present the attack  loss and the resulting $\ell_2$-distortion against  iteration numbers for solving both unconstrained and constrained adversarial attack problems, namely, \eqref{eq: attack_general_uncons} and \eqref{eq: attack_general} in Appendix\,\ref{app: exp_para}, over $100$ randomly selected images.
%(\textcolor{Sijia_color}{with the original image ID $6$ in ImageNet}).
Here every algorithm is initialized by zero   perturbation. Thus, as the iteration increases, the attack loss decreases until it converges to $0$ (indicating successful attack) while the distortion could increase. At this sense, the best attack performance   should correspond to the best tradeoff between the fast convergence to $0$ attack loss  and the low distortion power (evaluated by  $\ell_2$ norm).
As we can see, ZO-AdaMM consistently outperforms other ZO methods in terms of the fast convergence of attack loss and   relatively small   perturbation. We also note that 
ZO-signSGD and ZO-NES have poor convergence accuracy in terms of  either large attack loss or large distortion at final iterations. This is not surprising, since it has been shown in \cite{liu2018signsgd} that ZO-signSGD only  converges to a neighborhood of a solution, and ZO-NES   can be regarded as a Euclidean projection based ZO-signSGD, which could induce convergence issues shown by Prop.\,\ref{prop: imp_distance}. We  refer readers to Table \ref{table: per_image_distortion} for  detailed experiment results.

\vspace{-0.05in}

\paragraph{Universal adversarial perturbation}
% \vspace*{-0.05in}
% In this experiment, we solve the constrained problem
% \eqref{eq: attack_general}  for designing a universal adversarial perturbation $\boldsymbol{\delta}$, where we  
% attack $M = 10$ images with  the true class label `brambling' and we set $\lambda = 10$ in \eqref{eq: attack_general}.
% % In \eqref{eq: attack_general},
% % we set \textcolor{Sijia_color}{[$\lambda = 10$]}
% % and choose $M = 10$ images from ImageNet with the  class label `brambling', $T=20,000$.
% %\textcolor{Sijia_color}{[Please elaborate on ZO-AdaMM parameter setting? $T$? and others?]}
% % The other parameter settings are 
% % similar to  Sec.\,\ref{sec: per_image_perturbation} except a) 
% The setting of algorithmic parameters is similar to   Sec.\,\ref{sec: per_image_perturbation} except  $T = 20000$. For ZO-AdaMM, we choose 
% $\alpha = 0.002$, $\beta_1 = 0.9$, and $\beta_2 = 0.3$, where 
% the sensitivity  of exponential moving average parameters $(\beta_1, \beta_2)$ is shown in  Fig.\,\ref{fig: beta_per_uncons_cons_uni}-(c).
%  For the other ZO algorithms, we greedily search $\alpha$
%  over $[10^{-2}, 10^{-4}]$ and choose the value that achieves the best convergence accuracy as shown  in Table\,\ref{table: alpha_uni}.
We now focus on designing a universal adversarial perturbation using the constrained attack problem formulation. Here   we  
attack $M = 100$ random selected images from ImageNet.
%with  the true class label `brambling'.
In Fig.\,\ref{fig: loss_dist_universal_attack}, we present the attack loss as well as the $\ell_2$ norm of universal perturbation at different iteration numbers.   
%Fig.\,\ref{fig: loss_dist_universal_attack}-(a) shows that   
As we can see, 
compared with the other ZO algorithms, ZO-AdaMM  has the fastest convergence speed  to reach the smallest adversarial perturbation (namely, strongest universal attack). 
% Accordingly, %Fig.\,\ref{fig: loss_dist_universal_attack}-(b) shows that 
% ZO-AdaMM finds  the strongest black-box universal attack  in the sense that the generated adversarial example is the most similar to the original image. 
\textcolor{black}{Moreover, in Table\,\ref{table: universal_vs_different_lr} we present detailed attack success rate and    $\ell_2$ distortion  over $T = 40000$ iterations.  Consistent with Fig.\,\ref{fig: loss_dist_universal_attack}, ZO-AdaMM   achieves  highest success rate with lowest distortion. 
%which imply the more optimal solution that ZO-AdaMM can obtain than others.
In Fig.\,\ref{table: universal_pattern} of Appendix\,\ref{table: universal_pattern}, we  visualize   patterns of the generated universal adversarial perturbations which further confirm the advantage of ZO-AdaMM. % ($M = 10$ with  true class label `brambling'),
%the number of iterations and the  $\ell_2$ distortion achieved at the first successful attack by different ZO optimization methods. As we can see, ZO-AdaMM requires more iterations than ZO-PSGD and ZO-SMD to  find the first successful adversarial example, but once it succeeds, the  $\ell_2$ distortion is the smallest among all the methods.   It is not   surprising that the other  methods may require fewer iterations to obtain the first successful attack, since their updates   keep  large perturbations, which enable crafted examples  to   cross the   boundary of  correct predictions at earlier iterations. 
}

\vspace{-0.1in}
\begin{minipage}{.5\textwidth}
\begin{figure}[H]
\begin{center}
\begin{tabular}{cc}
\includegraphics[width=.45\textwidth,height=!]{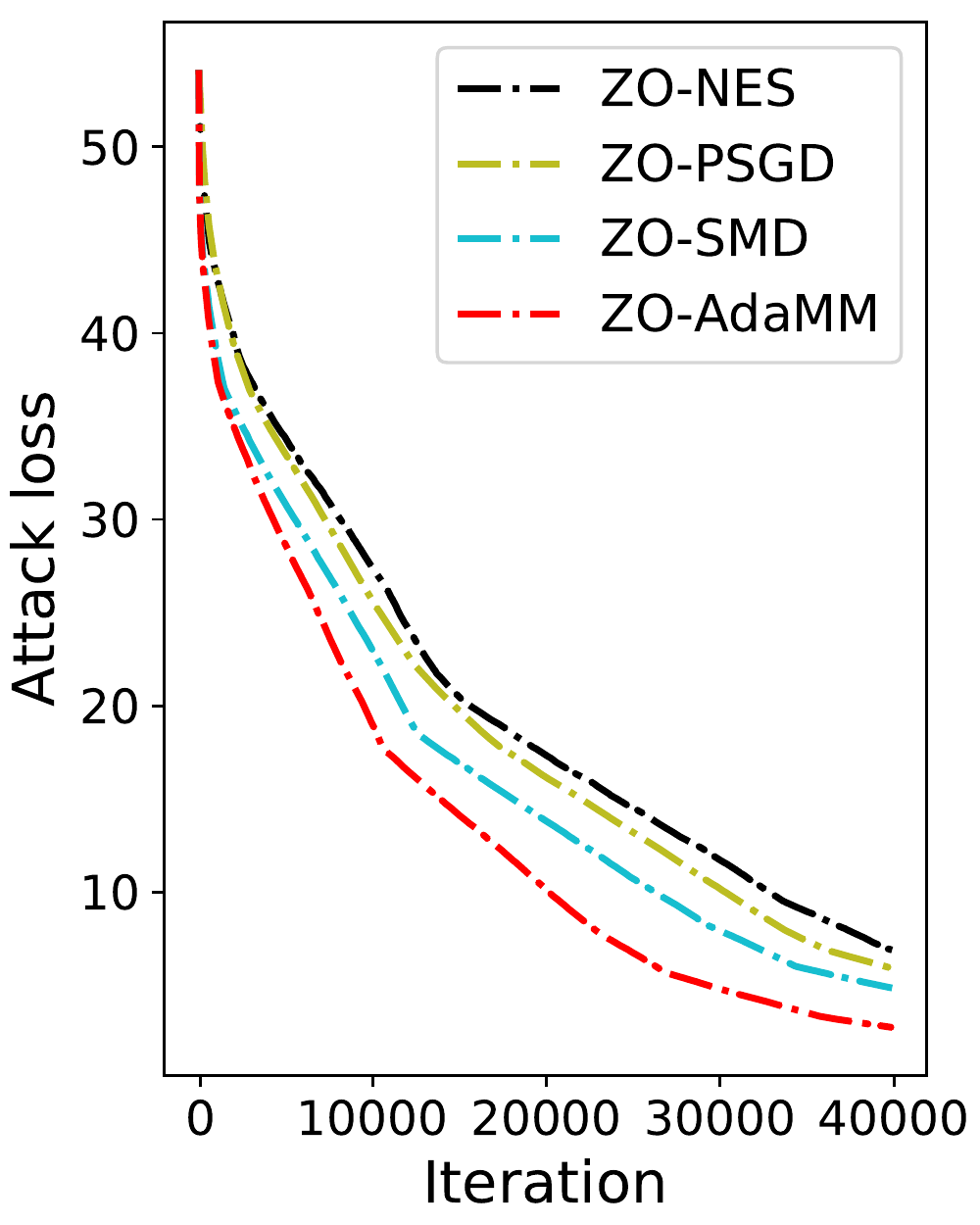}  & \hspace*{-0.15in}
\includegraphics[width=.45\textwidth,height=!]{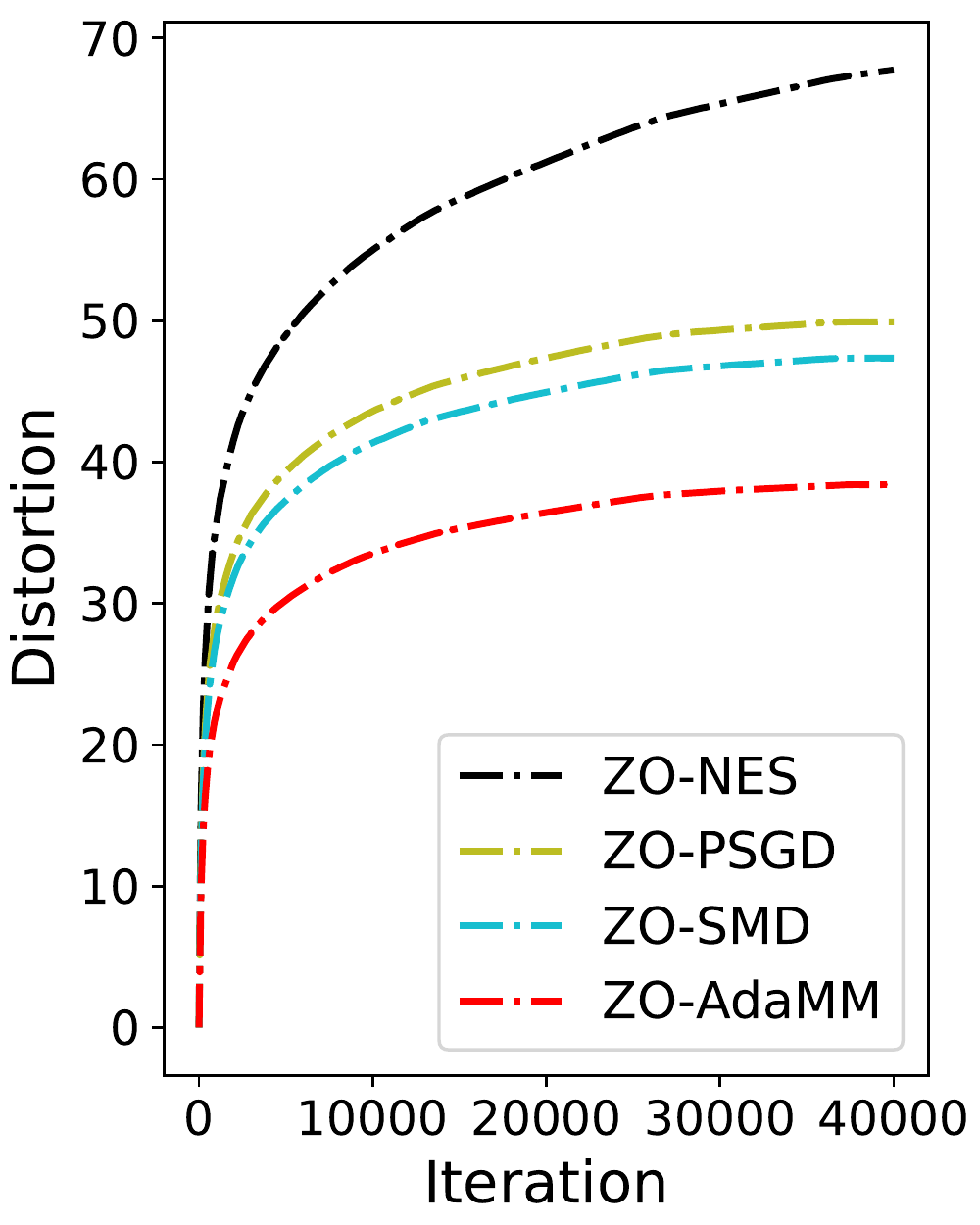} %\\
%\includegraphics[width=.25\textwidth,height=!]{Plots_SL/id_11_LossPlot.pdf}\\
%(a) \textcolor{Sijia_color}{Attack loss} & \hspace*{-0.0in} (b) distortion %$\| \boldsymbol{\delta}\|_2^2 $ % & (c) attack loss
\end{tabular}
\end{center}
\vspace*{-0.2in}
\caption{\footnotesize{Attack loss and   distortion of universal attack. 
}}
  \label{fig: loss_dist_universal_attack}
\end{figure}
\end{minipage}%
\hspace{0.2in}
\begin{minipage}{0.45\textwidth}
\begin{table}[H]
\begin{center}
\begin{adjustbox}{width= 0.9\textwidth }
\begin{tabular}{c|c|c}
\toprule[1pt]
          Methods             &
          \begin{tabular}[c]{@{}c@{}}  Attack  \\
success rate
\end{tabular}
          %Iteration number to achieve 1st successful attack
                        &        \begin{tabular}[c]{@{}c@{}}  Final \\ $\| \boldsymbol{\delta}_{T}\|_2^2 $
\end{tabular}   \\
                        \midrule[1pt]
\multirow{1}{*}{ZO-NES} % & $1\times10^{-2}$        & Fail  & NA                              \\
                        % & $1\times10^{-3}$         & 244.94/2531                   & 147.76           \\
                      % & $4\times10^{-4}$         & 65.38/8188                    & 56.99            \\
                        & {74\%}                         & {67.74}   \\
                      %  & $2\times10^{-5}$       & Fail   & NA                               \\
                      % \midrule[1pt]
                      \hline
                       
\multirow{1}{*}{ZO-PSGD} % & $1\times10^{-2}$         & Fail   & NA                                  \\
                        %& $1\times10^{-3}$         & 72.23/6107                    & 60.98         \\
                        %& $6\times10^{-4}$         & 38.86/13543                   & 36.86         \\
                        & 78\%                       & {49.92}\\
                        %& $4\times10^{-4}$         & Fail   & NA                       \\
                      % \midrule[1pt]
                        \hline
                        
\multirow{1}{*}{ZO-SMD} % & $1\times10^{-1}$         & Fail   & NA                                  \\
                        %& $4\times10^{-2}$         & 116.12/12370                   & 107.74   \\
                        %& $7\times10^{-3}$         & 68.36/15130                    & 65.64    \\
                        & 79\%                           & {47.36}\\
                        %& $5\times10^{-3}$         & Fail   & NA                       \\
%\midrule[1pt]
\hline 
\multirow{1}{*}{ZO-AdaMM}  
 & \textbf{84}\%                           & \textbf{38.40}\\ 
% & \begin{tabular}[c]{@{}c@{}} $ 2\times10^{-3} $ \\
% $(\beta_2 = 0.1)$
% \end{tabular} 
% %$2\times10^{-3} (\beta_1 = 0.9, \beta_2 = 0.1)$  
% & 19.39/12071   & 19.97   \\  
%                             &  % $2\times10^{-3} (\beta_1 = 0.9, \beta_2 = 0.2)$  
%                             \begin{tabular}[c]{@{}c@{}} $ 2\times10^{-3} $ \\
% $( \beta_2 = 0.2)$
% \end{tabular} 
%                             & 15.12/15921   & 14.62  \\ 
%                             & \begin{tabular}[c]{@{}c@{}} $ 2\times10^{-3} $ \\
% $( \beta_2 = 0.3)$
% \end{tabular}   
%                             %$2\times10^{-3} (\beta_1 = 0.9, \beta_2 = 0.3)$  
%                             &12.81/18192    & \textbf{12.63}    \\  
\bottomrule[1pt]
\end{tabular}
\end{adjustbox}
\end{center}
\caption{ \footnotesize{ 
% Performance of universal attack with different learning rate for ZOPSGD, ZOSMD, ZONES and ZOAdaMM. 
\textcolor{black}{Summary of attack success rate and eventual $\ell_2$ distortion for universal attack against $100$ images under  $T = 40000$ iterations.
%Summary of iterations required to achieve $1$st success universal attack and $\ell_2$ distortion of adversarial perturbation $\boldsymbol{\delta}_t$ versus iteration $t$ given a total number of  $T = 20000$ iterations. 
} } }
\label{table: universal_vs_different_lr}
\end{table}
\end{minipage}

\section{Conclusion}
%\vspace{-0.05in}

In this paper, 
we propose ZO-AdaMM, the first effort to integrate adaptive momentum methods with ZO optimization. In theory, we show that ZO-AdaMM has convergence  guarantees for both convex and nonconvex constrained optimization. Compared with (first-order) AdaMM, it suffers a slowdown factor of $O(\sqrt{d})$. % %, leading to $O(d/\sqrt{T})$  convergence rate. 
%guarantees for both convex and nonconvex optimization, and 
Particularly, we establish a new Mahalanobis distance based convergence measure whose necessity and importance are provided in characterizing    the convergence behavior of ZO-AdaMM on nonconvex constrained problems.
%\textcolor{XC_color}{which successfully characterizes the convergence behavior of ZO-AdaMM on nonconvex constrained problems. Further, we demonstrate the importance of Mahalanobis distance based projection by providing examples where the Euclidean distance based projection lead to non-convergence, which implies signSGD should be implemented with Mahalanobis projection in constrained problems. }
%[which circumvents the non-convergence issue of using the conventional Euclidean distance based convergence measure in the constrained nonconvex analysis.]
%\textcolor{XC_color}{[XC: It seems the non-convergence example is caused by replacing projection operator, it is not due to use of the Euclidean distance based convergence measure, I changed the sentence a little bit.]} 
To demonstrate the utility of the algorithm, we show the superior performance of ZO-AdaMM for designing   adversarial examples from black-box neural networks. %In both examples of per-image and universal adversarial perturbation, 
Compared with $6$ state-of-the-art ZO methods, ZO-AdaMM has the fastest empirical convergence  to  strong black-box adversarial attacks that require the minimum   distortion strength. 
{{
\bibliographystyle{IEEEbib}
\bibliography{Ref_ZO,refs2,Ref_Sijia,ref}
}}

\newpage 
\setcounter{section}{0}

\section*{Appendix}

\setcounter{section}{0}
\setcounter{figure}{0}
\makeatletter 
\renewcommand{\thefigure}{A\@arabic\c@figure}
\makeatother
\setcounter{table}{0}
\renewcommand{\thetable}{A\arabic{table}}
\setcounter{mylemma}{0}
\renewcommand{\themylemma}{A\arabic{mylemma}}

\section{Smoothing Function and Random Gradient Estimate}
\label{app: grad_properties}

\begin{mylemma} 
\label{lemma: smooth_f_random}
% The ZO gradient estimator $\hat{\nabla} f$ in \eqref{eq: grad_rand} and the smoothing function $f_{\mu}$ in \eqref{eq: fmu_smooth} have the following properties.
a) \textit{Relationship between $f_{\mu}$ and $f$:}
If $f$ is convex, then $f_\mu$ is   convex.
If $f$ is $L_{c}$-Lipschitz continuous, then $f_{\mu}$ is $L_{c}$-Lipschitz continuous.
Moreover for any $\mathbf x \in \mathbb R^d$,
\begin{align}
 | f_\mu(\mathbf x) - f(\mathbf x) | \leq
      L_c \mu.  \label{eq: dist_f_smooth_true_random_Lip}
\end{align}
If $f$ has $L_{g}$-Lipschitz continuous gradient, then $f_{\mu}$ has $L_{g}$-Lipschitz continuous gradient. Moreover
%b-1)
%b-2)
for any $\mathbf x \in \mathbb R^d$,
\begin{align}
& | f_\mu(\mathbf x) - f(\mathbf x) | \leq
      {L_g \mu^2}/{2}   
\label{eq: dist_f_smooth_true_random}  \\
& \| \nabla f_\mu (\mathbf x) - \nabla f(\mathbf x) \|_2^2 \leq 
{\mu^2 d^2 L_g^2}/{4}. 
 \label{eq: dist_smooth_true}  
\end{align}
b) \textit{Statistical properties of $\hat \nabla f$:}
For any $\mathbf x \in \mathbb R^d$,
{ \begin{align}
%\small
 &  \mathbb E_{\mathbf u} %_{ \{ \mathbf u_j \}}
    \left [
    \hat \nabla f(\mathbf x)
    \right ] 
%     =  \mathbb E_{\mathbf u \sim U_{\mathrm{B}}} \left  [
%   \frac{d}{\mu}  f(\mathbf x +\mu \mathbf u ) \mathbf u  
%     \right ]
    =   \nabla f_\mu (\mathbf x). \label{eq: smooth_est_grad}
    \end{align}% 
}If $f$ has $L_g$-Lipschitz continuous gradient, then
{ \begin{align}
   \mathbb E_{\mathbf u} \left [
\| 
\hat \nabla f(\mathbf x)
\|_2^2
\right ] \leq 
2 d \| \nabla f(\mathbf x) \|_2^2  + {\mu^2 L_g^2 d^2}/{2}.  %, \quad  \text{under \textbf{A2}}.
\label{eq: second_moment_grad_random}
\end{align}%
}
\end{mylemma}
\textbf{Proof:}
We refer   readers to  \cite[Lemma\,4.1]{gao2014information} 
%and \cite[Lemma\,8]{shamir2017optimal} 
for the detailed proof of a)-b) except the Lipschitz continuity of $f_\mu$ and \eqref{eq: dist_f_smooth_true_random_Lip}.
Suppose that $f$ is $L_c$-Lipschitz continuous, based on the definition of $f_\mu$ in \eqref{eq: fmu_smooth}, we obtain
\begin{align*}
 |f_\mu (\mathbf x) - f_\mu (\mathbf y)| 
\leq &\frac{1}{\alpha(d)}\int_{B} |f(\mathbf x + \mu \mathbf u) - f(\mathbf y + \mu \mathbf u)| d  \mathbf u %\nonumber \\
\leq  L_c \| \mathbf x - \mathbf y \|_2,
\end{align*}
where $\alpha(d)$ denotes the volume of the unit ball $B$ in $\mathbb R^d$. 

Moreover, we prove \eqref{eq: dist_f_smooth_true_random_Lip} as below.
\begin{align*}
| f_\mu(\mathbf x) - f(\mathbf x) | = \left| \frac{1}{\alpha(d)}\int_{B} f(\mathbf x + \mu \mathbf u) - f(\mathbf x) d  \mathbf u \right| \leq \frac{\mu L_c}{\alpha(d)}\int_{B} \| \mathbf u \|_2 d  \mathbf u = \frac{\mu L_c d}{d + 1} \leq \mu L_c,
\end{align*}
\textcolor{black}{where the first equality holds due to \eqref{eq: fmu_smooth}, Jensen's inequality and Lipschitz  continuity of $f$, and the last equality holds 
since $(1/\alpha(d)) \int_{B} \| \mathbf u \|_2^p d \mathbf u = \frac{n}{n+p}$ \cite[Lemma\,6.3.a]{gao2014information}.}
\hfill $\square$

In 
Lemma\,\ref{lemma: smooth_f_random}, it is clear from  \eqref{eq: dist_smooth_true} and \eqref{eq: smooth_est_grad}  that the ZO gradient estimate \eqref{eq: grad_rand} becomes unbiased to the true gradient $\nabla f$ only when $\mu \to 0$. However,   if $\mu$ is too small, then the difference of empirical function values is also too small to represent the function differential \cite{lian2016comprehensive,liu2018_NIPS}. Thus, the tolerance on the smoothing parameter $\mu$ is an important factor to indicate the convergence performance of ZO optimization methods. 
%see Table\,\ref{table: SZO_complexity_T}.
It is also known from \eqref{eq: second_moment_grad_random} that regardless of the value of $\mu$, the
variance of the ZO gradient estimate is always proportional to the dimension $d$. This is one of reasons for the dimension-dependent slowdown in convergence   of ZO optimization methods. {This also introduces technical difficulties for} {analyzing the effect of adaptive learning rate on the convergence of ZO-AdaMM in nonconvex optimization.}

\section{Proof for  Nonconvex Optimization}
\subsection{Proof of Proposition \ref{prop: imp_distance}}\label{app: prop_divergence}
%\textbf{Proof of Proposition \ref{prop: imp_distance}}:
Let us consider a special case of Algorithm\,\ref{alg:zoadam} with the average ZO gradient estimate $\hat{\nabla}f  (\mathbf x ) =
    \frac{d}{q\mu}\sum_{i=1}^q \left \{ [  f ( \mathbf x + \mu \mathbf u_i ) - f ( \mathbf x )]  \mathbf u_i \right \}$ under $\beta_{1,t} = \beta_2 \to 0$,  $\mu \to 0$ and $q \to \infty $. The conditions of $\beta_{1,t} = \beta_2 \to 0$ enables Algorithm\,\ref{alg:zoadam} to reduce to ZO-signSGD in \cite{liu2018signsgd}, and the conditions of  $\mu \to 0$ and $q \to \infty $ makes the ZO gradient estimate   unbiased to $\nabla f(\mathbf x)   $ and its variance close to $0$ \citep[Proposition\,2]{liu2018signsgd}. As a result, we obtain $\hat{\mathbf g}_t \to \nabla f(\mathbf x_t)$, and 
Algorithm\,1 becomes signSGD \cite{bernstein2018signsgd},
\begin{align}\label{eq: zo-nes}
   \mathbf  x_{t+1} = & \Pi_{\mathcal X, \mathbf I} (\mathbf x_t  - \alpha_t \mathrm{sign}(\nabla f(\mathbf x_t))) %\nonumber \\
    % = & \Pi_{\mathcal X,\mathbf I} (\mathbf x_t  - \alpha_t \hat V_t^{-1/2}\nabla f(x_t)))
\end{align}
where  
$\mathrm{sign}(x) = 1$ if $x > 0$ and $-1$ if $x < 0$, and it is taken elementwise for a vector argument. 
%where $\hat V_t = \mathrm{diag}(\nabla f(x_t)^2)$ and $\nabla f(x_t)^2$ is element-wise square.

% It is easy to see that if we set $\hat g_t = \nabla f(x_t)$ (first order noiseless optimization) and run ZO-NES on \eqref{eq: counter_example}, it is exactly ZO-AdaMM with $\beta_2 = 0$ and the distance metric in projection replaced by euclidean distance (the $\max$ operation in updating $\hat V_t$ is not active since gradient is a constant vector).

Let $f(\mathbf x) = -2 x_1 - x_2$ in \eqref{eq: counter_example}.
We then run \eqref{eq: zo-nes} at $x_1 = x_2 = 0.5$, which yields
% Now we analyze whether replacing the Mahalanobis distance with Euclidean distance in ZO-AdaMM will work. 
% Suppose we run ZO-NES any point $x_1 = x_2 = 0.5$ for \eqref{eq: counter_example}, we have
% \begin{align}
%     \mathrm{sign}(\nabla f(x_t))) = [-1,-1]^T
% \end{align}
% and 
\begin{align}\label{eq: zo-nes_specify}
   \mathbf  x_{t+1} =& \Pi_{\mathcal X} ([0.5,0.5]^T  - \alpha_t [-1,-1]^T) = \Pi_{X} ([0.5+\alpha_t,0.5 + \alpha_t]^T) = [0.5,0.5]^T,
\end{align}
where $\mathcal X$ encodes the constraint $| x_1 + x_2 | \leq 1$.

It is clear that the updating rule \eqref{eq: zo-nes_specify} will converge to $\mathbf x = [0.5,0.5]^T$ regardless of the choice of   $\alpha_t$. 
%this is a fix point of the algorithm ant it has 0 projected gradient.
The remaining question is whether or not  it is a stationary point.
Recall that a point $\mathbf x^*$ is a  stationary point if it satisfies the following conditions:
\begin{align}\label{eq: stationary_def}
\inp{\nabla f(\mathbf x^*)}{\mathbf   x - \mathbf  x^* }
 \geq 0, \, \forall \mathbf  x \in  \mathcal X. 
\end{align}
Since the gradient at $[0.5,0.5]^T$ is  $[-2,-1]^T$, and the inequality \eqref{eq: stationary_def} at $\mathbf x = [0.6,0.4]^T \in X$ does \textit{not} hold, given by $\inp{ [-2,-1]^T}{ [0.6,0.4]^T - [0.5,0.5]^T} = -0.1 <0 $. This implies that $\mathbf x^* = [0.5,0.5]^T$ is \textit{not} a stationary point of problem \eqref{eq: counter_example}.

Next, we apply the Mhalanobis distance $\hat{\mathbf V_t} = \diag(\nabla f(\mathbf x_t)^2)$ to \eqref{eq: zo-nes},
\begin{align}\label{eq: zo-nes_corrected}
    \mathbf x_{t+1} = & \Pi_{\mathcal X, \hat {\mathbf V}_t^{1/2}} (\mathbf x_t  - \alpha_t \mathrm{sign}(\nabla f(\mathbf x_t))) 
    = \Pi_{\mathcal X, \hat {\mathbf V}_t^{1/2}} (\mathbf x_t  - \alpha_t  \hat{\mathbf V}_t^{-1/2} \nabla f(\mathbf x_t) ) .
\end{align}
%where $\hat V_t = \mathrm{diag}(\nabla f(x_t)^2)$.
Similar to \eqref{eq: zo-nes},
we then consider the impact of fixed point  $\mathbf x_{t+1} = \mathbf x_{t}$ on \eqref{eq: zo-nes_corrected}.
By the definition of projection operator, we have
\begin{align} \label{eq: fix_point_project}
 \mathbf x_{t} = \argmin_{\mathbf x \in \mathcal X} \|\hat {\mathbf  V}^{1/4} (\mathbf x - \mathbf x_{t}  + \alpha_t \hat {\mathbf V}_t^{-1/2}\nabla f(\mathbf x_t))) \|   
\end{align}
The optimality condition of \eqref{eq: fix_point_project} is given by 
\begin{align}
   \inp{ \hat {\mathbf V}^{1/2} (\mathbf x_t - \mathbf x_t + \alpha_t \hat {\mathbf V}_t^{-1/2}\nabla f(\mathbf x_t))}{\mathbf x - \mathbf x_t }  \geq 0  ,\, \forall x \in X, \nonumber 
\end{align}
which reduces to 
\begin{align}
    \langle  \nabla f(\mathbf x_t), \mathbf  x - \mathbf x_t \rangle \geq 0 , \, \forall \mathbf x \in X.
\end{align}
It thus means that $\mathbf x_t$ is a stationary point by \eqref{eq: stationary_def}. 
% From the above argument, we can see that the Mahalanobis distance in projection is necessary to make the algorithm converge to a stationary point even in noiseless first order case, not to mention zeroth-order optimization.
\hfill $\square$

\subsection{Proof of Proposition \ref{prop: zo_adamm_nonconvex}}\label{app: prop_main_uncons}
Before proving the main result Proposition \ref{prop: zo_adamm_nonconvex}, we first prove a few auxiliary lemmas. 
% \textcolor{Sijia_color}{[SL: Lemma\,\ref{lem: z_t} should hold for any $\hat{\mathbf g}_t$, including the first-order case. Thus, we can directly cite from our ICLR work? Right? Any additional technical difficulty?]}\textcolor{XC_color}{[XC: Just changed $\beta_{1,t}$ to $\beta_1$ compared to our ICLR work, I commented the proof out for clarity.]}
\begin{lemma}\label{lem: z_t}
	%If $ \beta_{1,t+1}\leq \beta_{1,t} \leq \beta_1$, \
% 	Let $\mathbf{x}_0 \Def \mathbf{x}_1$ 
% 	{\color{red}[your algorithm descrption has no $x_0$]} \textcolor{XC_color}{[This is just a syntax sugar for simplicity of presentations in the following sections, $x_0$ is a conceptual quantity.]} in Algorithm\,1,
{Given $\{ \mathbf x_t\}$ from Algorithm\,\ref{alg:zoadam},}
	consider the sequence
	\begin{align}\label{eq:z}
	&\mathbf z_t =  \mathbf {x}_{t} + \frac{\beta_{1}}{1-\beta_{1}}(\mathbf x_t - \mathbf x_{t-1}), ~ \forall t \geq 1,
	\end{align}
	{where let $\mathbf{x}_0 \Def \mathbf{x}_1$.}
	Then for $\beta_{1,t}=\beta_1$ and $\mathcal{X}=\mathbb{R}^d$, $\forall t > 1$
	\begin{align*}
	&\mathbf z_{t+1}-\mathbf{z}_t \\
	=&  
	-  \frac{\beta_{1}}{1-\beta_{1}}\left({\alpha_t}\mathbf{\hat V}_t ^{-1/2}-{\alpha_{t-1}}  \mathbf{\hat V}_{t-1} ^{-1/2}\right)  \mathbf{m}_{t-1} - \alpha_t  \mathbf{\hat V}_t^{-1/2} \mathbf{\hat g}_t
	\end{align*}
	and 
	\begin{align*}
	\mathbf{z}_2 - \mathbf{z}_1 =  - \alpha_1   \mathbf{ \hat g}_1/\sqrt{\mathbf{\hat v}_1}.
	\end{align*}
\end{lemma}
%\textcolor{XL_color}{[XL: Is $v_t$ a scalar here or a vector as in Alg 1?]} \textcolor{XC_color}{[XC: $v_t$ is a vector, $\mathbf {\hat V}_t^{-1/2} \mathbf{m}_t$ is coordinate-wise division.]}

\noindent \textbf{Proof of Lemma \ref{lem: z_t}:} The proof follows from  Lemma 6.1 in \cite{chen2018convergence} by setting $\beta_{1,t} = \beta_1$.
\begin{lemma}\label{lem: telescope}
By ZO-AdaMM update rule, we have 
   \begin{align}\label{eq: descent_lemma_main}
    \mathbb E[f_{\mu}(\mathbf z_{t+1}) - f_{\mu}(\mathbf{z}_1)] 
    \leq &  \sum_{t=1}^TE\left[ \langle \nabla f_{\mu}(\mathbf x_t), \mathbf z_{t+1} - \mathbf z_t \rangle \right] +  \frac{4L_g + 5L_g\beta_1^2}{2(1-\beta_1)^2}\sum_{t=1}^T E\left[ \|\mathbf{x}_{t+1} - \mathbf {x}_{t}\|^2\right]
\end{align}

%\textcolor{XL_color}{[XL: Both AMSGrad and ZO-AdaMM are used. Should we unify the terminology?]} \textcolor{XC_color}{[XC: Changed all to ZO-AdaMM.]}

\end{lemma}
\noindent \textbf{Proof of Lemma \ref{lem: telescope}:}
By smoothness of function $f$, we can have
\begin{align} \label{eq: descent_lemma}
   & f_{\mu}(\mathbf z_{t+1}) - f_{\mu}(\mathbf z_t)\nonumber \\
    \leq &   \langle \nabla f_{\mu}(\mathbf z_t), \mathbf z_{t+1} - \mathbf z_t \rangle + \frac{L_g}{2} \| \mathbf z_{t+1} -\mathbf{z}_t\|^2 \nonumber\\
     = &\langle \nabla f_{\mu}(\mathbf x_t), \mathbf z_{t+1} - \mathbf z_t \rangle + \frac{L_g}{2} \| \mathbf z_{t+1}
     -\mathbf{z}_t\|^2  + \langle \nabla f_{\mu}(\mathbf z_t)- \nabla f_{\mu}(\mathbf x_t), \mathbf z_{t+1} - \mathbf z_t \rangle \nonumber\\
     \leq &\langle \nabla f_{\mu}(\mathbf x_t), \mathbf z_{t+1} - \mathbf z_t \rangle + \frac{L_g}{2} \| \mathbf z_{t+1}
     -\mathbf{z}_t\|^2  + \frac{1}{2}(\frac{1}{L_g}\| \nabla f_{\mu}(\mathbf z_t)- \nabla f_{\mu}(\mathbf x_t) \|^2 + L_g\|  \mathbf z_{t+1} - \mathbf z_t \|^2 )\nonumber \\
     \leq &  \langle \nabla f_{\mu}(\mathbf x_t), \mathbf z_{t+1} - \mathbf z_t \rangle + L_g \nonumber \| \mathbf z_{t+1}
     -\mathbf{z}_t\|^2 + \frac{1}{2}L_g\| \mathbf z_t - \mathbf x_t \|^2 \nonumber\\ 
     = &  \langle \nabla f_{\mu}(\mathbf x_t), \mathbf z_{t+1} - \mathbf z_t \rangle + L_g  \| \mathbf z_{t+1}
     -\mathbf{z}_t\|^2  + \frac{1}{2}L_g\|  \frac{\beta_1}{1-\beta_1}(\mathbf x_t - \mathbf x_{t-1}) \|^2
\end{align}
Further, by \eqref{eq:z}, we have
\begin{align}
    \mathbf z_{t+1}- \mathbf z_t =   \frac{1}{1-\beta_{1}}(\mathbf{x}_{t+1} - \mathbf {x}_{t})  + \frac{\beta_1}{1-\beta_1} (\mathbf x_t - \mathbf x_{t-1}) \nonumber
\end{align}
and thus
\begin{align} \label{eq: diff_sqr}
  \|\mathbf z_{t+1}- \mathbf z_t\|^2 
    \leq   \frac{2}{(1-\beta_{1})^2}\|\mathbf{x}_{t+1} - \mathbf {x}_{t}\|^2  + \frac{2\beta_1^2}{(1-\beta_1)^2} \|\mathbf x_t - \mathbf x_{t-1}\|^2
\end{align}

Substituting \eqref{eq: diff_sqr} into \eqref{eq: descent_lemma}, we get
\begin{align}
    & f_{\mu}(\mathbf z_{t+1}) - f_{\mu}(\mathbf z_t) \nonumber \\
    \leq&   \langle \nabla f_{\mu}(\mathbf x_t), \mathbf z_{t+1} - \mathbf z_t \rangle +  \frac{2L_g }{(1-\beta_{1})^2}\|\mathbf{x}_{t+1} - \mathbf {x}_{t}\|^2  + \frac{2\beta_1^2L_g }{(1-\beta_1)^2} \|\mathbf x_t - \mathbf x_{t-1}\|^2 \nonumber \\
     & + \frac{1}{2}L_g\frac{\beta_1^2}{(1-\beta_1)^2}\|  \mathbf x_t - \mathbf x_{t-1} \|^2 \ \nonumber \\
     = & \langle \nabla f_{\mu}(\mathbf x_t), \mathbf z_{t+1} - \mathbf z_t \rangle +   \frac{2L_g}{(1-\beta_{1})^2}\|\mathbf{x}_{t+1} - \mathbf {x}_{t}\|^2  + \frac{5L_g\beta_1^2}{2(1-\beta_1)^2} \|\mathbf x_t - \mathbf x_{t-1}\|^2  
\end{align}
Summing $t$ from $1$ to $T$ and take expectation, we get
\begin{align}
    &\mathbb E[f_{\mu}(\mathbf z_{t+1}) - f_{\mu}(\mathbf{z}_1)] \nonumber \\
    \leq & \mathbb  E\bigg[ \sum_{t=1}^T \bigg( \langle \nabla f_{\mu}(\mathbf x_t), \mathbf z_{t+1} - \mathbf z_t \rangle +   \frac{2L_g}{(1-\beta_{1})^2}\|\mathbf{x}_{t+1} - \mathbf {x}_{t}\|^2  + \frac{5L_g\beta_1^2}{2(1-\beta_1)^2} \|\mathbf x_t - \mathbf x_{t-1}\|^2  \bigg)\bigg] \nonumber \\
    \leq &  \sum_{t=1}^T \mathbb E\left[ \langle \nabla f_{\mu}(\mathbf x_t), \mathbf z_{t+1} - \mathbf z_t \rangle \right] +  \frac{4L_g + 5L_g\beta_1^2}{2(1-\beta_1)^2}\sum_{t=1}^T \mathbb E\left[ \|\mathbf{x}_{t+1} - \mathbf {x}_{t}\|^2\right] \nonumber
\end{align}
\hfill $\square$

\begin{lemma} \label{lem: first_order_split}
Assume $\|\hat{g}_t\|_\infty \leq G_{zo},\, \forall t \in [T]$ and $m_0=0$, By ZO-AdaMM update rule, we have
    \begin{align} \label{eq:descent_tele}
    \sum_{t=1}^T \mathbb E[\langle \nabla f_{\mu}(\mathbf x_t), \mathbf z_{t+1} - \mathbf z_t \rangle ]  
    \leq & \mathbb E \left[\left(\frac{\eta G_{zo}}{1-\beta_1} + \eta^2\right)  \sum_{i=1}^d \frac{\alpha_{1}}{\sqrt{\mathbf{\hat v}_{0,i}}}   \right] \nonumber\\
    &-\sum_{t=1}^T\mathbb E\left[\langle \nabla f_{\mu}(\mathbf x_t),  \alpha_t  \mathbf{\hat V}_t^{-1/2} \nabla f_{\mu}(\mathbf x_t) \rangle \right]. 
\end{align}
%where we assume {\color{red}[in where? Assumption A]}$\|\nabla f_{\mu}(\mathbf x_t)\|_\infty \leq \eta$ and $\|\hat{g}_t\|_\infty \leq G_{zo}$.
%\textcolor{Sijia_color}{[SL: Why not use $\ell_2$ norm?]}
\end{lemma}
\noindent \textbf{Proof of Lemma \ref{lem: first_order_split}:} 
By Lemma \ref{lem: z_t}, %\textcolor{Sijia_color}{[SL: which lemma?]}, 
we have
\begin{align} \label{eq: descent_bias}
&     \langle \nabla f_{\mu}(\mathbf x_t), \mathbf z_{t+1} - \mathbf z_t \rangle  \nonumber \\
    = & \langle \nabla f_{\mu}(\mathbf x_t), -  \frac{\beta_{1}}{1-\beta_{1}}\left({\alpha_t}\mathbf {\hat V}_t^{-1/2}-{\alpha_{t-1}}\mathbf {\hat V}_{t-1}^{-1/2}  \right)  \mathbf{m}_{t-1}  - \alpha_t \mathbf{\hat V}_t^{-1/2}\mathbf{\hat  g}_t \rangle \nonumber \\
    =& \langle \nabla f_{\mu}(\mathbf x_t), -  \frac{\beta_{1}}{1-\beta_{1}}\left({\alpha_t}\mathbf {\hat V}_t^{-1/2}-{\alpha_{t-1}}\mathbf {\hat V}_{t-1}^{-1/2}  \right)  \mathbf{m}_{t-1}\rangle - \langle \nabla f_{\mu}(\mathbf x_t),  \alpha_t  \mathbf {\hat V}_t^{-1/2} \mathbf{\hat g}_t \rangle ,
\end{align}
%\textcolor{Sijia_color}{[SL: is the last inequality redundant?]}\textcolor{XC_color}{[XC: yes, just removed.]}
and 
\begin{align} \label{eq:descent_split}
    &\langle \nabla f_{\mu}(\mathbf x_t),  \alpha_t \mathbf{\hat V}_t^{-1/2}\mathbf{\hat  g}_t \rangle \nonumber \\
    =& \langle \nabla f_{\mu}(\mathbf x_t),  \alpha_t \mathbf{\hat V}_t^{-1/2} \nabla f_{\mu}(\mathbf x_t) \rangle  + \langle \nabla f_{\mu}(\mathbf x_t),  \alpha_t \mathbf{\hat V}_t^{-1/2}   (\mathbf{\hat g}_t - \nabla f_{\mu}(\mathbf x_t)) \rangle \nonumber \\
     =& \langle \nabla f_{\mu}(\mathbf x_t),  \alpha_t \mathbf{\hat V}_t^{-1/2}   \nabla f_{\mu}(\mathbf x_t) \rangle  + \langle \nabla f_{\mu}(\mathbf x_t),  \alpha_{t-1} \mathbf{\hat V}_{t-1}^{-1/2}   (\mathbf{\hat g}_t - \nabla  f_{\mu}(\mathbf x_t))  \rangle \nonumber\\
     &+  \langle  \nabla f_{\mu}(\mathbf x_t),  \left({\alpha_{t}}\mathbf{\hat V}_t^{-1/2} - {\alpha_{t-1}}\mathbf{\hat V}_{t-1}^{-1/2}\right)   (\mathbf{\hat g}_t - \nabla  f_{\mu}(\mathbf x_t))  \rangle.
\end{align}

Substitute \eqref{eq:descent_split} into \eqref{eq: descent_bias}, we have 
\begin{align}
   & \langle \nabla f_{\mu}(\mathbf x_t), \mathbf z_{t+1} - \mathbf z_t \rangle \nonumber \\
    \leq & \langle \nabla f_{\mu}(\mathbf x_t), -  \frac{\beta_{1}}{1-\beta_{1}}\left({\alpha_t}\mathbf {\hat V}_t^{-1/2}-{\alpha_{t-1}}\mathbf {\hat V}_{t-1}^{-1/2}  \right) \odot \mathbf{m}_{t-1}\rangle  \nonumber \\
    &-\langle \nabla f_{\mu}(\mathbf x_t),  \alpha_t \mathbf{\hat V}_t^{-1/2}   \nabla f_{\mu}(\mathbf x_t) \rangle - \langle \nabla f_{\mu}(\mathbf x_t),  \alpha_{t-1}  \mathbf{\hat V}_{t-1}^{-1/2} (\mathbf{\hat g}_t - \nabla  f_{\mu}(\mathbf x_t))  \rangle \nonumber\\
     &-  \langle  \nabla f_{\mu}(\mathbf x_t),  \left({\alpha_{t}}\mathbf{\hat V}_t^{-1/2}-{\alpha_{t-1}}\mathbf{\hat V}_{t-1}^{-1/2}\right)    (\mathbf{\hat g}_t - \nabla  f_{\mu}(\mathbf x_t))  \rangle \nonumber \\
     = & \langle \nabla f_{\mu}(\mathbf x_t), -  \left({\alpha_t}\mathbf {\hat V}_t^{-1/2}-{\alpha_{t-1}}\mathbf {\hat V}_{t-1}^{-1/2}  \right)  \frac{ \mathbf m_{t}  }{1-\beta_1} \rangle  \nonumber \\ 
     & - \inp{ \nabla f_{\mu}(\mathbf x_t) }{ -  \left({\alpha_t}\mathbf {\hat V}_t^{-1/2}-{\alpha_{t-1}}\mathbf {\hat V}_{t-1}^{-1/2}  \right) \nabla  f_{\mu}(\mathbf x_t) } \nonumber \\
    &-\langle \nabla f_{\mu}(\mathbf x_t),  \alpha_t \mathbf{\hat V}_t^{-1/2}  \nabla f_{\mu}(\mathbf x_t)\rangle   - \langle \nabla f_{\mu}(\mathbf x_t),  \alpha_{t-1} \mathbf{\hat V}_{t-1}^{-1/2}  (\mathbf{\hat g}_t - \nabla  f_{\mu}(\mathbf x_t))  \rangle \nonumber \\
    \leq &(\frac{\eta G_{zo}}{1-\beta_1} + \eta^2) \sum_{i=1}^d \left | \frac{\alpha_{t-1}}{\sqrt{\mathbf{\hat v}_{t-1,i}}} -\frac{\alpha_{t}}{\sqrt{\mathbf{\hat v}_{t,i}}} \right|   \nonumber\\
    &-\langle \nabla f_{\mu}(\mathbf x_t),  \alpha_t \mathbf{\hat V}_t^{-1/2}   \nabla f_{\mu}(\mathbf x_t) \rangle  - \langle \nabla f_{\mu}(\mathbf x_t),  \alpha_{t-1} \mathbf{\hat V}_{t-1}^{-1/2}    (\mathbf{\hat g}_t - \nabla  f_{\mu}(\mathbf x_t)) \rangle
\end{align}
where the last inequality follows from the assumption that $\mathbf {\hat V}_{t} = \mathrm{diga}(\mathbf {\hat v}_t)$, $\|\nabla f_{\mu}(\mathbf x_t)\|_\infty \leq \eta$ and $\|\mathbf{\hat g}_t\|_\infty \leq G_{zo}$.

{The upper bound on $\|\mathbf{m}_t\|_{\infty}$ can be proved by a simple induction. Recall that $\mathbf{m}_t = \beta_{1,t} \mathbf{m}_{t-1} + (1-\beta_{1,t})\mathbf{\hat g}_t$, suppose $\|\mathbf{m}_{t-1}\| \leq G_{zo}$, we have
	\begin{align}
	\|\mathbf{m}_t\|_{\infty}
	\leq &(\beta_{1,t} + (1-\beta_{1,t}))\max (\|\mathbf{\hat g}_t\|_{\infty},\|\mathbf{m}_{t-1}\|_{\infty}) \nonumber \\
	=& \max(\|\mathbf{\hat g}_t\|_{\infty},\|\mathbf{m}_{t-1}\|_{\infty}) \leq G_{zo}. \label{eq: mt_H_bound}
	\end{align} Then since $\mathbf m_0=0$, we have $\|\mathbf m_0\| \leq G_{zo}$, which completes the induction.
}

% \textcolor{Sijia_color}{[SL: How to bound $\mathbf{m}_t$ is not clear in the last inequality. Also, we should have $\frac{\alpha_t}{\sqrt{\hat{v}_t}} \leq \frac{\alpha_{t-1}}{\sqrt{\hat{v}_{t-1}}}$ given $\alpha_t$ is non-increasing, right? 
% %Also, why not use Proposition\,\ref{lemma: Vt_mt} to handle the relationship $\hat{v}_t$ and $\mathbf{m}_t$ and $\mathbf{m}_{t-1}$?
% ]}\textcolor{XC_color}{[XC: The decreasing property is used later. We have $\|\mathbf{m}_t\|_\infty \leq G_{zo}$ when $\|\mathbf{\hat g}_t\|_\infty \leq G_{zo}$.]}\textcolor{XL_color}{[XL: It is more clear to provide a short discussion/proof how to get $\|\mathbf{m}_t\|_\infty \leq G_{zo}$ from $\|\mathbf{\hat g}_t\|_\infty \leq G_{zo}$].}\textcolor{XC_color}{[XC: Added a proof for bound on $\mathbf{m}_t$]}

Sum $t$ from 1 to $T$ and take expectation over randomness of $\mathbf{\hat g}_t$, we have
\begin{align} 
    &\sum_{t=1}^T \mathbb E[\langle \nabla f_{\mu}(\mathbf x_t), \mathbf z_{t+1} - \mathbf z_t \rangle ] \nonumber\\
    \leq & \mathbb E \left[\sum_{t=1}^T\left(\frac{\eta G_{zo}}{1-\beta_1} + \eta^2\right) \sum_{i=1}^d \left|\frac{\alpha_{t-1}}{\sqrt{\mathbf{\hat v}_{t-1,i}}} -\frac{\alpha_{t}}{\sqrt{\mathbf{\hat v}_{t,i}}} \right| \right]  \nonumber \\
    &-\sum_{t=1}^T\mathbb E \left[\langle \nabla f_{\mu}(\mathbf x_t),  \alpha_t \mathbf{\hat V}_t^{-1/2}   \nabla f_{\mu}(\mathbf x_t) \rangle \right]  - \sum_{t=1}^T\mathbb E \left[\langle \nabla f_{\mu}(\mathbf x_t),  \alpha_{t-1}  \mathbf{\hat V}_{t-1}^{-1/2} (\mathbf{\hat g}_t - \nabla  f_{\mu}(\mathbf x_t))  \rangle \right]\nonumber \\
    \leq & \mathbb E \left[\left(\frac{\eta G_{zo}}{1-\beta_1} + \eta^2\right)  \sum_{i=1}^d \frac{\alpha_{1}}{\sqrt{\mathbf{\hat v}_{0,i}}}   \right]  -\sum_{t=1}^T\mathbb E \left[\langle \nabla f_{\mu}(\mathbf x_t),  \alpha_t \mathbf{\hat V}_t^{-1/2}  \nabla f_{\mu}(\mathbf x_t) \rangle \right] \nonumber
\end{align}
where the last inequality follows from following facts.

1. Since $\mathbf{\hat v}_{t} = \max(\mathbf{\hat v}_{t-1}, \mathbf{v}_t)$, we know $\mathbf{\hat v}_t$ is non-decreasing. Given the fact that $\alpha_t$ is non-increasing (by our choice), we have $\alpha_{t-1}/\mathbf{\hat v}_{t-1,i} - \alpha_t/\mathbf{\hat v}_{t,i} \geq 0 $. Thus, following inequality holds.
\begin{align}
&\mathbb E \left[\sum_{t=1}^T  \sum_{i=1}^d \left |\frac{\alpha_{t-1}}{\sqrt{\mathbf{\hat v}_{t-1,i}}} -\frac{\alpha_{t}}{\sqrt{\mathbf{\hat v}_{t,i}}}  \right| \right]= \mathbb E \left[ \sum_{i=1}^d \sum_{t=1}^T \left| \frac{\alpha_{t-1}}{\sqrt{\mathbf{\hat v}_{t-1,i}}} -\frac{\alpha_{t}}{\sqrt{\mathbf{\hat v}_{t,i}}} \right|   \right]\nonumber \\
=& \mathbb E \left[ \sum_{i=1}^d \sum_{t=1}^T  \left(\frac{\alpha_{t-1}}{\sqrt{\mathbf{\hat v}_{t-1,i}}} -\frac{\alpha_{t}}{\sqrt{\mathbf{\hat v}_{t,i}}} \right)    \right]  
\leq  \mathbb E \left[  \sum_{i=1}^d\frac{\alpha_{1}}{\sqrt{\mathbf{\hat v}_{0,i}}} \right] 
\end{align}
2.   We have $\mathbb E[\mathbf{\hat g}_t|\mathbf{\hat g}_{1:t-1}] = \nabla f_{\mu}(\mathbf x_t)$ by the assumption that $\mathbb E[\mathbf{\hat g}_t] = \nabla f_{\mu}(\mathbf x_t)$ and the noise on $\mathbf{\hat g}_t$ is independent of $\mathbf{\hat g}_{1:t-1}$. Thus, the following holds
\begin{align}\label{eq: unbiased_gradient}
    &\mathbb E \left[\langle \nabla f_{\mu}(\mathbf x_t),  \alpha_{t-1} \mathbf{\hat V}_{t-1}^{-1/2}  (\mathbf{\hat g}_t - \nabla  f_{\mu}(\mathbf x_t))  \rangle \right] % \nonumber \\
    % = & \mathbb E_{\mathbf{\hat g}_{1:t-1}}\left[E_{\mathbf{\hat g}_t}\left[\langle \nabla f_{\mu}(\mathbf x_t),  \alpha_{t-1} \mathbf{\hat V}_{t-1}^{-1/2}  (\mathbf{\hat g}_t - \nabla  f_{\mu}(\mathbf x_t))  \rangle \bigg| \mathbf{\hat g}_{1:t-1} \right]\right]\nonumber \\
    %= & \mathbb E_{\mathbf{\hat g}_{1:t}}\left[0 \right] 
    =0 
\end{align}

% \textcolor{Sijia_color}{SL: Is $\mathbb E[g_t|g_{1:t-1}] = \nabla f_{\mu}(\mathbf x_t)$ correct? Here it should be $\hat{g}_t$ \eqref{eq: grad_rand} rather than the stochastic gradient estimate. The aforementioned proof is correct if $f$ is replaced with $f_{\mu}$ \eqref{eq: fmu_smooth}. }\textcolor{XC_color}{[XC: Yes, we are proving first order AMSGrad here, the $f$ is actually $f_{\mu}$ if adapted to zeroth-order version.]}\textcolor{red}{[XC: Changed all previous $f$ to $f_\mu$ and $g$ to $\mathbf{\hat g}$ in new update]}\textcolor{XL_color}{[XL: The last inequality is still not clear to me. Why do $\alpha_{t-1}/\mathbf{\hat v}_{t-1} - \alpha_t/\mathbf{\hat v}_t \geq 0 $ and $\mathbb E[g_t|g_{1:t-1}] = \nabla f_{\mu}(\mathbf x_t)$ hold need some more explain/reference.]}\textcolor{XC_color}{[XC: Added a discussion on these.]}

\hfill $\square$
\begin{lemma}\label{lem: quadractic_bound}
    Assume $\gamma \Def \beta_1/\beta_2 < 1$, ZO-AdaMM yields %{\color{red}[any assumptions? $\beta_1/\beta_2<1$?]}
    \begin{align} \label{eq: sqr_bound}
    \|\mathbf{x}_{t+1} - \mathbf {x}_{t}\|^2 
    \leq & \alpha_t^2 d \frac{1-\beta_1}{1-\beta_2}\frac{1}{1-\gamma}
\end{align}
\end{lemma}
%Now let's consider $\|\mathbf{x}_{t+1} - \mathbf {x}_{t}\|^2$.
\noindent \textbf{Comment:}{This is an important lemma for ZO-AdaMM, it shows the squared update quantity is not dependent on size of stochastic gradient, thus giving a tighter dependency on $d$ compared with \citep{reddi2018convergence}.}

\noindent \textbf{Proof of Lemma \ref{lem: quadractic_bound}:} By the update rule, we have
\begin{align} 
    & \|\mathbf{x}_{t+1} - \mathbf {x}_{t}\|^2 
    =\alpha_t^2 \left\| \frac{\mathbf{m}_t}{\sqrt{\mathbf{\hat v}_t}}\right\|^2\nonumber \\
    \leq& \alpha_t^2 \sum_{i=1}^d \frac{((1-\beta_1)\sum_{j=0}^{t-1} \beta_1^{t-j}\mathbf{\hat g}_{j,i})^2}{(1-\beta_2) \sum_{j=0}^{t-1} \beta_2^{t-j} \mathbf{\hat g}_{j,i}^2}
    \leq  \alpha_t^2 \sum_{i=1}^d \frac{(1-\beta_1)^2(\sum_{j=0}^{t-1}\beta_1^{t-j})(\sum_{j=0}^{t-1} \beta_1^{t-j}\mathbf{\hat g}_{j,i}^2)}{(1-\beta_2) \sum_{j=0}^{t-1} \beta_2^{t-j} \mathbf{\hat g}_{j,i}^2} \nonumber\\
    \leq & \alpha_t^2 \sum_{i=1}^d \frac{(1-\beta_1)\sum_{j=0}^{t-1} \beta_1^{t-j}\mathbf{\hat g}_{j,i}^2}{(1-\beta_2) \sum_{j=0}^{t-1} \beta_2^{t-j} \mathbf{\hat g}_{j,i}^2} 
    \leq  \alpha_t^2 \sum_{i=1}^d \sum_{j=0}^{t-1} \frac{(1-\beta_1) \beta_1^{t-j}\mathbf{\hat g}_{j,i}^2}{(1-\beta_2)  \beta_2^{t-j} \mathbf{\hat g}_{j,i}^2} \nonumber\\
    \leq & \alpha_t^2 d  \sum_{j=0}^{t-1} \frac{1-\beta_1}{1-\beta_2}  \gamma^{t-j} 
    \leq \alpha_t^2 d \frac{1-\beta_1}{1-\beta_2}\frac{1}{1-\gamma} \nonumber
\end{align}
where the second inequality is due to Cauchy-Schwarz and $\gamma = \beta_1/\beta_2 < 1$.

% \textcolor{Sijia_color}{Why not use the conclusion in Proposition\,\ref{lemma: Vt_mt}? It does not hold in nonconvex setting?}
\hfill $\square$

%\textcolor{XC_color}{[XC: will change the convergence measure and reorganize the theorems on Monday]}
%\begin{theorem}
 %   Pick $R$ uniformly randomly from 1 to $T$ and set $\alpha_t = \alpha = 1/\sqrt{Td}$, AMSGrad yields
 %   \begin{align} \label{eq: gradient_bound}
%    \mathbb E \left[ \| \nabla f_{\mu}(x_R)\|^2 \right] 
 %   \leq & \frac{\sqrt{d}}{\sqrt{T}}D_fG _{zo} + \frac{G_{zo}}{T}\mathbb E \left[\left(\frac{\eta G_{zo}}{1-\beta_1} + \eta^2\right)  \left\| \frac{1}{\sqrt{\mathbf{\hat v}_{1}}} \right\|_1  \right]  \nonumber\\
 %   &+ \frac{\sqrt{d}}{\sqrt{T}} G_{zo} \frac{4L +  5L\beta_1^2}{2(1-\beta_1)^2} \frac{1-\beta_1}{1-\beta_2}\frac{1}{1-\gamma} 
%\end{align}
%\end{theorem}
\noindent \textbf{Proof of Proposition \ref{prop: zo_adamm_nonconvex}:}
Substitute \eqref{eq: sqr_bound} and \eqref{eq:descent_tele} into \eqref{eq: descent_lemma_main}, we get
\begin{align}
    &\mathbb E[f_{\mu}(\mathbf z_{t+1}) - f_{\mu}(\mathbf{z}_1)] \nonumber \\
    \leq &  \sum_{t=1}^T \mathbb E[ \langle \nabla f_{\mu}(\mathbf x_t), \mathbf z_{t+1} - \mathbf z_t \rangle]   + \frac{4L_g + 5L_g\beta_1^2}{2(1-\beta_1)^2}\sum_{t=1}^T \mathbb E[ \|\mathbf{x}_{t+1} - \mathbf {x}_{t}\|^2] \nonumber \\
    \leq & \mathbb E \left[\left(\frac{\eta G_{zo}}{1-\beta_1} + \eta^2\right)  \left\| \frac{\alpha_{1}}{\sqrt{\mathbf{\hat v}_{0}}} \right\|_1  \right]  -\sum_{t=1}^T\mathbb E \left[\langle \nabla f_{\mu}(\mathbf x_t),  \alpha_t  \mathbf{\hat V}_t^{-1/2} \nabla f_{\mu}(\mathbf x_t) \rangle \right] \nonumber \\
    & + \sum_{t=1}^T\alpha_t^2 d\frac{4L_g +  5L_g\beta_1^2}{2(1-\beta_1)^2} \frac{1-\beta_1}{1-\beta_2}\frac{1}{1-\gamma}
\end{align}
Rearrange and assume $f_{\mu}(\mathbf{z}_1) - \min_{\mathbf z} f_{\mu}(\mathbf z) \leq D_f$, we get %\textcolor{XC_color}{[XC: The extra dependency on $d$ in ZO-Admm comes from $G$ in LHS of \eqref{eq: extra_d} (it becomes $G_{zo}$ in Theorem \ref{thm: zo_adamm_nonconvex} which is $O(d)$, we may improve the bound below using the fact that $G_{zo}$ is $O(\sqrt{d\log dT})$ with high probability in spherical distribution). Is there other possible improvement?]} \textcolor{Sijia_color}{[SL: Great for the current improvement.]}
\begin{align}\label{eq: tele_uncon}
 & \sum_{t=1}^T\mathbb E \left[\langle \nabla f_{\mu}(\mathbf x_t),  \alpha_t  \mathbf{\hat V}_t^{-1/2} \nabla f_{\mu}(\mathbf x_t) \rangle \right]   \nonumber\\
    \leq & D_f + \mathbb E \left[\left(\frac{\eta G_{zo}}{1-\beta_1} + \eta^2\right)  \left\| \frac{\alpha_{1}}{\sqrt{\mathbf{\hat v}_{0}}} \right\|_1  \right]  + \sum_{t=1}^T\alpha_t^2 d\frac{4L_g +  5L_g\beta_1^2}{2(1-\beta_1)^2} \frac{1-\beta_1}{1-\beta_2}\frac{1}{1-\gamma}
\end{align}
%\begin{align}\label{eq: extra_d}
%    &\sum_{t=1}^T \alpha_t \mathbb E \left[ \| \mathbf{\hat V}^{-1/2} \nabla f_{\mu}(\mathbf x_t)\|^2 \right]  \nonumber \\
%    \leq & \sum_{t=1}^T\mathbb E \left[\langle \nabla %f_{\mu}(\mathbf x_t),  \alpha_t   \nabla f_{\mu}(\mathbf x_t)/\sqrt{\mathbf{\hat v}_t} \rangle \right]   \nonumber\\
%    \leq & D_f + \mathbb E \left[\left(\frac{L_c G_{zo}}{1-\beta_1} + \eta^2\right)  \left\| \frac{\alpha_{1}}{\sqrt{\mathbf{\hat v}_{1}}} \right\|_1  \right] \nonumber \\
%    &+ \sum_{t=1}^T\alpha_t^2 d\frac{4L +  5L\beta_1^2}{2(1-\beta_1)^2} \frac{1-\beta_1}{1-\beta_2}\frac{1}{1-\gamma}
%\end{align}
Set $\alpha_t = \alpha = 1/\sqrt{Td}$ and divide both sides by $T\alpha$, uniformly randomly pick $R$ from 1 to $T$,
\begin{align}\label{eq: rate_first_order}
    &\mathbb E \left[ \| \mathbf{\hat V}_R^{-1/2} \nabla f_{\mu}(x_R)\|^2 \right] 
    =  \frac{1}{T} \sum_{t=1}^T  \mathbb E \left[ \| V_t^{-1/2} \nabla f_{\mu}(\mathbf x_t)\|^2 \right] \nonumber\\
    \leq & \frac{D_f}{T\alpha} + \frac{1}{T}\mathbb E \left[\left(\frac{
    \eta G_{zo}}{1-\beta_1} + \eta^2\right)  \left\| \frac{1}{\sqrt{\mathbf{\hat v}_{0}}} \right\|_1  \right]  + {\alpha} d \frac{4L_g +  5L_g\beta_1^2}{2(1-\beta_1)^2} \frac{1-\beta_1}{1-\beta_2}\frac{1}{1-\gamma} \nonumber\\
    = & \frac{\sqrt{d}}{\sqrt{T}}D_f + \frac{1}{T}\mathbb E \left[\left(\frac{\eta G_{zo} }{1-\beta_1} + \eta^2\right)  \left\| \frac{1}{\sqrt{\mathbf{\hat v}_{0}}} \right\|_1  \right]   + \frac{\sqrt{d}}{\sqrt{T}}  \frac{4L_g +  5L_g\beta_1^2}{2(1-\beta_1)^2} \frac{1-\beta_1}{1-\beta_2}\frac{1}{1-\gamma} 
\end{align}
%\begin{align}
 %   &\mathbb E \left[ \| \nabla f_{\mu}(x_R)\|^2 \right] \nonumber \\
 %   =& \sum_{t=1}^T \frac{1}{T} \mathbb E \left[ \| \nabla f_{\mu}(\mathbf x_t)\|^2 \right] \nonumber\\
 %   \leq & \frac{D_fG}{T\alpha} + \frac{G}{T}\mathbb E \left[\left(\frac{\eta G_{zo}}{1-\beta_1} + \eta^2\right)  \left\| \frac{1}{\sqrt{\mathbf{\hat v}_{1}}} \right\|_1  \right]  \nonumber\\
 %   &+ {\alpha} dG \frac{4L +  5L\beta_1^2}{2(1-\beta_1)^2} \frac{1-\beta_1}{1-\beta_2}\frac{1}{1-\gamma} \nonumber\\
 %   = & \frac{\sqrt{d}}{\sqrt{T}}D_fG _{zo} + \frac{G}{T}\mathbb E \left[\left(\frac{\eta G_{zo}}{1-\beta_1} + \eta^2\right)  \left\| \frac{1}{\sqrt{\mathbf{\hat v}_{1}}} \right\|_1  \right]  \nonumber\\
%    &+ \frac{\sqrt{d}}{\sqrt{T}} G \frac{4L +  5L\beta_1^2}{2(1-\beta_1)^2} \frac{1-\beta_1}{1-\beta_2}\frac{1}{1-\gamma} 
%\end{align}

%\noindent \textbf{Proof of Theorem \ref{thm: zo_adamm_nonconvex}:}

Since $\mathbf{\hat V}_{0,ii}^{1/2} \geq c, \, \forall i \in [d]$. By Lemma \ref{lemma: smooth_f_random}, we have 
\begin{align}
    \|\mathbf{\hat V}_t^{-1/4} (\nabla f_{\mu}(x)-\nabla f_{\mu}(x))\|^2 \leq \frac{\mu ^2 d^2 L^2}{4c}
\end{align}

Then  we can easily adapt \eqref{eq: rate_first_order} to
\begin{align} 
     \mathbb E \left[ \| \mathbf{\hat V}_t ^{-1/4}\nabla f(x_R)\|^2 \right]  
    \leq  & \frac{\mu ^2 d^2 L^2}{2c} + 2\frac{\sqrt{d}}{\sqrt{T}}D_f  + 2\frac{1}{T}\mathbb E \left[\left(\frac{\eta G _{zo} }{1-\beta_1} + 2\eta2\right)  \left\| \frac{1}{\sqrt{\mathbf{\hat v}_{0}}} \right\|_1  \right]  \nonumber\\
    &+ \frac{\sqrt{d}}{\sqrt{T}}  \frac{4L_g +  5L_g\beta_1^2}{(1-\beta_1)^2} \frac{1-\beta_1}{1-\beta_2}\frac{1}{1-\gamma} \nonumber
\end{align}
Substituting into $\mu$ finishes the proof. \hfill $\square$

\subsection{Proof of Proposition \ref{prop: g_zo_bound}} \label{app: prop_concentration}
%\textbf{Proof of Proposition \ref{prop: g_zo_bound}}:
%\noindent\textbf{Proof of Proposition \ref{prop: g_zo_bound}:} %The proof is essentially bounding $G_{zo}$ by $O(\sqrt{d\log(dT)})$ with high probability. 
Upon defining $G_{\mathrm{zo},i} \Def \max_{t \in [T] } |\hat{g}_{t,i}|$
%Define $G_{zo,i} \Def \max_{t=1 }^T |\hat{g}_{t,i}|$ to be,

by  \citep[Lemma\,2.2]{Dasgupta2003elementary}, for a vector $\mathbf u $ sampled from a unit sphere in $\mathbb{R}^d$, we have for any $i \in [d]$,
%\textcolor{XL_color}{[XL: I did not find Lemma 5.1 in the ref. Maybe another one?]} \textcolor{XC_color}{[XC: It is Lemma 2.2 with $k=1$, just corrected.]}
%\textcolor{Sijia_color}{[SL: confused notation in $|\cdot|$. What is it? $\ell_1$ or $\ell_\infty$ or $\ell_2$? And for what $\xi$? Any condition on $\xi$?]}
\begin{align} \label{eq: prob_basic1}
    P[|  u_i | \geq \sqrt{\xi/d} ] \leq \exp\left(\left(1-\xi + \log \xi\right)/2\right).
\end{align}
Let $\xi = 4 \log \frac{dT}{\delta}$, and by the assumption of $\max(d,T) \geq 3$ we have $ 1 +  \log \xi \leq \xi/2 $.
Thus, we obtain from \eqref{eq: prob_basic1} that
%and thus
\begin{align}\label{eq: sphere_concentrate}
    P[| u_i| \geq \sqrt{\xi/d} ] % \leq& \exp\left(\frac{1}{2}\left(1-\xi + \log \xi\right)\right) \nonumber\\
    \leq & \exp\left(-\xi/4 \right) = \exp\left(-\log \left( {dT/\delta} \right)\right)  
    =   \delta/dT.
\end{align}

Recall that  the ZO gradient estimate $\hat{\mathbf g}_t$ is given by the form
\begin{align}\label{eq: est_temp}
    \hat{\nabla}f  (\mathbf x ) =
    (d/\mu)   [  f ( \mathbf x + \mu \mathbf u ) - f ( \mathbf x )]  \mathbf u.
\end{align}
By  Lipschitz of $f$ under \textbf{A2}, 
% \textcolor{Sijia_color}{[SL: we did not specify A2 in the claim of Prop. and does A2 imply the Lipschtiz?]} \textcolor{XC_color}{[XC: I assumed A1 and A2 hold for the hold paper, probably I should state it in the proposition, bounded $L_2$ norm on gradient is equivalent to Lipschitz consinuity on $f$]}, 
the $i$th coordinate of the ZO gradient estimate \eqref{eq: est_temp} is upper bounded by $dL_c|u_{i}|$. Since $\mathbf u$ is drawn uniformly randomly from a unit sphere, by \eqref{eq: sphere_concentrate}  we have
\begin{align}\label{eq: sphere_concentrate_v2}
    P[ d L_c | u_i| \geq L_c \sqrt{\xi d} ] % \leq& \exp\left(\frac{1}{2}\left(1-\xi + \log \xi\right)\right) \nonumber\\
    \leq  \delta/dT.
\end{align}
Also, since $| \hat{g}_{t,i}  | \leq d L_c |u_i|$, based on \eqref{eq: sphere_concentrate_v2} we obtain that 
\begin{align}\label{eq: single_bound_v0}
 P[|\hat{g}_{t,i}| \geq L_c \sqrt{\xi d}]  \leq 
 P[ d L_c | u_i| \geq L_c \sqrt{\xi d} ]
  \leq \delta/dT.
%    P[|\hat{g}_{t,i}| \geq 2\eta\sqrt{ d\log ( {dT}/{\delta} ) } ] \leq \delta/dT
\end{align}
Substituting $\xi = 4 \log \frac{dT}{\delta}$ into \eqref{eq: single_bound_v0}, we have
\begin{align}\label{eq: single_bound_v1}
        P[|\hat{g}_{t,i}| \geq 2 L_c\sqrt{ d\log ( {dT}/{\delta} ) } ] \leq \delta/dT
\end{align}
% \textcolor{Sijia_color}{[It is not clear how to obtain the above inequality. Is there any mistake at RHS of the above inequality?.]} [\textcolor{XC_color}{[XC: $u_i$ is from a unit spere distributionsubstitute ]}

Then by the union bound and \eqref{eq: single_bound_v1}, we have 
\begin{align}
    & P[|\hat{g}_{t,i}| \geq 2L_c \sqrt{ d\log ( {dT}/{\delta} ) }, \, \forall i, t] \nonumber  \\
    \leq& \sum_{t \in [T]}  \sum_{i\in [d]} P[|\hat{g}_{t,i}| \geq 2L_c\sqrt{ d\log ( {dT}/{\delta} ) }] \nonumber \leq dT (\delta/dT) = \delta,
\end{align}
which implies the inequality \eqref{eq: single_bound}.
%The proof is finished by the union bound to make \eqref{eq: single_bound} satisfied for all $t \in [T]$ and $i \in [d]$. 
%and substituting reduced $G_{zo}$ into \eqref{eq: final_bound}. 
\hfill $\square$
\textcolor{black}{
\subsection{Proof of Theorem \ref{thm: zo_adamm_high_prob}}\label{app: proof_adamm_high_prob}
The idea is to prove a similar result as Proposition \ref{prop: zo_adamm_nonconvex} conditioned on the event in Proposition \ref{prop: g_zo_bound} ($\max_{t\in [T]} \{  \| \hat{\mathbf g}_t \|_\infty \} \leq 2 L_c \sqrt{ d\log ({dT}/{\delta}) }$). Thus, the proof follows the same flow as Proposition \ref{prop: zo_adamm_nonconvex}.} 
\textcolor{black}{
The difference is that \eqref{eq: unbiased_gradient} does not hold conditioned on the event and more efforts are need to bound the corresponding term in \eqref{eq: unbiased_gradient}.} 
\textcolor{black}{
Denote the event that $\max_{t\in [T]} \{  \| \hat{\mathbf g}_t \|_\infty \} \leq 2 L_c \sqrt{ d\log ({dT}/{\delta}) }$ to be $U(\delta)$, we need to upper bound 
\begin{align}
    \mathbb E \left[\langle \nabla f_{\mu}(\mathbf x_t),  \alpha_{t-1} \mathbf{\hat V}_{t-1}^{-1/2}  (\mathbf{\hat g}_t - \nabla  f_{\mu}(\mathbf x_t))  \rangle  | U(\delta)\right]  %\nonumber \\
\end{align}
where $\mathbb E[\cdot|U(\delta)]$ is conditional expectation conditioned on $U(\delta)$.}

\textcolor{black}{
By Proposition \ref{prop: g_zo_bound}, we know $P(U(\delta)) \geq 1- \delta$ and using the fact that  $E[\cdot|A] = \frac{E[\cdot] - E[\cdot|A^c]P(A^c)}{P(A)}$ for any event $A$ and its complimentary event $A^c$, we have
\begin{align}
        &\mathbb E \left[\langle \nabla f_{\mu}(\mathbf x_t),  \alpha_{t-1} \mathbf{\hat V}_{t-1}^{-1/2}  (\mathbf{\hat g}_t - \nabla  f_{\mu}(\mathbf x_t))  \rangle  | U(\delta)\right] \nonumber \\
        \leq & \frac{\mathbb E \left[\langle \nabla f_{\mu}(\mathbf x_t),  \alpha_{t-1} \mathbf{\hat V}_{t-1}^{-1/2}  (\mathbf{\hat g}_t - \nabla  f_{\mu}(\mathbf x_t))  \rangle \right]  }{1-\delta} \nonumber \\
        & +\frac{\delta \left|\mathbb E \left[\langle \nabla f_{\mu}(\mathbf x_t),  \alpha_{t-1} \mathbf{\hat V}_{t-1}^{-1/2}  (\mathbf{\hat g}_t - \nabla  f_{\mu}(\mathbf x_t))  \rangle  | U(\delta)^c\right]\right|  }{1-\delta}
\end{align}
and further we have 
\begin{align}
     &\left|\mathbb E \left[\langle \nabla f_{\mu}(\mathbf x_t),  \alpha_{t-1} \mathbf{\hat V}_{t-1}^{-1/2}  (\mathbf{\hat g}_t - \nabla  f_{\mu}(\mathbf x_t))  \rangle  | U(\delta)^c\right]\right| \nonumber \\
     \leq &d\frac{\alpha_{t-1}}{c}(\eta^2 + \eta \max_{t\in [T]} \|\hat g_t\|_{\infty} ) \nonumber \\
     \leq & d\frac{\alpha_{t-1}}{c}(\eta^2 + \eta d L_c   ) 
\end{align}
where the first inequality is due to $\| \nabla f_{\mu}(x_t)\|_{\infty} \leq \eta$ and $\hat v_{t-1}^{1/2} \geq \hat {\mathbf v}_{0}^{1/2} \geq c \mathbf 1 $, the second inequality is due to \eqref{eq: grad_rand} and Lipschitz continuity of $f(\mathbf x; \boldsymbol{\xi })$. }

\textcolor{black}{
Using the fact that $\mathbb E \left[\langle \nabla f_{\mu}(\mathbf x_t),  \alpha_{t-1} \mathbf{\hat V}_{t-1}^{-1/2}  (\mathbf{\hat g}_t - \nabla  f_{\mu}(\mathbf x_t))  \rangle \right] = 0$ proved in in \eqref{eq: unbiased_gradient} and set $\delta = 1/Td^{0.5}$, we have for $T \geq 2$
\begin{align}\label{eq: bound_bias_high_prob}
    & \mathbb E \left[\langle \nabla f_{\mu}(\mathbf x_t),  \alpha_{t-1} \mathbf{\hat V}_{t-1}^{-1/2}  (\mathbf{\hat g}_t - \nabla  f_{\mu}(\mathbf x_t))  \rangle  | U(1/Td^{0.5})\right] \nonumber \\
    \leq & 2 \frac{1}{T d^{0.5}}d\frac{\alpha_{t-1}}{c}(\eta^2 + \eta d L_c   ) = 2\frac{d^{1.5}}{T} \frac{\alpha_{t-1}}{c} \eta L_c + 2\frac{d^{0.5}}{T}  \frac{\alpha_{t-1}}{c}\eta^2 
\end{align}
}

\textcolor{black}{
Replacing \eqref{eq: unbiased_gradient} with \eqref{eq: bound_bias_high_prob} and going through the rest of the proof of Proposition \eqref{prop: zo_adamm_nonconvex}, one can finally get 
\begin{align}\label{eq: conditioned_rate}
     & \mathbb E \left[ \| \mathbf{\hat V}_t ^{-1/4}\nabla f(x_R)\|^2  \big | U(1/Td^{0.5})\right ]   \nonumber \\
    \leq  & \frac{\mu ^2 d^2 L^2}{2c} + 2\frac{\sqrt{d}}{\sqrt{T}}D_f  + 2\frac{1}{T}\mathbb E \left[\left(\frac{\eta G _{zo} }{1-\beta_1} + 2\eta2\right)  \left\| \frac{1}{\sqrt{\mathbf{\hat v}_{0}}} \right\|_1 \bigg| U(1/Td^{0.5}) \right]  \nonumber\\
    &+ \frac{\sqrt{d}}{\sqrt{T}}  \frac{4L_g +  5L_g\beta_1^2}{(1-\beta_1)^2} \frac{1-\beta_1}{1-\beta_2}\frac{1}{1-\gamma} +  2 \frac{d^{1.5}}{T} \frac{\eta L_c}{c}  + 2 \frac{d^{0.5}}{T}  \frac{\eta^2}{c}.  \nonumber
\end{align}
Since in the event of $U(1/Td^{0.5})$, we have 
\begin{align}
   G_{zo} =  \max_{t\in [T]} \{  \| \hat{\mathbf g}_t \|_\infty \} \leq 2 L_c \sqrt{ d\log ({d^{1.5}T^2}) } = 2 L_c \sqrt{ d }\sqrt{1.5 \log d + 2 \log T }.
\end{align}
Substituting the above inequality into \eqref{eq: conditioned_rate}, we get the desired result.
%and $ \hat V_{t,ii}^{-1/2} \geq {1}/{\max_{t\in [T]} \{  \| \hat{\mathbf g}_t \|_\infty \} }$ (by the update rule),
%\begin{align} 
%    & \frac{\mathbb E \left[ \|\nabla f(x_R)\|^2 \right | U(1/Td^{0.5})]}{2 L_c \sqrt{ d }\sqrt{\log d + 3 \log T }}  
%     \leq \mathbb E \left[ \| \mathbf{\hat V}_t ^{-1/4}\nabla f(x_R)\|^2 \right | U(1/Td^{0.5})] \nonumber \\  
%    \leq  & \frac{\mu ^2 d^2 L^2}{2c} + 2\frac{\sqrt{d}}{\sqrt{T}}D_f  + 2\frac{1}{T}\mathbb E \left[\left(\frac{\eta G _{zo} }{1-\beta_1} + 2\eta2\right)  \left\| \frac{1}{\sqrt{\mathbf{\hat v}_{0}}} \right\|_1 | U(1/Td^{0.5}) \right]  \nonumber\\
 %   &+ \frac{\sqrt{d}}{\sqrt{T}}  \frac{4L_g +  5L_g\beta_1^2}{(1-\beta_1)^2} \frac{1-\beta_1}{1-\beta_2}\frac{1}{1-\gamma} +  2 \frac{d^{1.5}}{T} \frac{\eta L_c}{c}  +2  \frac{d^{0.5}}{T}  \frac{\eta^2}{c}.  \nonumber
%\end{align}
%Rearrange the above inequality and take big O notation, we get the desired result. 
\hfill $\square$}
% \section{Proof for constrained Nonconvex optimization}

\subsection{Proof of Theorem \ref{thm: nonconvex_cons}}\label{app: proof_uncons}
To proceed into proof of Theorem \ref{thm: nonconvex_cons}, 
we give a few technical lemmas for the properties of \eqref{eq: prox}.

\begin{lemma} \label{lem: inner_project}
For any symmetric $\mathbf H \succeq 0, \mathbf g, \omega$, we have
\begin{equation}  
    \langle \mathbf g, P_{\mathcal{X},\mathbf H}(\mathbf x^-, \mathbf g,\omega)  \rangle \geq  \|\mathbf H^{1/2} P_{\mathcal{X},\mathbf H}(\mathbf x^-, \mathbf g,\omega) \|^2
\end{equation}
\end{lemma}
\noindent \textbf{Proof of Lemma \ref{lem: inner_project}:} By definition of $\mathbf x^+$, the optimality condition of \eqref{eq: project} is 
\begin{align}
    \langle \mathbf g + \frac{1}{\omega} \mathbf H (\mathbf x^+-\mathbf x^-), \mathbf x  - \mathbf x^+ \rangle \geq 0 \quad \forall \mathbf x  \in \mathcal{X} \nonumber
\end{align}

Thus 
\begin{align}
    \langle \mathbf g + \frac{1}{\omega} \mathbf H (\mathbf x^+-x), \mathbf x - \mathbf x^+ \rangle \geq 0 \nonumber
\end{align}
which can be rearranged to 
\begin{align}
    &\langle \mathbf g, P_{\mathcal{X},\mathbf H}(\mathbf x^-, \mathbf g,\omega)  \rangle = \frac{1}{\omega}\langle \mathbf g  , x - \mathbf x^+ \rangle  
    \geq \frac{1}{\omega^2} \langle  \mathbf H (x- \mathbf x^+), x - \mathbf x^+ \rangle = \|\mathbf H^{1/2} P_{\mathcal{X},\mathbf H}(\mathbf x^-, \mathbf g,\omega) \|^2 \nonumber
\end{align}
This completes the proof.
\hfill $\square$

\begin{lemma} \label{lem: non_exp}
Let $\mathbf{x}_1^+$ and $\mathbf{x}_2^+$ be given by \eqref{eq: project} with $\mathbf g$ replaced by $\mathbf{g}_1$ and $\mathbf{g}_2$, with $H \succ 0$, we have %{\color{red}[are we assuming $H$ is pd? how about V? define $\lambda_{\min}$?]}\textcolor{XC_color}{[Added a assumption]}
\begin{align}
    \|\mathbf{x}_1^+ - \mathbf{x}_2 ^+ \| & \leq \frac{\omega}{\lambda_{\min}(\mathbf H)} \|\mathbf{g}_1 - \mathbf{g}_2 \|\label{eq: non_exp_1}\\
    \|\mathbf{H}^{1/2}(\mathbf{x}_1^+ - \mathbf{x}_2 ^+) \| & \leq \omega \|\mathbf{H}^{-1/2}(\mathbf{g}_1 - \mathbf{g}_2) \|\label{eq: non_exp_2}.
\end{align}
where $\lambda_{\min}(\mathbf H)$ is the minimum eigenvalue of $\mathbf H$.
\end{lemma}

\noindent \textbf{Proof of Lemma \ref{lem: non_exp}:} By definition of $\mathbf x^+$, the optimality condition of \eqref{eq: project} is 
\begin{align}
    \langle \mathbf g + \frac{1}{\omega} \mathbf H (\mathbf x^+-\mathbf x^-), \mathbf x  - \mathbf x^+ \rangle \geq 0 \quad \forall \mathbf x  \in \mathcal{X} \nonumber
\end{align}

Thus, we have 
\begin{align}
    \langle \mathbf{g}_1 + \frac{1}{\omega} \mathbf H (\mathbf{x}_1^+-\mathbf x^-, \mathbf{x}_2^+ - \mathbf{x}_1^+ \rangle \geq 0 \nonumber 
\end{align}
\begin{align}
    \langle \mathbf{g}_2 + \frac{1}{\omega} \mathbf H (\mathbf{x}_2^+-\mathbf x^-, \mathbf{x}_1^+ - \mathbf{x}_2^+ \rangle \geq 0  \nonumber
\end{align}

Summing up the above two inequalities, we get
\begin{align} \label{eq: sum_optimality}
 \hspace{-0.4cm}   \langle \mathbf{g}_1 - \mathbf{g}_2, \mathbf{x}_2^+ - \mathbf{x}_1^+ \rangle \geq \frac{1}{\omega} \langle  \mathbf H (\mathbf{x}_2^+-\mathbf{x}_1^+), \mathbf{x}_2^+ - \mathbf{x}_1^+ \rangle 
\end{align}
By Cauchy-Schwarz inequality, we get
\begin{align}
    \| \mathbf{g}_1 - \mathbf{g}_2\| \| \mathbf{x}_2^+ - \mathbf{x}_1^+ \|
    \geq  \langle \mathbf{g}_1 - \mathbf{g}_2, \mathbf{x}_2^+ - \mathbf{x}_1^+ \rangle 
    \geq & \frac{1}{\omega} \langle  \mathbf H (\mathbf{x}_2^+-\mathbf{x}_1^+), \mathbf{x}_2^+ - \mathbf{x}_1^+ \rangle \nonumber \\
     \geq  & \frac{1}{\omega} \lambda_{\min}(\mathbf H)  \|\mathbf{x}_2^+-\mathbf{x}_1^+\|^2 \nonumber
\end{align}
which gives \eqref{eq: non_exp_1}.

Further, by \eqref{eq: sum_optimality} and Cauchy-Schwarz, we also have
\begin{align}
    & \| \mathbf H^{-1/2} (\mathbf{g}_1 - \mathbf{g}_2)\| \| \mathbf H^{1/2} (\mathbf{x}_2^+ - \mathbf{x}_1^+) \| \geq \langle \mathbf{g}_1 - \mathbf{g}_2, \mathbf{x}_2^+ - \mathbf{x}_1^+ \rangle \nonumber \\ 
    \geq & \frac{1}{\omega} \langle  \mathbf H (\mathbf{x}_2^+-\mathbf{x}_1^+), \mathbf{x}_2^+ - \mathbf{x}_1^+ \rangle =\frac{1}{\omega}  \|\mathbf H^{1/2}(\mathbf{x}_2^+-\mathbf{x}_1^+)\|^2 \nonumber
\end{align}
which gives \eqref{eq: non_exp_2}. This completes the proof. \hfill $\square$

The following lemma characterizes the difference between projected points if different distance matrices are used in ZO-AdaMM.

\begin{lemma}\label{lem: project_diff_v}
Assume $\mathbf V_t^{1/2} \geq c \mathbf I$, ZO-AdaMM yields 
\begin{align} \label{eq: diff_bound_con}
  & \left \| (P_{\mathcal{X},{\mathbf{\hat V}_{t-1}^{1/2}}}(\mathbf x_t,\nabla f_{\mu}(\mathbf x_t),\alpha_t)  - P_{\mathcal{X},{\mathbf{\hat V}_t^{1/2}}}(\mathbf x_t,\nabla f_{\mu}(\mathbf x_t),\alpha_t) ) \right \| ^2 \leq \sum_{i=1}^d v_{t,i}^{1/2}  (\hat v_{t,i}^{1/2}  - \hat v_{t-1,i}^{1/2}) \frac{1}{c^4} \eta^2.
\end{align}
\end{lemma}
\noindent \textbf{Proof of Lemma \ref{lem: project_diff_v}: }
Recall the optimality condition of \eqref{eq: project} is 
\begin{align}\label{eq: opt_condition}
    \langle \mathbf g + \frac{1}{\omega} \mathbf H (\mathbf x^+-\mathbf x^-, \mathbf x  - \mathbf x^+ \rangle \geq 0 \quad \forall \mathbf x  \in \mathcal{X}
\end{align}
Let us define 
\begin{align*}
    \mathbf x_t^* & \triangleq \mathbf x_t - \alpha_t P_{\mathcal{X},{\mathbf{\hat V}_{t-1}^{1/2}}}(\mathbf x_t,\nabla f_{\mu}(\mathbf x_t),\alpha_t) \\ \mathbf{\tilde x}_t^* &\triangleq \mathbf x_t - \alpha_t P_{\mathcal{X},{\mathbf{\hat V}_t^{1/2}}}(\mathbf x_t,\nabla f_{\mu}(\mathbf x_t),\alpha_t).
\end{align*}
By optimality condition \eqref{eq: opt_condition}, we have
\begin{align}
    \langle \nabla f_{\mu} (\mathbf x_t) + \frac{1}{\alpha_t} \mathbf{\hat V}_t^{1/2} (\mathbf{\tilde x}_t^*- \mathbf x_t), \mathbf x_t^* - \mathbf{\tilde x}_t^* \rangle &\geq 0 \nonumber\\
    \langle \nabla f_{\mu} (\mathbf x_t) + \frac{1}{\alpha_t} \mathbf{\hat V}_{t-1}^{1/2} (\mathbf x_t^*- \mathbf x_t), \mathbf{\tilde x}_t^* - \mathbf x_t^* \rangle &\geq 0 \nonumber
\end{align}
Summing the above up, we get 
\begin{align}
    \langle  \mathbf{\hat V}_t^{1/2} (\mathbf{\tilde x}_t^*- \mathbf x_t) - \mathbf{\hat V}_{t-1}^{1/2} (\mathbf x_t^*- \mathbf x_t), \mathbf x_t^* - \mathbf{\tilde x}_t^* \rangle \geq 0 \nonumber
\end{align}
which is equivalent to 
\begin{align}
& \langle  (\mathbf{\hat V}_t^{1/2}  - \mathbf{\hat V}_{t-1}^{1/2}) (\mathbf x_t^*- \mathbf x_t), \mathbf x_t^* - \mathbf{\tilde x}_t^* \rangle \nonumber \\
& \quad 
    + \langle  \mathbf{\hat V}_t^{1/2} (\mathbf{\tilde x}_t^* - \mathbf x_t^*) , \mathbf x_t^* - \mathbf{\tilde x}_t^* \rangle \geq 0. \nonumber
\end{align}
Further rearranging, we have
\begin{align}
    & \langle   (\mathbf{\hat V}_t^{1/2}  - \mathbf{\hat V}_{t-1}^{1/2}) (\mathbf x_t^*- \mathbf x_t), \mathbf x_t^* - \mathbf{\tilde x}_t^* \rangle 
    \geq  \| \mathbf{\hat V}_t^{1/4} (\mathbf{\tilde x}_t^* - \mathbf x_t^* ) \|^2 \geq c \|(\mathbf{\tilde x}_t^* - \mathbf x_t^* ) \|^2 \nonumber
\end{align}
which implies (by using Cauchy-Swartz on the left hand side and then squaring both sides)
\begin{align}
  c^2\| (\mathbf{\tilde x}_t^* - \mathbf x_t^* ) \| ^2 \leq & \|   ( \mathbf{\hat V}_t^{1/2}  - \mathbf{\hat V}_{t-1}^{1/2}) (\mathbf x_t^*- \mathbf x_t)\|^2 =\sum_{i=1}^d   (\hat v_{t,i}^{1/2}  - \hat v_{t-1,i}^{1/2})^2 (\hat x_{t,i}^*-x_{t,i})^2  \nonumber \\
   \stackrel{(a)}{\leq} & \sum_{i=1}^d \hat v_{t,i}^{1/2}   (\hat v_{t,i}^{1/2}  - \hat v_{t-1,i}^{1/2}) \|\hat x_{t}^*-x_{t}\|^2 
   \stackrel{(b)}{\leq} \sum_{i=1}^d \hat v_{t,i}^{1/2}   (\hat v_{t,i}^{1/2}  - \hat v_{t-1,i}^{1/2}) \frac{1}{c^2}\alpha_t^2 \|\nabla f_{\mu}(x_t)\|^2 \nonumber \\
   \leq & \sum_{i=1}^d \hat v_{t,i}^{1/2}   (\hat v_{t,i}^{1/2}  - \hat v_{t-1,i}^{1/2}) \frac{1}{c^2}\alpha_t^2 \eta^2
\end{align}
where (a) is due to $\hat v_{t,i}^{1/2} \geq \hat v_{t-1,i}^{1/2}$ and (b) is due to Lemma \ref{lem: non_exp} by treating $\mathbf g_1 = \nabla f_{\mu}(\mathbf x_t),\, \mathbf g_2 = 0,\, \mathbf x^- = \mathbf x_t,\, H  = \mathbf{\hat V}_t^{1/2}$. %{\color{red}[it is not clear to me how the last inequality obtains.]}\textcolor{XC_color}{[Added more explanations in the inequalities.]}
Substituting \eqref{eq: prox} into LHS of the above inequality and rearrange, we get \eqref{eq: diff_bound_con}.
% \begin{align}
%   &\| (P_{\mathcal{X},{\mathbf{\hat V}_{t-1}^{1/2}}}(x_t,\nabla f_{\mu}(\mathbf x_t),\alpha_t) - P_{\mathcal{X},{\mathbf{\hat V}_t^{1/2}}}(\mathbf x_t,\nabla f_{\mu}(\mathbf x_t),\alpha_t) ) \| ^2 \nonumber\\
%   \leq & \sum_{i=1}^d v_{t,i}^{1/2}  (\hat v_{t,i}^{1/2}  - \hat v_{t-1,i}^{1/2}) \frac{1}{c^4} \eta^2
% \end{align}
This completes the proof. \hfill $\square$

Now we are ready to prove our main theorem.

\noindent \textbf{Proof of Theorem \ref{thm: nonconvex_cons}:}

We start with standard decent lemma in nonconvex optimization. By Lipschitz smoothness of $f_{\mu}$, we have
\begin{align}\label{eq:descent_constraint}
    f_{\mu}(\mathbf x_{t+1}) \leq & f_{\mu}(\mathbf x_t)  - \alpha_t \langle \nabla f_{\mu}(\mathbf x_t), P_{\mathcal{X},{\mathbf{\hat V}_t^{1/2}}}(\mathbf x_t, \mathbf{\hat g}_t,\alpha_t) \rangle   + \frac{L}{2}\alpha_t^2 \|P_{\mathcal{X},{\mathbf{\hat V}_t^{1/2}}}(\mathbf x_t,\mathbf{\hat g}_t,\alpha_t)\|^2.
\end{align}
We need to upper bound RHS of the above inequality and split out a descent quantity. %\textcolor{XC_color}{[XC: will change overlength equations if we use double column in appendix]}\textcolor{XC_color}{[This was an outdated comment]} {\color{red}[I think we can use single-column for the appendix; I did it last year.]}
\begin{align} \label{eq: split_bounds}
  &-\langle \nabla f_{\mu}(\mathbf x_t), P_{\mathcal{X},{\mathbf{\hat V}_t^{1/2}}}(\mathbf x_t,\mathbf{\hat g}_t,\alpha_t) \rangle  \nonumber\\
  = & -\langle  \mathbf{\hat g}_t, P_{\mathcal{X},{\mathbf{\hat V}_t^{1/2}}}(\mathbf x_t,\mathbf{\hat g}_t,\alpha_t) \rangle   + \langle  \mathbf{\hat g}_t- \nabla f_{\mu}(\mathbf x_t) , P_{\mathcal{X},{\mathbf{\hat V}_t^{1/2}}}(\mathbf x_t,\mathbf{\hat g}_t,\alpha_t) \rangle \nonumber\\
  \leq & -\|\mathbf{\hat V}_t^{1/4} P_{\mathcal X,{\mathbf{\hat V}_t^{1/2}}}(\mathbf x_t, \mathbf{\hat g}_t,\alpha_t) \|^2  + \langle  \mathbf{\hat g}_t - \nabla f_{\mu}(\mathbf x_t) , P_{\mathcal{X},{\mathbf{\hat V}_t^{1/2}}}(\mathbf x_t,\mathbf{\hat g}_t,\alpha_t) \rangle
\end{align}
where the inequality is by Lemma \eqref{lem: inner_project} and some simple substitutions. %{\red[notation issue: in the lemma g does not have hat.]}\textcolor{XC_color}{[The lemma is for any $\mathbf g$, there are some substitutions here e.g. $ \mathbf g = \mathbf  {\hat g}_t$.]}

Further, for the last term in RHS of \eqref{eq: split_bounds} we have
\begin{align} \label{eq: ABC}
    &\langle  \mathbf{\hat g}_t - \nabla f_{\mu}(\mathbf x_t) , P_{\mathcal{X},{\mathbf{\hat V}_t^{1/2}}}(\mathbf x_t,\mathbf{\hat g}_t,\alpha_t) \rangle \nonumber \\
    & =  \begin{rcases}+\langle\mathbf{\hat g}_t - \nabla  f_{\mu}(\mathbf x_t) , P_{\mathcal{X},{\mathbf{\hat V}_t^{1/2}}}(\mathbf x_t,\mathbf{\hat g}_t,\alpha_t) \rangle\\
  -\langle\mathbf{\hat g}_t - \nabla  f_{\mu}(\mathbf x_t), P_{\mathcal{X},{\mathbf{\hat V}_{t}^{1/2}}}(\mathbf x_t,\nabla f_{\mu}(\mathbf x_t),\alpha_t)\rangle
  \end{rcases}A\nonumber\\
  & \begin{rcases}
  {+\langle  \mathbf{\hat g}_t- \nabla  f_{\mu}(\mathbf x_t),P_{\mathcal{X},{\mathbf{\hat V}_{t}^{1/2}}}(\mathbf x_t,\nabla f_{\mu}(\mathbf x_t),\alpha_t)}\rangle  \\
   { -\langle \mathbf{\hat g}_t- \nabla  f_{\mu}(\mathbf x_t),  P_{\mathcal{X},{\mathbf{\hat V}_{t-1}^{1/2}}}(\mathbf x_t,\nabla f_{\mu}(\mathbf x_t),\alpha_t)\rangle}
  \end{rcases} B\nonumber \\ 
  & + \underbrace{\langle  \mathbf{\hat g}_t -\nabla  f_{\mu}(\mathbf x_t) , P_{\mathcal{X},{\mathbf{\hat V}_{t-1}^{1/2}}}(\mathbf x_t,\nabla f_{\mu}(\mathbf x_t),\alpha_t)\rangle}_{C}
\end{align}
Next, we bound the three terms in RHS of \eqref{eq: ABC}. 

Let's bound term $A$ first, with the assumption $\mathbf{\hat V}^{1/2} \geq c \mathbf I$, by Lemma \ref{lem: non_exp}, \eqref{eq: prox} and Cauchy-Schwartz inequality, we have:
\begin{align}
    A =& \langle    \mathbf{\hat g}_t - \nabla f_{\mu}(\mathbf x_t),  P_{\mathcal{X},{\mathbf{\hat V}_t^{1/2}}}(\mathbf x_t,\mathbf{\hat g}_t,\alpha_t) - P_{\mathcal{X},{\mathbf{\hat V}_{t}^{1/2}}}(\mathbf x_t,\nabla f_{\mu}(\mathbf x_t),\alpha_t)\rangle \leq  \frac{1}{c} \|\mathbf{\hat g}_t - f_{\mu}(\mathbf x_t)\|^2 
\end{align}

Now let's bound term $C$, because $\mathbb {E}[\mathbf{\hat g}_t] = \nabla f_\mu(\mathbf x_t)$ and the noise in $\mathbf{\hat g}_t$ is independent of $\nabla f_\mu(\mathbf x_t)$ and $\mathbf{\hat V}_{t-1}$, we have
\begin{align}
    \mathbb {E}[\langle \nabla  f_{\mu}(\mathbf x_t) - \mathbf{\hat g}_t, P_{\mathcal{X},{\mathbf{\hat V}_{t-1}^{1/2}}}(\mathbf x_t,\nabla f_{\mu}(\mathbf x_t),\alpha_t)\rangle] = 0
\end{align}
Substituting the above bounds for A and C, into \eqref{eq: ABC} and \eqref{eq: split_bounds}, using Young's inequality on term B, we have
\begin{align} \label{eq: split_decrease_con}
    & -\mathbb {E}[\langle \nabla f_{\mu}(\mathbf x_t), P_{\mathcal{X},{\mathbf{\hat V}_t^{1/2}}}(\mathbf x_t,\mathbf{\hat g}_t,\alpha_t) \rangle ] \nonumber\\
    \leq & -\mathbb {E}[\|\mathbf{\hat V}_t^{1/4} P_{\mathcal X,{\mathbf{\hat V}_t^{1/2}}}(\mathbf x_t, \mathbf{\hat g}_t,\alpha_t) \|^2] + \frac{1}{c} \mathbb {E}[\|\mathbf{\hat g}_t - f_{\mu}(\mathbf x_t)\|^2]  + \frac{1}{2} \mathbb {E}[\|\mathbf{\hat g}_t - f_{\mu}(\mathbf x_t)\|^2]   + \frac{1}{2} \mathbb {E}[B_2]
\end{align}
where we define
\begin{align*}
    B_2 \Def  & \left \| (P_{\mathcal{X},{\mathbf{\hat V}_{t-1}^{1/2}}}(\mathbf x_t,\nabla f_{\mu}(\mathbf x_t),\alpha_t)  - P_{\mathcal{X},{\mathbf{\hat V}_t^{1/2}}}(\mathbf x_t,\nabla f_{\mu}(\mathbf x_t),\alpha_t) ) \right \| ^2.
\end{align*}
% $B_2 \triangleq {\mathbb {E}[\|P_{\mathcal{X},{\mathbf{\hat V}_{t}^{1/2}}}(\mathbf x_t,\nabla f_{\mu}(\mathbf x_t),\alpha_t) - P_{\mathcal{X},{\mathbf{\hat V}_{t-1}^{1/2}}}(\mathbf x_t,\nabla f_{\mu}(\mathbf x_t),\alpha_t)\|^2]}.$
What remains is to bound the term $B2$ which is given by Lemma \ref{lem: project_diff_v}.

Combining \eqref{eq:descent_constraint}, \eqref{eq: split_decrease_con}, \eqref{eq: diff_bound_con}, we have
\begin{align}
    \mathbb {E}[f_{\mu}(\mathbf x_{t+1})] 
    \leq & \mathbb {E}[f_{\mu}(\mathbf x_t)]  -\alpha_t \mathbb {E}[\|\mathbf{\hat V}_t^{1/4} P_{\mathcal X,{\mathbf{\hat V}_t^{1/2}}}(\mathbf x_t, \mathbf{\hat g}_t,\alpha_t) \|^2]  + \alpha_t(\frac{1}{c}+\frac{1}{2}) \mathbb {E}[\|\mathbf{\hat g}_t - f_{\mu}(\mathbf x_t)\|^2] \nonumber \\
    & \hspace*{-0.3in}+ \alpha_t \frac{1}{2} \mathbb {E}\big[\sum_{i=1}^d \hat v_{t,i}^{1/2}  (\hat v_{t,i}^{1/2}  - \hat v_{t-1,i}^{1/2}) \frac{1}{c^4} \eta^2\big]  + \frac{L}{2}\alpha_t^2\mathbb {E}\left[\frac{1}{c^2}\| \mathbf{\hat V}_t^{1/4}P_{\mathcal{X},{\mathbf{\hat V}_t^{1/2}}}(\mathbf x_t,\mathbf{\hat g}_t,\alpha_t)\|^2\right] 
\end{align}
which can be rearranged into
\begin{align}\label{eq:tele_con}
   &  (\alpha_t-\frac{L}{2c^2}\alpha_t^2) \mathbb {E}[\|\mathbf{\hat V}_t^{1/4} P_{\mathcal X,{\mathbf{\hat V}_t^{1/2}}}(\mathbf x_t, \mathbf{\hat g}_t,\alpha_t) \|^2] \nonumber \\
    \leq &  \mathbb {E}[f_{\mu}(\mathbf x_t)] - \mathbb {E}[f_{\mu}(\mathbf x_{t+1})]    + \alpha_t(\frac{1}{c}+\frac{1}{2}) \mathbb {E}[\|\mathbf{\hat g}_t - f_{\mu}(\mathbf x_t)\|^2] \nonumber \\
    & + \alpha_t \frac{1}{2} \mathbb {E}\left[\sum_{i=1}^d \hat v_{t,i}^{1/2}  (\hat v_{t,i}^{1/2}  - \hat v_{t-1,i}^{1/2}) \frac{1}{c^4} \eta^2\right]. 
\end{align}

In addition, we have
\begin{align} \label{eq: split_variance}
    & \|\mathbf{\hat V}_t^{1/4} P_{\mathcal{X},{\mathbf{\hat V}_t^{1/2}}}(\mathbf x_t,\nabla f(\mathbf x_t),\alpha_t) \|^2 
    \leq  3 \|\mathbf{\hat V}_t^{1/4} P_{\mathcal{X},{\mathbf{\hat V}_t^{1/2}}}(\mathbf x_t, \mathbf{\hat g}_t ,\alpha_t) \|^2 \nonumber \\
    & \hspace*{0.5in}+ 3 \left. \|\mathbf{\hat V}_t^{1/4} (P_{\mathcal{X},{\mathbf{\hat V}_t^{1/2}}}(\mathbf x_t,\nabla f_{\mu}(\mathbf x_t) , \alpha_t) \right. \left. - P_{\mathcal{X},{\mathbf{\hat V}_t^{1/2}}}(\mathbf x_t,\nabla f(\mathbf x_t) , \alpha_t))\|^2 \right. \nonumber \\
    &\hspace*{0.5in} + \left. 3 \|\mathbf{\hat V}_t^{1/4}(P_{\mathcal{X},{\mathbf{\hat V}_t^{1/2}}}(\mathbf x_t, \mathbf{\hat g}_t,\alpha_t) \right.  \left. - P_{\mathcal{X},{\mathbf{\hat V}_t^{1/2}}}(\mathbf x_t,\nabla f_{\mu}(\mathbf x_t),\alpha_t)) \|^2 \right. \nonumber \\
    \leq & 3\|\mathbf{\hat V}_t^{1/4} P_{\mathcal{X},{\mathbf{\hat V}_t^{1/2}}}(\mathbf x_t, \mathbf{\hat g}_t ,\alpha_t) \|^2   + \frac{3}{c} \|\nabla f_{\mu}(\mathbf x_t) - \nabla f(\mathbf x_t) \|^2 + \frac{3}{c}\|\mathbf{\hat g}_t - \nabla f_{\mu}(\mathbf x_t) \|^2
\end{align}
where the second inequality is by \eqref{eq: prox} and Lemma \eqref{lem: non_exp}

Combining \eqref{eq: split_variance} and \eqref{eq:tele_con}, we have
\begin{align}
    &\left(\alpha_t-\frac{L}{2c^2}\alpha_t^2\right) \|\mathbf{\hat V}_t^{1/4} P_{\mathcal{X},{\mathbf{\hat V}_t^{1/2}}}(\mathbf x_t,\nabla f(\mathbf x_t),\alpha_t) \|^2 \nonumber \\
     \leq & 3( \mathbb {E}[f_{\mu}(\mathbf x_t)] - \mathbb {E}[f_{\mu}(\mathbf x_{t+1})] )  + (3\alpha_t(\frac{1}{c}+\frac{1}{2}) + \frac{3}{c}(\alpha_t-\frac{L}{2c^2}\alpha_t^2)) \mathbb {E}[\|\mathbf{\hat g}_t - f_{\mu}(\mathbf x_t)\|^2] \nonumber \\
    & + \frac{3}{2}\alpha_t  \mathbb {E}[\sum_{i=1}^d \hat v_{t,i}^{1/2}  (\hat v_{t,i}^{1/2}  - \hat v_{t-1,i}^{1/2}) \frac{1}{c^4} \eta^2]   + \frac{3}{c}(\alpha_t-\frac{L}{2c^2}\alpha_t^2) \|\nabla f_{\mu}(\mathbf x_t) - \nabla f(\mathbf x_t) \|^2 
\end{align}
Summing over $t$ from 1 to $T$, setting $\alpha_t = \alpha$, and dividing both sides by $T(\alpha-\frac{L_g\alpha^2}{2c^2})$, we get  
\begin{align}\label{eq:final_con}
    & \frac{1}{T}\sum_{t=1} ^T  \mathbb {E}[\|\mathbf{\hat V}_t^{1/4} P_{\mathcal{X},{\mathbf{\hat V}_t^{1/2}}}(\mathbf x_t \nabla f(\mathbf x_t),\alpha_t) \|^2] \nonumber \\
    \leq & \frac{3}{T(\alpha-\frac{L_g\alpha^2}{2c})}(\mathbb {E}[f_{\mu}(\mathbf{x}_1)]  - \mathbb {E}[f_{\mu}(x_{T+1})])   + \left(\frac{3\alpha (c+2)}{2 T c (\alpha-\frac{L_g\alpha^2}{2c})}   + \frac{3}{Tc}\right)\sum_{t=1}^T  \mathbb {E}[\|\mathbf{\hat g}_t - f_{\mu}(\mathbf x_t)\|^2] \nonumber \\
    & + \frac{3\alpha}{2T(\alpha-\frac{L_g\alpha^2}{2c})}   \mathbb {E}[\sum_{i=1}^d \hat v_{T,i}  ]\frac{1}{c^4} \eta^2  + \frac{3}{Tc}\sum_{t=1}^T \mathbb {E}[\|\nabla f_{\mu}(\mathbf x_t) - \nabla f(\mathbf x_t) \|^2].
\end{align}
Choose $\alpha \leq \frac{c}{L}$, we have 
\begin{align}
    \alpha-\frac{L_g\alpha^2}{2c}  = \alpha\left(1- \frac{L_g\alpha}{2c}\right) \geq \alpha(1-\frac{1}{2}) = \frac{\alpha}{2}
\end{align}
and \eqref{eq:final_con} becomes 
\begin{align}\label{eq:final_con_2}
    & \frac{1}{T}\sum_{t=1} ^T  \mathbb {E} \left [\|\mathbf{\hat V}_t^{1/4} P_{\mathcal{X},{\mathbf{\hat V}_t^{1/2}}}(\mathbf x_t \nabla f(\mathbf x_t),\alpha_t) \|^2 \right ] \nonumber \\
    \leq & \frac{6}{T\alpha}D_f  + \frac{1}{T} (\frac{9}{c}+3)\sum_{t=1}^T  \mathbb {E}\left [\|\mathbf{\hat g}_t - f_{\mu}(\mathbf x_t)\|^2 \right ] + \frac{3}{T}  \frac{1}{c^4} \eta^2 \mathbb {E}\left [\sum_{i=1}^d \hat v_{T,i} \right ]  + \frac{3}{c}  \frac{\mu^2 d^2 L_g^2}{4}
\end{align}
where we defined $D_f \Def \mathbb {E}[f_{\mu}(\mathbf{x}_1)]  - \min_{x}f_{\mu}(x)$ and \textcolor{black}{used the fact that $\|\nabla f_{\mu}(\mathbf x_t) - \nabla f(\mathbf x_t) \|^2 \leq \frac{\mu^2 d^2 L_g^2}{4}$ by Lemma \ref{lemma: smooth_f_random}.}

\textcolor{black}{
Further, we have 
\begin{align}\label{eq: v_t_bound}
    & \mathbb {E}\left[\sum_{i=1}^d \hat v_{T,i}\right]   = \mathbb {E}\left[\sum_{i=1}^d \max_{t\in[T]} (1-\beta_2) \sum_{k=1}^t  \beta_2^{t-k} \hat g_{k,i}^2 \right] \nonumber \\
    \leq & \mathbb {E}\left[d \max_{t\in[T]} (1-\beta_2) \sum_{k=1}^t  \beta_2^{t-k}  \|\hat{g}_k\|_{\infty} \right] \nonumber \\
    \leq & \mathbb {E}\left[d \max_{t\in[T]}    \|\hat{g}_t\|_{\infty} \right] \nonumber \\
%=& (1-\beta_2) \sum_{k=1}^T \beta_2^{T-k} \mathbb {E}[  \| \hat {\mathbf g}_{k}\|^2  ]  
%    \leq (1-\beta_2)  \sum_{k=1}^T \beta_2^{T-k} 2 ( \mathbb {E}[\| \hat g_{k} -  \nabla f_{\mu}(\mathbf x_k)\|^2] +  \mathbb {E}[\|    \nabla f_{\mu}(\mathbf x_k)\|^2])  \nonumber \\
%    \leq &  2 ( \max_{t \in [T]} \mathbb {E}[\| \hat g_{t} -  \nabla f_{\mu}(\mathbf x_t)\|^2] + d \eta^2)
\end{align}
where the last inequality holds since $\sum_{k=1}^T \beta_2^{T-k} \leq 1/(1-\beta_2)$.%, and  $\mathbb {E}[\|    \nabla f_{\mu}(\mathbf x_k)\|^2] \leq d \mathbb {E}[\|    \nabla f_{\mu}(\mathbf x_k)\|_\infty^2] \leq d \eta^2 $ by \textbf{A2}.
%by our assumption $\sigma_{\hat g} \geq \mathbb {E}[\| \hat g_{k} -  \nabla f_{\mu}(\mathbf x_T)\|^2]$   and \textbf{A2}.
}
%\sijia{What is $\sigma_{\hat g}$?}

%is the upper bound of variance of $\hat g_k$. 

Uniformly randomly picking $R$ from $1$ to $T$ and substituting  \eqref{eq: v_t_bound} into \eqref{eq:final_con_2}  finishes the proof.
\hfill $\square$

\section{Proof for Convex Optimization}
\subsection{Proof of Proposition \ref{prop: regret_mid}}\label{app: prf_prop_regret_mid}
%\textbf{Proof of Proposition \ref{prop: regret_mid}}: 
We follow the analytic framework in \cite[Theorem 4]{reddi2018convergence} Based on Lemma\,\ref{lemma: smooth_f_random}, we obtain that $f_{t,\mu}$ defined in \eqref{eq: fmu_smooth} (with respect to $f_t$) is convex.
% and 
% $
% \hat {\mathbf g}_t 
% $ is an \textit{unbiased} estimate of $\nabla f_{t,\mu}$ \eqref{eq: smooth_est_grad}. 
The convexity of $f_{t,\mu}$ yields
\begin{align}\label{eq: convexity_fmu}
    f_{t,\mu} (\mathbf x_t) - f_{t,\mu} (\mathbf x^*) \leq   
    \langle \mathbb E_{\mathbf u} [\hat {\mathbf g}_t] , \mathbf x_t - \mathbf x^* \rangle, 
\end{align}
where we have used the fact that $\mathbb E_{\mathbf u} [\hat {\mathbf g}_t] = \nabla f_{t,\mu} (\mathbf x_t) $ given by Lemma\,\ref{lemma: smooth_f_random}. 
Taking the expectation with respect to all the randomness in \eqref{eq: convexity_fmu}, we then obtain
\begin{align}\label{eq: convexity_fmu_E}
    \mathbb E 
[     f_{t,\mu} (\mathbf x_t) - f_{t,\mu} (\mathbf x^*) ] \leq \mathbb E \langle \hat{\mathbf g}_t, \mathbf x_t - \mathbf x^* \rangle.
\end{align}

% Furthermore, 
% from \eqref{eq: dist_f_smooth_true_random} we have 
% $|f_{t,\mu}(\mathbf x) - f_t(\mathbf x)| \leq  %\mu L_c
% \mu^2 L_g/2
% $, and hence 
% \begin{align}\label{eq: dist_func_random_grad}
%     f_{t,\mu}(\mathbf x_t) - f_{t,\mu}(\mathbf x^*) - (  f_{t}(\mathbf x_t)  -  f_{t}(\mathbf x^*) ) \geq
%     %- 2 \mu L_c . 
%     -\mu^2 L_g.
% \end{align}

% Combining \eqref{eq: convexity_fmu_E} and \eqref{eq: dist_func_random_grad}, we obtain that
% \begin{align}
%  \mathbb E \left [ \sum_{t=1}^T \left [  f_{t}(\mathbf x_t)  -  f_{t}(\mathbf x^*) \right ]
%  \right ] 
%  \leq  %2  \mu L_c T 
%   \mu^2 L_g T
%  + \mathbb E \left [ \sum_{t=1}^T [f_{t,\mu} (\mathbf x_t) - f_{t,\mu} (\mathbf x^*) ] \right ].
% \end{align}

% which gives \eqref{eq: regret_mid}.
% %\hfill $\square$

%\subsection{Proof of Proposition \ref{prop: optgap2var_convex}}\label{app: prf_prop_optgap2var}
%\textbf{Proof of Proposition \ref{prop: optgap2var_convex}}:
Further, recall that $\Pi_{\mathcal X, \sqrt{\hat {\mathbf V}_t}}(\mathbf x^*) = \arg\min_{\mathbf x \in \mathcal X} \| \hat {\mathbf V}_t^{1/4} (\mathbf x - \mathbf x^*) \|^2 = \mathbf x^*$, where for ease of notation, let $\| \cdot \|$ denote the Euclidean norm.
Applying \citep[Lemma\,4]{reddi2018convergence} to ZO-AdaMM, we obtain that
\begin{align}
     \left \| \hat {\mathbf V}_t^{1/4} (\mathbf x_{t+1} - \mathbf x^*)  \right \|^2  
   & \leq \left \| \hat {\mathbf V}_t^{1/4} (\mathbf x_{t} - \alpha_t \hat{\mathbf V}_t^{-1/2} \mathbf m_t - \mathbf x^*)  \right \|^2 \nonumber \\
 & \hspace*{-1in} =  \left \| \hat {\mathbf V}_t^{1/4} (\mathbf x_{t} - \mathbf x^*)  \right \|^2  + \alpha_t^2 \| \hat{\mathbf V}_t^{-1/4} \mathbf m_t  \|^2    - 2 \alpha_t \langle \beta_{1,t} \mathbf m_{t-1} + (1-\beta_{1,t}) \hat {\mathbf g}_t,\mathbf x_t - \mathbf x^* \rangle.
\end{align}
Rearranging the above inequality, and using the Cauchy-Schwarz inequality 
$2 \langle \mathbf a, \mathbf b \rangle \leq c \| \mathbf a \|^2 + \cfrac{1}{c} \| \mathbf b \|^2$ for $c > 0$, we obtain
\begin{align}\label{eq: inner_prod_ghat}
     \langle \hat{\mathbf g}_t, \mathbf x_t - \mathbf x^* \rangle  \leq  
     & \frac{ \| \hat{\mathbf V}_t^{1/4} (\mathbf x_t - \mathbf x^*) \|^2 - \| \hat{\mathbf V}_t^{1/4} (\mathbf x_{t+1} - \mathbf x^*)\|^2 }{2 \alpha_t (1-\beta_{1,t})}
    %  \left  [
    % \| \hat{\mathbf V}_t^{1/4} (\mathbf x_t - \mathbf x^*) \|_2 - \| \hat{\mathbf V}_t^{1/4} (\mathbf x_{t+1} - \mathbf x^*)\|^2 
    % \right ] 
    + \frac{\alpha_t \| \hat{\mathbf V}_t^{-1/4} \mathbf m_t \|^2}{2(1-\beta_{1,t})}   \nonumber \\
    &  + 
    \frac{\beta_{1,t}}{1-\beta_{1,t}} \frac{\alpha_t  \| \hat{\mathbf V}_t^{-1/4} \mathbf m_{t-1} \|^2 }{ 2 }  +  \frac{\beta_{1,t}}{1-\beta_{1,t}} \frac{\| \hat{\mathbf V}_t^{1/4} (\mathbf x_t - \mathbf x^*) \|^2}{2 \alpha_t}.
    \end{align}
Taking the sum over $t$ for \eqref{eq: inner_prod_ghat}, we obtain 
\begin{align}\label{eq: inner_prod_ghat_sum}
    & \mathbb E \left [ \sum_{t=1}^T \langle \hat{\mathbf g}_t, \mathbf x_t - \mathbf x^* \rangle \right ]  
     \leq   \frac{1}{2(1-\beta_{1})}  \mathbb E  \underbrace{
     \left [  
     \sum_{t=1}^T        { \alpha_t \| \hat{\mathbf V}_t^{-1/4} \mathbf m_t \|^2}
     \right ]   }_{A}  +  \frac{\beta_1}{2(1-\beta_1)} \mathbb E  \underbrace{ \left [ \sum_{t=1}^T
  \alpha_t \| \hat{\mathbf V}_t^{-1/4} \mathbf m_{t-1} \|^2  
     \right ]   }_{B} \nonumber \\
         & + \sum_{t=1}^T \mathbb E \left [   \frac{  \| \hat{\mathbf V}_t^{1/4} (\mathbf x_t - \mathbf x^*) \|^2 - \| \hat{\mathbf V}_t^{1/4} (\mathbf x_{t+1} - \mathbf x^*)\|^2 }{2 \alpha_t (1-\beta_{1,t})}   \right ]  + \sum_{t=1}^T \mathbb E \left [ \frac{\beta_{1,t}}{2 \alpha_t (1-\beta_{1})}  \| \hat{\mathbf V}_t^{1/4} (\mathbf x_t - \mathbf x^*) \|^2 \right ],
    \end{align}
    where we have used the facts that $\beta_{1,t} \leq \beta_1$ and $1/(1-\beta_{1,t}) \leq 1/(1-\beta_1)$.
    
We next bound term $A$  in \eqref{eq: inner_prod_ghat_sum}.  
Based on \eqref{eq: mt}, we can directly apply \citep[Lemma\,2]{reddi2018convergence} to obtain that 
\begin{align}\label{eq: result_lemma2_v0}
  A
    \leq & 
    \frac{\alpha \sqrt{1+\log{T}}}{(1-\beta_1)(1-\gamma)\sqrt{1-\beta_2}} \sum_{i=1}^d \| \hat{\mathbf g}_{1:T,i} \|_2.  %\nonumber \\
    % \leq &    \frac{\alpha \sqrt{1+\log{T}}}{(1-\beta_1)(1-\gamma)\sqrt{1-\beta_2}} \sqrt{ d \sum_{t=1}^T \mathbb E \| \hat {\mathbf g}_t \|^2 },
\end{align}

%%%%%%%%%%%%%%%%%%%%%%%% Remove begin: %%%%%%%%%%%%%%%%%%%%%%%%%
\iffalse
In \eqref{eq: result_lemma2_v0}, we can further obtain that
\begin{align}\label{eq: gt_square}
   & \frac{1}{d} \sum_{i=1}^d \sqrt{\| \hat{\mathbf g}_{1:T,i} \|_2^2} \leq \sqrt{\frac{1}{d} \sum_{i=1}^d \| \hat{\mathbf g}_{1:T,i} \|^2 } \nonumber \\
    =  & \sqrt{\frac{1}{d} \sum_{i=1}^d \sum_{t=1}^T \| \hat{\mathbf g}_{t,i} \|^2 }  = \sqrt{\frac{1}{d} \sum_{t=1}^T \| \hat{\mathbf g}_{t} \|^2 },
\end{align}
where the first inequality holds due to $(\sum_{i=1}^d a_i)^2 \leq d \sum_{i=1}^d a_i^2 $.
Substituting \eqref{eq: gt_square} into \eqref{eq: result_lemma2_v0}, we have
\begin{align}\label{eq: result_lemma2}
  A
      \leq &    \frac{\alpha \sqrt{1+\log{T}}}{(1-\beta_1)(1-\gamma)\sqrt{1-\beta_2}} \sqrt{ d \sum_{t=1}^T \mathbb E \| \hat {\mathbf g}_t \|^2 }.
\end{align}
\fi
%%%%%%%%%%%%%%%%%%%%%%%% Remove end %%%%%%%%%%%%%%%%%%%%%%%%%

Furthermore, we bound term $B$ in \eqref{eq: inner_prod_ghat_sum}. 
%This is similar to \eqref{eq: result_lemma2} but with slight modifications.
Based on \eqref{eq: mt}, we obtain that
\begin{align}\label{eq: Vt_mt_1}
 B
     = &
    \sum_{t=1}^{T-1} \alpha_t \| \hat{\mathbf V}_t^{-1/4} \mathbf m_{t-1} \|^2 + 
    \alpha_T  \sum_{i=1}^d \frac{m_{T-1,i}^2}{ \sqrt{\hat{v}_{T,i}} } \nonumber \\
    \leq  &  \sum_{t=1}^{T-1} \alpha_t \| \hat{\mathbf V}_t^{-1/4} \mathbf m_{t-1} \|^2 + 
    \alpha_T  \sum_{i=1}^d \frac{m_{T-1,i}^2}{ \sqrt{{v}_{T,i}} },
%     \nonumber \\
%     \overset{\eqref{eq: ineq_correct}}{\leq } & \sum_{t=1}^{T-1} \alpha_t \| \hat{\mathbf V}_t^{-1/4} \mathbf m_{t-1} \|^2  + \alpha \sum_{i=1}^d 
%     \frac{ \left ( \sum_{j=1}^{T-1} 
%     \left ( 
%     \prod_{k=1}^{T-1-j} \beta_{1,T-k}
%     \right ) \right )
%     \left (
%     \sum_{j=1}^{T-1} 
%     \left ( 
%     \prod_{k=1}^{T-1-j} \beta_{1,T-k}
%     \right ) \hat g_{j,i}^2
%     \right )
%     }{ \sqrt{T((1-\beta_2) \sum_{j=1}^T \beta_2^{T-j} \hat g_{j,i}^2)} } \nonumber \\
%     \overset{\beta_{1,t}\leq \beta_1}{\leq} & \sum_{t=1}^{T-1} \alpha_t \| \hat{\mathbf V}_t^{-1/4} \mathbf m_t \|^2 + \frac{\alpha}{(1-\beta_1)\sqrt{T(1-\beta_2)}} \sum_{i=1}^d 
%     \frac{ 
%     \sum_{j=1}^{T-1} 
%   \beta_{1}^{T-1-j}
%   \hat g_{j,i}^2
%     }{ \sqrt{\sum_{j=1}^T \beta_2^{T-j} \hat g_{j,i}^2} } \nonumber \\
%     \leq & \sum_{t=1}^{T-1} \alpha_t \| \hat{\mathbf V}_t^{-1/4} \mathbf m_t \|^2 + \frac{\alpha}{(1-\beta_1)\sqrt{T(1-\beta_2)}} \sum_{i=1}^d  \sum_{j=1}^T 
%     \frac{ 
%   \beta_{1}^{T-1-j}
%   \hat g_{j,i}^2
%     }{ \sqrt{\beta_2^{T-j} \hat g_{j,i}^2} } \nonumber \\
%      = & \sum_{t=1}^{T-1} \alpha_t \| \hat{\mathbf V}_t^{-1/4} \mathbf m_t \|^2 + \frac{\alpha}{\textcolor{blue}{\beta_1}(1-\beta_1)\sqrt{T(1-\beta_2)}} \sum_{i=1}^d  \sum_{j=1}^T 
%   \gamma^{T-j} |\hat g_{j,i}|.
\end{align}
where we have used the fact that $\mathbf v_t \leq \hat{\mathbf v}_t$ given in Algorithm\,\ref{alg:zoadam}.
The last term in \eqref{eq: Vt_mt_1} can be further derived via \eqref{eq: mt},
\begin{align}\label{eq: ineq_correct}
     \alpha_T \sum_{i=1}^d \frac{m_{T-1,i}^2}{\sqrt{v_{T,i}}} 
    & =
    \alpha \sum_{i=1}^d 
    \frac{\left (\sum_{j=1}^{T-1} \left [ 
    \left ( 
    \prod_{k=1}^{T-j - 1} \beta_{1,T-k}
    \right ) \hat  g_{j,i} \textcolor{black}{(1-\beta_{1,j})}
    \right ] \right )^2}{ \sqrt{T(1-\beta_2) \sum_{j=1}^T ( \beta_2^{T-j} \hat g_{j,i}^2 )} } \nonumber \\
    & \leq  \alpha \sum_{i=1}^d 
    \frac{ \left ( \sum_{j=1}^{T-1}
   \beta_1^{T-1-j} (1-\beta_{1,j})^2 \right ) \left ( \sum_{j=1}^{T-1} \beta_1^{T-1-j} \hat g_{j,i}^2 \right )
    % \left (
    % \sum_{j=1}^T 
    % \left ( 
    % \prod_{k=1}^{T-j} \beta_{1,T-k+1}
    % \right ) g_{j,i}^2
    % \right )
    }{ \sqrt{T(1-\beta_2) \sum_{j=1}^T (\beta_2^{T-j} \hat  g_{j,i}^2)} } \nonumber \\
    & \leq \alpha \sum_{i=1}^d 
    \frac{ \left ( \sum_{j=1}^T 
   \beta_1^{T-1-j}
    \right )  \left ( \sum_{j=1}^{T-1} \beta_1^{T-1-j} \hat g_{j,i}^2 \right )
    % \left (
    % \sum_{j=1}^T 
    % \left ( 
    % \prod_{k=1}^{T-j} \beta_{1,T-k+1}
    % \right ) g_{j,i}^2
    % \right )
    }{ \sqrt{T(1-\beta_2) \sum_{j=1}^T ( \beta_2^{T-j} \hat  g_{j,i}^2)} } \nonumber \\
    & \leq 
    \frac{\alpha}{(1-\beta_1)\sqrt{T(1-\beta_2)}} \sum_{i=1}^d  \sum_{j=1}^T 
    \frac{ 
   \beta_{1}^{T-1-j}
  \hat g_{j,i}^2
    }{ \sqrt{\beta_2^{T-j} \hat g_{j,i}^2} } \nonumber \\
    & = \frac{\alpha}{\textcolor{black}{\beta_1}(1-\beta_1)\sqrt{T(1-\beta_2)}} \sum_{i=1}^d  \sum_{j=1}^T 
   \gamma^{T-j} |\hat g_{j,i}|,
\end{align}
where the first inequality holds due to Cauchy-Schwarz inequality and $\beta_{1,T-k} \leq \beta_1$ for $\forall k$, the second inequality holds due to $1- \beta_{1,j} \leq 1$, and the third inequality holds due to $\sum_{j=1}^T \beta_1^{T-1-j} \leq 1/(1-\beta_1)$ and $\beta_2^{T-j} \hat{g}_{j,i}^2 \leq  \sum_{j=1}^{T} \beta_2^{T-j} \hat g_{j,i}^2$.
Based on \eqref{eq: ineq_correct}, we then applies the proof of \citep[Lemma\,2]{reddi2018convergence}, which yields
%%%%%%%%%%%%%%%%%%%%%%%% Remove from the original proof Begin: %%%%%%%%%%%%%%%%%%%%%%%%%
\iffalse
\begin{align}\label{eq: result_lemma2_2}
   B \leq  &
    \frac{\alpha \sqrt{1+\log{T}}}{{\color{black}\beta_1}(1-\beta_1)(1-\gamma)\sqrt{1-\beta_2}} \sum_{i=1}^d \| \hat{\mathbf g}_{1:T,i} \|_2 \nonumber \\
    \leq & \frac{\alpha \sqrt{1+\log{T}}}{{\color{black}\beta_1}(1-\beta_1)(1-\gamma)\sqrt{1-\beta_2}} \sqrt{d \sum_{t=1}^T \| \hat{\mathbf g}_{t} \|_2}
\end{align}
\fi
%%%%%%%%%%%%%%%%%%%%%%%% Remove from the original proof End: %%%%%%%%%%%%%%%%%%%%%%%%%
\begin{align}\label{eq: result_lemma2_2}
   B \leq  &
    \frac{\alpha \sqrt{1+\log{T}}}{{\color{black}\beta_1}(1-\beta_1)(1-\gamma)\sqrt{1-\beta_2}} \sum_{i=1}^d \| \hat{\mathbf g}_{1:T,i} \|_2 
    % \nonumber \\
    % \leq & \frac{\alpha \sqrt{1+\log{T}}}{{\color{black}\beta_1}(1-\beta_1)(1-\gamma)\sqrt{1-\beta_2}} \sqrt{d \sum_{t=1}^T \| \hat{\mathbf g}_{t} \|_2}
\end{align}
\iffalse
Compared with \eqref{eq: result_lemma2}, an extra factor $1/\beta_1$ appears in \eqref{eq: result_lemma2_2}.
\fi

Substituting %\eqref{eq: result_lemma2} 
\eqref{eq: result_lemma2_v0}
and \eqref{eq: result_lemma2_2} into  \eqref{eq: inner_prod_ghat_sum}, we obtain that
\iffalse
\begin{align}\label{eq: opt_gap_simplify1}
  & \mathbb E \left [ \sum_{t=1}^T \langle \hat{\mathbf g}_t, \mathbf x_t - \mathbf x^* \rangle \right ]  
     \leq    \frac{\alpha \sqrt{1+\log{T}}  \sqrt{d} \sqrt{  \sum_{t=1}^T \mathbb E \| \hat {\mathbf g}_t \|^2 }  }{(1-\beta_1)^2(1-\gamma)\sqrt{1-\beta_2}}  \nonumber \\
         & + \mathbb E \underbrace{  \left [ \sum_{t=1}^T   \frac{  \| \hat{\mathbf V}_t^{1/4} (\mathbf x_t - \mathbf x^*)  \|^2 - \| \hat{\mathbf V}_t^{1/4} (\mathbf x_{t+1} - \mathbf x^*)\|^2 }{2 \alpha_t (1-\beta_{1,t})}   \right ] }_{C} \nonumber \\
    & +  \mathbb E   \underbrace{ \left [  \sum_{t=1}^T \frac{\beta_{1,t}  \| \hat{\mathbf V}_t^{1/4} (\mathbf x_t - \mathbf x^*) \|^2 }{2 \alpha_t (1-\beta_{1})} \right ] }_{D},
\end{align}
\fi
\begin{align}\label{eq: opt_gap_simplify1}
  & \mathbb E \left [ \sum_{t=1}^T \langle \hat{\mathbf g}_t, \mathbf x_t - \mathbf x^* \rangle \right ]  
     \leq    \frac{\alpha \sqrt{1+\log{T}}      \sum_{i=1}^d \mathbb E \| \hat {\mathbf g}_{1:T,i} \|  }{(1-\beta_1)^2(1-\gamma)\sqrt{1-\beta_2}}  \nonumber \\
         & + \mathbb E \underbrace{  \left [ \sum_{t=1}^T   \frac{  \| \hat{\mathbf V}_t^{1/4} (\mathbf x_t - \mathbf x^*)  \|^2 - \| \hat{\mathbf V}_t^{1/4} (\mathbf x_{t+1} - \mathbf x^*)\|^2 }{2 \alpha_t (1-\beta_{1,t})}   \right ] }_{C}  +  \mathbb E   \underbrace{ \left [  \sum_{t=1}^T \frac{\beta_{1,t}  \| \hat{\mathbf V}_t^{1/4} (\mathbf x_t - \mathbf x^*) \|^2 }{2 \alpha_t (1-\beta_{1})} \right ] }_{D}.
\end{align}
\iffalse
where we have used  Jensen's inequality $\phi(\mathbb E [X]) \geq \mathbb E [\phi (X)]$ for a concave function $\phi(x) = \sqrt{x}$,
\begin{align}
 \mathbb E \sqrt{  \sum_{t=1}^T \| \hat{\mathbf g}_{t} \|^2 }  \leq \sqrt{   \sum_{t=1}^T \mathbb E \| \hat {\mathbf g}_t \|^2 }. \nonumber 
\end{align}
\fi

%%%%%%%%%%%%%%%%%%%% Wrong proof in Reddi's paper
\iffalse
In \eqref{eq: opt_gap_simplify1}, the term $C$ can be bounded as 
\begin{align} \label{eq: C_convex}
    C = & \frac{  \| \hat{\mathbf V}_1^{1/4} (\mathbf x_1 - \mathbf x^*)  \|^2 }{2\alpha_1 (1-\beta_1)}
    +  
    \sum_{t=2}^T \left [ 
    \frac{\| \hat {\mathbf V}_t^{1/4} (\mathbf x_t - \mathbf x^*) \|^2}{ 2 \alpha_t (1-\beta_{1,t})}
    -  \frac{\| \hat {\mathbf V}_{t-1}^{1/4} (\mathbf x_t - \mathbf x^*) \|^2}{2\alpha_{t-1}(1-\beta_{1,t})}
    \right ]
    \nonumber\\
    \leq & \frac{D_{\infty}^2\sum_{i=1}^d \hat{v}_{1,i}^{1/2} }{2\alpha_1 (1-\beta_1)}
    + \frac{ D_\infty^2   \sum_{i=1}^d \sum_{t=2}^T \left [ \frac{\hat{v}_{t,i}^{1/2}}{\alpha_t} - \frac{\hat{v}_{t-1,i}^{1/2}}{\alpha_{t-1}}  \right ] }{2(1-\beta_1)} = \frac{D_\infty^2 \sum_{i=1}^d \hat{v}_{T,i}^{1/2} }{2 \alpha_T (1-\beta_1)},
\end{align}
where the first inequality holds due to $\| \mathbf x - \mathbf x^* \|^2 \leq D_\infty^2$ for $\mathbf x \in \mathcal X$, and $\hat{v}_t^{1/2}/\alpha_t \geq \hat{v}_{t-1}^{1/2}/\alpha_{t-1}$.
\fi 
%%%%%%%%%%%%%%%%%%%%%%%%%%%%%%%%%%%%%%%%%%%%%%%%

In \eqref{eq: opt_gap_simplify1}, the term $D$ yields
\begin{align}\label{eq: D_convex}
    D \leq \frac{\beta_1 D_\infty^2}{2(1-\beta_1)} \sum_{t=1}^T \sum_{i=1}^d \frac{ \hat{v}_{t,i}^{1/2} }{\alpha_t}.
\end{align}

\textcolor{black}{\textit{We remark that it was shown in \cite{phuong2019convergence} that
the proof in \cite{reddi2018convergence} to bound the term $C$ is problematic. Compared to \cite{phuong2019convergence}, we propose a simpler fix to bound $C$ when $0 < \beta_{1,t} \leq \beta_{1,t-1}  \leq 1$.} We rewrite $C$ in \eqref{eq: opt_gap_simplify1} as
} 
\textcolor{black}{
\begin{align}\label{eq: shift_seq}
    % & \sum_{t=1}^T \mathbb E \left [   \frac{  \| \hat{\mathbf V}_t^{1/4} (\mathbf x_t - \mathbf x^*) \|^2 - \| \hat{\mathbf V}_t^{1/4} (\mathbf x_{t+1} - \mathbf x^*)\|^2 }{2 \alpha_t (1-\beta_{1,t})}   \right ] \nonumber\\
    %  = & \sum_{t=1}^T \mathbb E \left [   \frac{  \| \hat{\mathbf V}_t^{1/4} (\mathbf x_t - \mathbf x^*) \|^2 - \| \hat{\mathbf V}_t^{1/4} (\mathbf x_{t+1} - \mathbf x^*)\|^2 }{2 \alpha_t (1-\beta_{1,t})}   \right ] \nonumber\\
   C  = & \frac{  \| \hat{\mathbf V}_1^{1/4} (\mathbf x_{1} - \mathbf x^*) \|^2 }{2 \alpha_1 (1-\beta_{1,1})} + \sum_{t=2}^T
   %\mathbb E \left [  
   \frac{  \| \hat{\mathbf V}_t^{1/4} (\mathbf x_{t} - \mathbf x^*) \|^2 }{2 \alpha_t (1-\beta_{1,t})} \nonumber \\
   & - \sum_{t=2}^T \frac{ \| \hat{\mathbf V}_{t-1}^{1/4} (\mathbf x_{t} - \mathbf x^*)\|^2}{2 \alpha_{t-1} (1-\beta_{1,t-1})}   %\right ] 
%   \nonumber\\
%     &
    - 
    %\left [
    \frac{  \| \hat{\mathbf V}_T^{1/4} (\mathbf x_{T+1} - \mathbf x^*) \|^2 }{2 \alpha_T (1-\beta_{1,T})}  %\right ]
    \nonumber  \\
    =& \sum_{t=2}^T \left [ \frac{  \| \hat{\mathbf V}_t^{1/4} (\mathbf x_{t} - \mathbf x^*) \|^2 }{2 \alpha_t (1-\beta_{1,t})} - \frac{ \| \hat{\mathbf V}_{t-1}^{1/4} (\mathbf x_{t} - \mathbf x^*)\|^2}{2 \alpha_{t-1} (1-\beta_{1,t-1})}   \right ] \nonumber \\
    & +  \frac{  \| \hat{\mathbf V}_1^{1/4} (\mathbf x_{1} - \mathbf x^*) \|^2 }{2 \alpha_1 (1-\beta_{1,1})} - \frac{  \| \hat{\mathbf V}_T^{1/4} (\mathbf x_{T+1} - \mathbf x^*) \|^2 }{2 \alpha_T (1-\beta_{1,T})}.
\end{align}
Further, the first term in RHS of \eqref{eq: shift_seq} can be bounded as
\begin{align}\label{eq: shift_seq_term1}
    &\sum_{t=2}^T % \mathbb E
    \left [   \frac{  \| \hat{\mathbf V}_t^{1/4} (\mathbf x_{t} - \mathbf x^*) \|^2 }{2 \alpha_t (1-\beta_{1,t})} -  \frac{ \| \hat{\mathbf V}_{t-1}^{1/4} (\mathbf x_{t} - \mathbf x^*)\|^2}{2 \alpha_{t-1} (1-\beta_{1,t-1})}   \right ] \nonumber \\
    = & \sum_{t=2}^T % \mathbb E 
    \left [   \frac{  \| \hat{\mathbf V}_t^{1/4} (\mathbf x_{t} - \mathbf x^*) \|^2 }{2 \alpha_t (1-\beta_{1,t})} -  \frac{ \| \hat{\mathbf V}_{t-1}^{1/4} (\mathbf x_{t} - \mathbf x^*)\|^2}{2 \alpha_{t-1} (1-\beta_{1,t})}   \right ] \nonumber \\
    & + \sum_{t=2}^T % \mathbb E 
    \left [ \left(\frac{1}{1-\beta_{1,t}} - \frac{1}{1-\beta_{1,t-1}}\right) \frac{ \| \hat{\mathbf V}_{t-1}^{1/4} (\mathbf x_{t} - \mathbf x^*)\|^2}{2 \alpha_{t-1}}  \right ]\nonumber \\
   \overset{(a)}{\leq }   & \frac{1}{2(1-\beta_1)} \sum_{t=2}^T \left [ 
    \sum_{i=1}^d \left ( \frac{ \hat v_{t,i}^{1/2} (x_{t,i} - x_i^*)^2}{\alpha_t} - \frac{ \hat v_{t-1,i}^{1/2} (x_{t,i} - x_i^*)^2}{\alpha_{t-1}}
    \right )
    \right ] \nonumber \\
    % & +  \sum_{t=2}^T % \mathbb E 
    % \left [ \left(\frac{1}{1-\beta_{1,t}} - \frac{1}{1-\beta_{1,t-1}}\right) \frac{ \| \hat{\mathbf V}_{t-1}^{1/4} (\mathbf x_{t} - \mathbf x^*)\|^2}{2 \alpha_{t-1}}  \right ]\nonumber \\
    \overset{(b)}{\leq} & \frac{ D_\infty^2   \sum_{i=1}^d \sum_{t=2}^T \left [ \frac{\hat{v}_{t,i}^{1/2}}{\alpha_t} - \frac{\hat{v}_{t-1,i}^{1/2}}{\alpha_{t-1}}  \right ] }{2(1-\beta_1)}
    %+ \frac{1}{1-\beta_1} \sum_{i=1}^d \frac{\hat{v}_{T-1,i}^{1/2}}{2\alpha_{T-1} }D_{\infty}^2 
    %\nonumber \\
    %\leq & 
    \leq 
    \frac{D_\infty^2 \sum_{i=1}^d \hat{v}_{T,i}^{1/2} }{2 \alpha_T (1-\beta_1)}
\end{align}
where the inequality (a) holds since $\beta_{1,t} \leq \beta_{1,t-1} \leq \beta_1$ and $1/(1-\beta_{1,t}) - 1/(1-\beta_{1,t-1}) \leq 0$, and the inequality (b) holds due to $\|x_t - x^*\|_{\infty} \leq D_{\infty}$  and $\frac{\hat{v}_{t,i}^{1/2}}{\alpha_t} - \frac{\hat{v}_{t-1,i}^{1/2}}{\alpha_{t-1}} \geq 0 $.
% the first inequality is due to $\|x_t - x^*\|_{\infty} \leq D_{\infty}$ and $\beta_{1,t} \leq \beta_{1,t-1} \leq \beta_1$ and $\frac{\hat{v}_{t,i}^{1/2}}{\alpha_t} - \frac{\hat{v}_{t-1,i}^{1/2}}{\alpha_{t-1}} \geq 0 $, the last inequality is due to $\frac{\hat{v}_{t,i}^{1/2}}{\alpha_t} - \frac{\hat{v}_{t-1,i}^{1/2}}{\alpha_{t-1}} \geq 0 $.
Substituting \eqref{eq: shift_seq_term1} into \eqref{eq: shift_seq}, we obtain that
\begin{align}
    C \leq    \frac{D_\infty^2 \sum_{i=1}^d \hat{v}_{T,i}^{1/2} }{2 \alpha_T (1-\beta_1)} + \frac{D_{\infty}^2 \sum_{i=1}^d {\hat{v}_{1,i}^{1/2}} }{2 \alpha_1 (1-\beta_1)} \leq \frac{D_\infty^2 \sum_{i=1}^d \hat{v}_{T,i}^{1/2} }{ \alpha_T (1-\beta_1)},
\end{align}
where the last inequality holds since $\hat{v}_{t+1,i}^{1/2} \geq \hat{v}_{t,i}^{1/2}$ and $ \alpha_1 \geq \alpha_T $.
}

\textit{We highlight that although the proof on bounding $C$ in \cite[Theorem\,4]{reddi2018convergence} is problematic, the conclusion of \cite[Theorem\,4]{reddi2018convergence} keeps correct. 
}

Substituting $C$ and $D$ into \eqref{eq: opt_gap_simplify1}, we obtain that
\iffalse
\begin{align}
    \label{eq: opt_gap_simplify2}
  & \mathbb E \left [ \sum_{t=1}^T \langle \hat{\mathbf g}_t, \mathbf x_t - \mathbf x^* \rangle \right ]  
     \leq    \frac{\alpha \sqrt{1+\log{T}}  \sqrt{d} \sqrt{  \sum_{t=1}^T \mathbb E \| \hat {\mathbf g}_t \|^2 }   }{(1-\beta_1)^2(1-\gamma)\sqrt{1-\beta_2}}  \nonumber \\
         & +   \frac{D_\infty^2 \sum_{i=1}^d \mathbb E  [\hat{v}_{t,i}^{1/2} ] }{2 \alpha_T (1-\beta_1)}   + \frac{ D_\infty^2}{2(1-\beta_1)} \sum_{t=1}^T \sum_{i=1}^d \frac{ \beta_{1,t} \mathbb E [ \hat{v}_{t,i}^{1/2} ] }{\alpha_t}.
\end{align}
\fi
\begin{align}
    \label{eq: opt_gap_simplify2}
  & \mathbb E \left [ \sum_{t=1}^T \langle \hat{\mathbf g}_t, \mathbf x_t - \mathbf x^* \rangle \right ]  
     \leq    \frac{\alpha \sqrt{1+\log{T}}      \sum_{i=1}^d \mathbb E \| \hat {\mathbf g}_{1:T,i} \|   }{(1-\beta_1)^2(1-\gamma)\sqrt{1-\beta_2}}  \nonumber \\
         & +   \frac{D_\infty^2 \sum_{i=1}^d \mathbb E  [\hat{v}_{T,i}^{1/2} ] }{\alpha_T (1-\beta_1)}   + \frac{ D_\infty^2}{2(1-\beta_1)} \sum_{t=1}^T \sum_{i=1}^d \frac{ \beta_{1,t} \mathbb E [ \hat{v}_{t,i}^{1/2} ] }{\alpha_t}.
\end{align}

% In \eqref{eq: opt_gap_simplify2}, we continue to analyze  $\mathbb E[ \hat{v}_{t,i}^{1/2} ]$.
% From\eqref{eq: vt}, we have $\mathbf v_t = (1-\beta_2)  \sum_{k=1}^t \beta_2^{t-k} \mathbf 1^T \hat{\mathbf g}_k^2$. Based on the definition of $\hat{\mathbf v}_t$ and  let $\hat {\mathbf v}_0 = \mathbf 0$, we then have 

% \iffalse
In \eqref{eq: opt_gap_simplify2},
%we continue to analyze the upper bound on $\mathbb E[ \hat{v}_{t,i}^{1/2} ]$. 
since $\sqrt{\cdot}$ is a concave function, the Jensen's inequality yields
\begin{align}\label{eq: bd_vt}
    \mathbb E[\sqrt{\hat {v}_{t,i}}] \leq 
    \sqrt{\mathbb E [ \hat {v}_{t,i} ]}.
\end{align}
Substituting \eqref{eq: bd_vt} into \eqref{eq: opt_gap_simplify2} and \eqref{eq: convexity_fmu_E}, we complete the proof. 
% Thus, we have
% \begin{align}\label{eq: bd_sum_vt}
%   &  \sum_{i=1}^d \mathbb E [\sqrt{\hat{v}_{t,i}}] \leq  \sum_{i=1}^d \sqrt{ \mathbb E [ \hat{v}_{t,i} ]}
%     \leq \sqrt{d \sum_{i=1}^d \mathbb E [\hat{v}_{t,i}]} \nonumber \\
%      = & \sqrt{d} \sqrt{\mathbb E[\mathbf 1^T \hat{\mathbf v}_t]} \overset{\eqref{eq: vt}}{=} 
%      \sqrt{d} \sqrt{\mathbb E \left [ (1-\beta_2)  \sum_{k=1}^t \beta_2^{t-k} \mathbf 1^T \hat{\mathbf g}_k^2 \right ]} \nonumber \\
%      = &  \sqrt{d} \sqrt{  (1-\beta_2)  \sum_{k=1}^t \beta_2^{t-k} \mathbb E [\| \hat{\mathbf g}_k \|_2^2 ]},
% \end{align}
% where the second inequality holds due to $( \sum_{i=1}^d  a_i )^2 \leq d \sum_{i=1}^d  a_i^2$.
% % \fi

\iffalse
Substituting \eqref{eq: bd_sum_vt} into \eqref{eq: opt_gap_simplify2} and based on $\alpha_t = \alpha/\sqrt{t}$,
we eventually reach \eqref{eq: inner_opt_gap_convex}.
\fi
\hfill $\square$
% \begin{align}
%     \label{eq: opt_gap_simplify3}
%   & \mathbb E \left [ \sum_{t=1}^T \langle \hat{\mathbf g}_t, \mathbf x_t - \mathbf x^* \rangle \right ]  
%      \leq    \frac{\alpha \sqrt{1+\log{T}}  \sum_{i=1}^d \mathbb E \| \hat{\mathbf g}_{1:T,i} \|_2   }{(1-\beta_1)^2(1-\gamma)\sqrt{1-\beta_2}}  \nonumber \\
%          & +   \frac{D_\infty^2 \sqrt{T} \sqrt{d} }{2 \alpha (1-\beta_1)} \sqrt{  (1-\beta_2)  \sum_{k=1}^t \beta_2^{t-k} \mathbb E [\| \hat{\mathbf g}_k \|_2^2 ]}  \nonumber \\
%          & + \frac{\beta_1 D_\infty^2 \sqrt{d}}{2(1-\beta_1)} \sum_{t=1}^T  \frac{   \sqrt{  (1-\beta_2)  \sum_{k=1}^t \beta_2^{t-k} \mathbb E [\| \hat{\mathbf g}_k \|_2^2 ]} }{\alpha_t} .
% \end{align}

\section{Supplementary Material of Experiments}\label{app: exp_para}

\subsection{Problem  and experiment setup}
It is  known that DNN-based image classifiers are vulnerable to adversarial examples---one can carefully craft images with imperceptible perturbations (a.k.a. adversarial perturbations or adversarial attacks) that can fool image classifiers even under a \textit{black box} threat model, where details of the model are unknown to the attacker \cite{chen2017zoo,ilyas2018blackbox,suya2017query,cheng2018query}.
%That is, the adversary has no knowledge on the configurations of the image classifier, and instead the only mode of interaction with this classifier is by function queries.

We focus on two problem settings of black-box adversarial attacks: per-image adversarial perturbation and universal adversarial perturbation.  
% The former has been studied  by the vast body of research 
% \cite{goodfellow2015explaining,carlini2017towards,chen2017zoo,ilyas2018blackbox,suya2017query,liu2018signsgd,cheng2018query}, where 
% each image is crafted individually to fool neural networks. 
% The latter aims to  find a single  perturbation  
% that  is applied
% to multiple images simultaneously in order to fool an image classifier 
% \cite{moosavi2017universal,liu2018_NIPS,perolat2018playing,athalye18}.
%We next formulate the black-box adversarial attack problems of our interest.
 Let ($\mathbf{x},t$) denote a legitimate image $\mathbf{x}$ with the true label $t \in \{1,2,\ldots,K\}$, where $K$ is the total number of image classes. And let $\mathbf x^\prime = \mathbf x+ \boldsymbol{\delta}$ denote an adversarial example, where $\boldsymbol{\delta}$ is the adversarial perturbation.
Our goal is to design $\boldsymbol{\delta}$ for a single image $\mathbf x$ or multiple images $\{ \mathbf x_i \}_{i=1}^M$.
 Spurred by \cite{carlini2017towards}, we consider the optimization problem
{\small \begin{align}\label{eq: attack_general}
\begin{array}{ll}
    \displaystyle \minimize_{\boldsymbol \delta } & \frac{\lambda}{M} \sum_{i=1}^Mf(\mathbf x_i + \boldsymbol \delta) 
    +  \| \boldsymbol{\delta} \|_2^2
    \\
  \st    &  (\mathbf x_i + \boldsymbol \delta) \in [-0.5,0.5]^d, \forall i, 
  %\\
 % & \| \boldsymbol{\delta} \|_2 \leq \epsilon,
\end{array}
\end{align}%
}where 
 $f(\mathbf x_0 + \boldsymbol \delta)$ denotes the (black-box) attack  loss function,  $\lambda >0$ is a regularization parameter that strikes a balance between minimizing the attack loss and the $\ell_2$  distortion, and we normalize the  pixel values  to $[-0.5,0.5]^d$. In problem \eqref{eq: attack_general}, we specify the loss function for untargeted attack \cite{carlini2017towards},
$
f(\mathbf x^\prime)=\max \{    Z(\mathbf x^\prime )_{t} - \max_{j \neq t}  Z(\mathbf x^\prime )_j, - \kappa \}
$, where $Z(\mathbf x^\prime)_k$ denotes the prediction score of class $k$ given the input $\mathbf x^\prime$, and the   parameter  $\kappa > 0$ governs the gap between the confidence of the  predicted   label and the true label $t$. In experiments, we choose $\kappa = 0$, and the attack loss $f$ reaches the minimum value $0$ as the perturbation succeeds to fool the neural network.

In problem \eqref{eq: attack_general}, if $M = 1$, then it becomes our first task to find per-image adversarial perturbations. If $M > 1$, then the problem corresponds to the task of finding universarial adversarial perturbations to $M$ images. 
Problem \eqref{eq: attack_general} yields a constrained   formulation for the design of black-box adversarial attacks. Since some   ZO algorithms are designed only for unconstrained optimization (see Table\,\ref{table: SZO_complexity_T}), we also consider the unconstrained version of  
problem  \eqref{eq: attack_general} \cite{liu2018_NIPS},
 {\small \begin{align}\label{eq: attack_general_uncons}
\hspace*{-0.15in} \begin{array}{ll}
    \displaystyle \minimize_{\mathbf w \in \mathbb R^d} &    % {\lambda} 
    \frac{\lambda}{M} \sum_{i=1}^M \left [  f\left (0.5\tanh(\tanh^{-1}(2 \mathbf x_i ) + \mathbf w)   \right ) \right .  \\
   &    \left. +  \|  0.5\tanh(\tanh^{-1}(2 \mathbf x_i ) + \mathbf w) - \mathbf x_i  \|_2^2 \right ], 
\end{array}
\end{align}%
}where  $\mathbf w \in \mathbb R^d$ are optimization variables, and we eliminate   the inequality constraint in  \eqref{eq: attack_general} by leveraging 
  $  0.5\tanh(\tanh^{-1}(2 \mathbf x_i) + \mathbf w) = \mathbf x_i + \boldsymbol \delta \in [-0.5,0.5]^d $.
 
The experiments of generating  black-box adversarial examples will be performed on Inception V3 \cite{szegedy2016rethinking}  under the dataset  %CIFAR-10 and 
ImageNet \cite{deng2009imagenet}. We will compare the proposed ZO-AdaMM method with $6$ existing ZO algorithms, ZO-SGD \cite{ghadimi2013stochastic}, ZO-SCD \cite{lian2016comprehensive} and ZO-signSGD \cite{liu2018signsgd} for unconstrained optimization, and ZO-PSGD \cite{ghadimi2016mini}, ZO-SMD \cite{duchi2015optimal} and ZO-NES \cite{ilyas2018blackbox} for constrained optimization.
The first $5$ methods have been summarized in Table\,\ref{table: SZO_complexity_T}, and
  ZO-NES refers to the black-box attack generation method 
 in \cite{ilyas2018blackbox}, which applies  a projected version of ZO-signSGD using
 natural evolution strategy (NES) based random gradient estimator. In all the aforementioned ZO algorithms, we  adopt the   random gradient estimator \eqref{eq: grad_rand_ave} and set $b = 1$ and $q = 10$ so that every method takes the same query cost per iteration. Accordingly, the total query complexity is consistent with the number of iterations. %\textit{We refer readers to Appendix\,\ref{app: exp_para} for more details on algorithmic parameter settings in our experiments.}

\begin{figure*}[htb]
\centerline{
\begin{tabular}{ccc}
     \includegraphics[width=.3\textwidth]{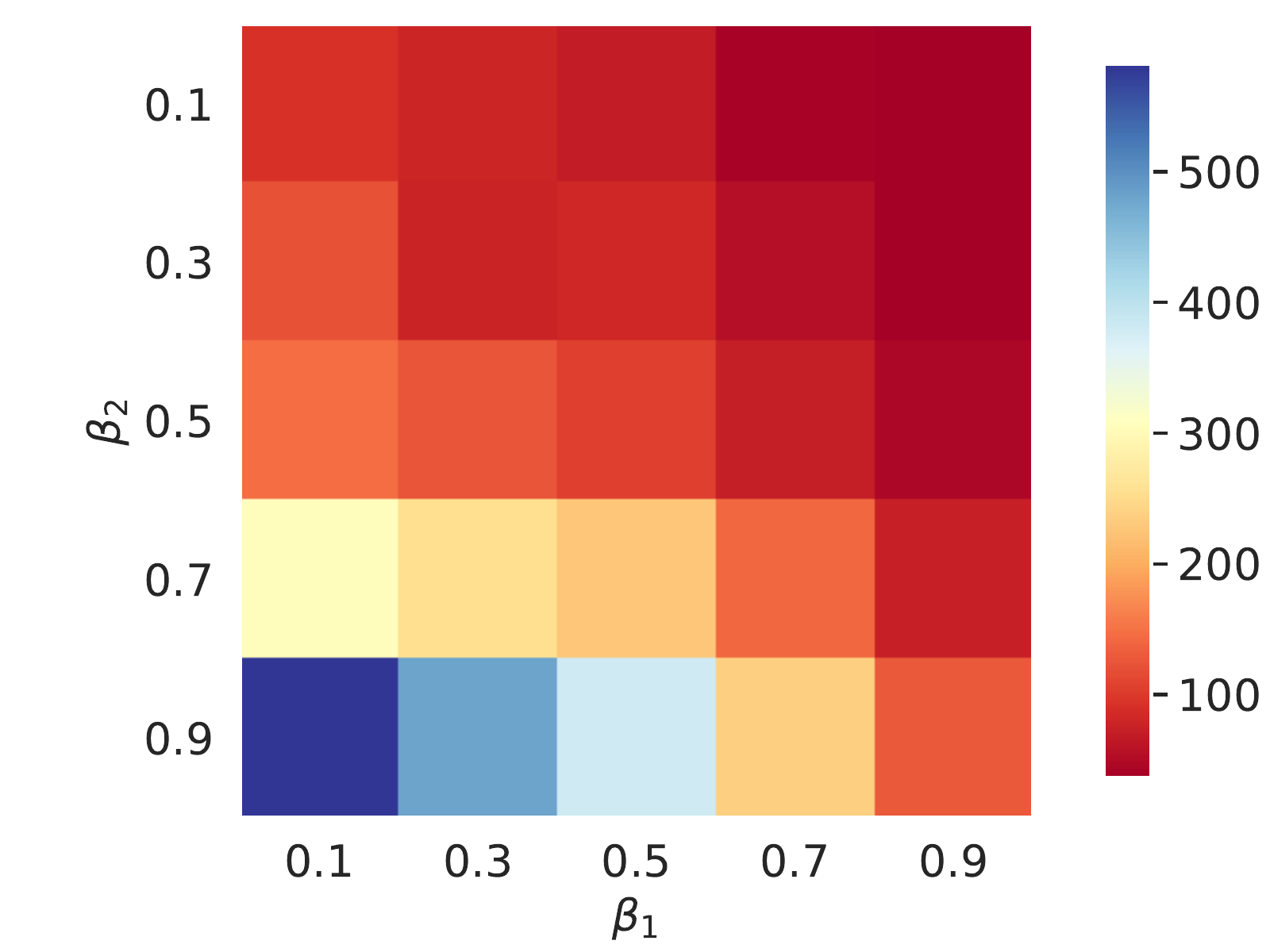} &  \hspace*{-0.15in}    \includegraphics[width=.3\textwidth]{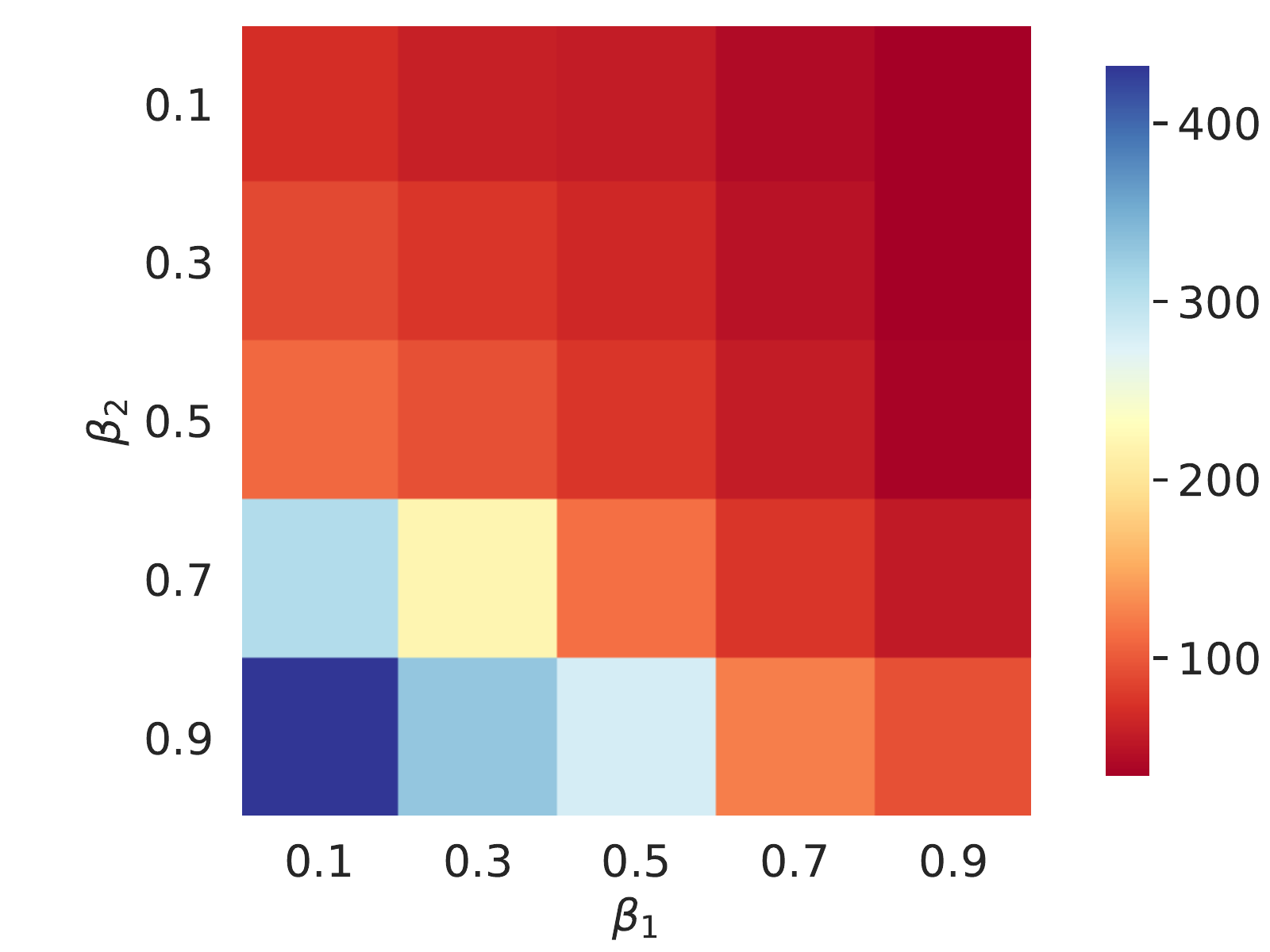}
     &  \hspace*{-0.15in}   
     \includegraphics[width=.3\textwidth]{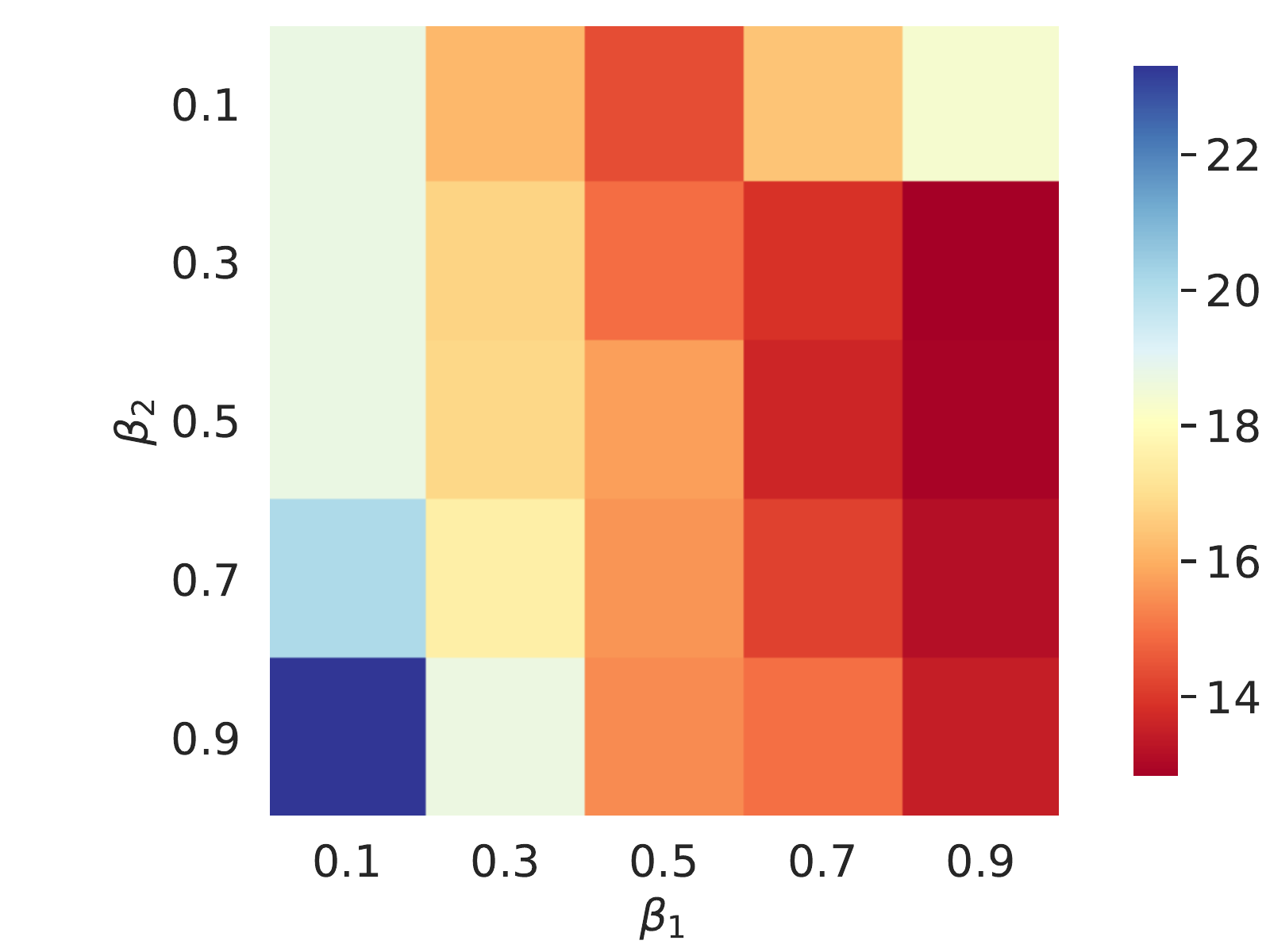}
     \\
     (a) &  \hspace*{-0.15in}   (b)  & \hspace*{-0.15in}  (c)
     \end{tabular}}
\caption{\footnotesize{The heat map of the converged objective value  at  $1000$ iterations versus different combinations of  $\beta_1$ and $\beta_2$ of ZO-AdaMM. (a) Unconstrained per-image (ID $11$) adversarial attack problem \eqref{eq: attack_general_uncons}; (b) Constrained per-image (ID $11$) adversarial attack problem \eqref{eq: attack_general}; (c)  Universal adversarial attack problem   \eqref{eq: attack_general} with $M = 10$.
}}
  \label{fig: beta_per_uncons_cons_uni}
\end{figure*}

In Fig.\,\ref{fig: beta_per_uncons_cons_uni}, we show the influence of
exponential averaging parameters $\beta_1$ and $\beta_2$ on the convergence of ZO-AdaMM, in terms of  the converged total loss  
while designing the per-image (ID $11$ in  ImageNet) and universal adversarial attack.  As we can see, the typical choice of $\beta_2 > 0.9$ is no longer the empirically optimal choice in the ZO setting. 
In all of our experiments, we find that  the choice of $\beta_1 \geq  0.9$ and $\beta_2 \in [0.3,0.5]$ performs well in practice. 
In Table\,\ref{table: alpha_per} and \ref{table: alpha_uni}, we present the best learning rate parameter $\alpha$ founded by greedy search at each experiment, in the sense that  the smallest objective function (corresponding to the successful attack) is achieved given the maximum number of iterations $T$. 

\begin{table}[htb]
\centering
\caption{Greedy search on the best learning rate parameter $\alpha$ for generating per-image adversarial perturbations.
}
\label{table: alpha_per}
\begin{adjustbox}{width=0.48\textwidth }
\begin{tabular}{c|c|c|c}
\toprule[1pt]
Methods  & Learning rate $\alpha$  & \begin{tabular}[c]{@{}c@{}}  Converged \\
objective value
\end{tabular} & \begin{tabular}[c]{@{}c@{}}  Success of \\
attack
\end{tabular} \\
\midrule[1pt]
\multirow{4}{*}{ZO-PSGD} & $4\times10^{-4}$     & 245.92  & $\times$ \\
& $2\times10^{-4}$   &78.66 & $\checkmark$\\
& \textcolor{blue}{$1\times10^{-4}$}   & \textcolor{blue}{31.42} & \textcolor{blue}{$\checkmark$} \\
& $9\times10^{-5}$   &30.98 &  $\times$ \\
\midrule[1pt]

\multirow{5}{*}{ZO-SMD} & $8\times10^{-4}$     & 245.92  & $\times$ \\
& $5\times10^{-4}$   &97.42 & $\checkmark$\\
& \textcolor{blue}{$3\times10^{-4}$}   & \textcolor{blue}{35.19} & \textcolor{blue}{$\checkmark$} \\
& $9\times10^{-5}$   & 36 &  $\times$ \\
\midrule[1pt]

\multirow{5}{*}{ZO-NES} & $5\times10^{-2}$     & 3997  & $\times$ \\
& $1\times10^{-2}$   &194.22 & $\checkmark$\\
& \textcolor{blue}{$9\times10^{-3}$}   & \textcolor{blue}{158.02} & \textcolor{blue}{$\checkmark$} \\
& $8\times10^{-3}$   &129.30 &  $\times$ \\
\midrule[1pt]

\multirow{5}{*}{ZO-SCD} & $8\times10^{-3}$     & 330.12  & $\times$ \\
& $2\times10^{-3}$   &77.14 & $\checkmark$\\
& \textcolor{blue}{$1\times10^{-3}$}   & \textcolor{blue}{42.87} & \textcolor{blue}{$\checkmark$} \\
& $9\times10^{-3}$   &39.60 &  $\times$ \\
\midrule[1pt]

\multirow{5}{*}{ZO-SGD} & $5\times10^{-3}$     & 1089.57  & $\times$ \\
& $8\times10^{-4}$   &33.60 & $\checkmark$\\
& \textcolor{blue}{$5\times10^{-4}$}   & \textcolor{blue}{31.11} & \textcolor{blue}{$\checkmark$} \\
& $4\times10^{-4}$   &33.13 &  $\times$ \\
\midrule[1pt]

\multirow{5}{*}{ZO-signSGD} & $8\times10^{-2}$     & 1590.02  & $\times$ \\
& $2\times10^{-2}$   &113.43 & $\checkmark$\\
& \textcolor{blue}{$1\times10^{-2}$}   & \textcolor{blue}{41.96} & \textcolor{blue}{$\checkmark$} \\
& $9\times10^{-3}$   &39.60 &  $\times$ \\
\midrule[1pt]

\end{tabular}
\end{adjustbox}
\end{table}

\begin{table}[htb]
\centering
\caption{Greedy search on the best learning rate parameter $\alpha$ for design of universal adversarial perturbations by solving problem \eqref{eq: attack_general}.
}
\label{table: alpha_uni}
\begin{adjustbox}{width=0.48\textwidth }
\begin{tabular}{c|c|c|c}
\toprule[1pt]
Methods  & Learning rate $\alpha$  & \begin{tabular}[c]{@{}c@{}}  Converged \\
objective value
\end{tabular} & \begin{tabular}[c]{@{}c@{}}  Success of \\
attack
\end{tabular} \\
\midrule[1pt]
\multirow{5}{*}{ZO-PSGD} & $1\times10^{-2}$     & 1072.05  & $\times$ \\
& $1\times10^{-3}$   &147.46 & $\checkmark$\\
& $4\times10^{-4}$   &56.99 & $\checkmark$\\
& \textcolor{blue}{$3\times10^{-4}$}   & \textcolor{blue}{36.86} & \textcolor{blue}{$\checkmark$} \\
& $2\times10^{-5}$   &24.91 &  $\times$ \\
\midrule[1pt]

\multirow{5}{*}{ZO-SMD} & $1\times10^{-2}$     & 788.46  & $\times$ \\
& $1\times10^{-3}$   &60.98 & $\checkmark$\\
& $6\times10^{-4}$   &36.86 & $\checkmark$\\
& \textcolor{blue}{$5\times10^{-4}$}   & \textcolor{blue}{29.56} & \textcolor{blue}{$\checkmark$} \\
& $4\times10^{-4}$   &24.91 &  $\times$ \\
\midrule[1pt]

\multirow{5}{*}{ZO-NES} & $1\times10^{-2}$     & 1230.15  & $\times$ \\
& $4\times10^{-2}$   &107.74 & $\checkmark$\\
& $7\times10^{-3}$   &65.64 & $\checkmark$\\
& \textcolor{blue}{$6\times10^{-3}$}   & \textcolor{blue}{54.00} & \textcolor{blue}{$\checkmark$} \\
& $5\times10^{-3}$   &42.57 &  $\times$ \\
\midrule[1pt]

\end{tabular}
\end{adjustbox}
\end{table}

\subsection{Per-image black-box adversarial attack}\label{app: exp1_para}
We  consider the task of per-image adversarial perturbation by solving
 problems \eqref{eq: attack_general} and 
\eqref{eq: attack_general_uncons}, where     $M = 1$ and $\lambda = 10$.
%\textcolor{Sijia_color}{[SL: ImageNet setting?]}.
In ZO-AdaMM (Algorithm\,1),  we set $\mathbf v_0 = \hat{\mathbf v}_0 = 10^{-5}$, $\mathbf m_0 = \mathbf 0$, $\beta_{1t} = \beta_1 = 0.9$, $\beta_2 = 0.3$ 
and $T = 1000$.  Here the exponential moving average parameters $(\beta_1, \beta_2)$ are exhaustively searched over $\{ 01, 0.3, 0.5, 0.7, 0.9\}^2$; see \textcolor{black}{Fig.\,\ref{fig: beta_per_uncons_cons_uni}-(a) \& (b)  in Appendix\,\ref{app: exp_para} as an example}. In ZO-AdaMM,
we also choose a decaying learning rate $\alpha_t = \alpha/\sqrt{t}$ with $\alpha = 0.01$. For fair comparison, we use the decaying strategy  for all other  ZO algorithms, and we determine  the best choice of  $\alpha $ by greedy search over the interval $[10^{-4}, 10^{-2}]$; see {Table\,\ref{table: alpha_per} in Appendix\,\ref{app: exp_para}} for more results on selecting $\alpha$.

In Table \ref{table: per_image_distortion}, we summarize the key statistics of each ZO optimization method for solving the per-image adversarial attack problem over $100$   images randomly selected from ImageNet.
For solving the unconstrained problem  \eqref{eq: attack_general_uncons},  ZO-SCD has the worst attack performance in general, i.e., leading to the largest number of iterations to reach the first successful attack and  the largest final distortion. 
%Our results are consistent with   \citep{chen2017zoo}, which uses the coordinate-wise gradient estimate as the descent direction to generate black-box adversarial examples. 
We also observe that  ZO-signSGD and ZO-AdaMM  achieve   better attack performance. However, the downside of ZO-signSGD is its poor convergence accuracy, given by the increase in distortion from the first successful attack to the final attack (i.e.,  $23.00 \to 28.52$ in Table\,\ref{table: per_image_distortion}).  For solving the constrained problem \eqref{eq: attack_general}, ZO-AdaMM achieves the best attack performance except for a slight drop in the attack success rate (ASR). Similar to ZO-signSGD, ZO-NES has a   poor convergence accuracy in terms of the increase in $\ell_2$ distortion after the attack becomes successful.

\begin{table}[htb]
\centering
\caption{\footnotesize{Performance of per-image attack  over $100$ images under $T = 1000$ iterations, where ASR represents attack success rate, and the  distortion $\|\boldsymbol{\delta} \|_2^2$ is averaged over successful attacks only.
 }}
\label{table: per_image_distortion}
\begin{adjustbox}{width= 0.7\textwidth }
\begin{threeparttable}
\begin{tabular}{c|c|c|c|c|c}
\toprule[1pt]
 \begin{tabular}[c]{@{}c@{}} Problem  \end{tabular}  & {Methods} &  {ASR} &  { \begin{tabular}[c]{@{}c@{}} Ave. iters \\ ($1$st succ.) \end{tabular} }   & {\begin{tabular}[c]{@{}c@{}} $\|\boldsymbol{\delta}_t \|_2^2$    \\ ($1$st succ.) \end{tabular}}& 
{\begin{tabular}[c]{@{}c@{}} Final\\ $\|\boldsymbol{\delta}_T \|_2^2$  \end{tabular}} \\
%\multicolumn{3}{c}{ Final distortion} \\ \cline{6-8}  
        %          &                    &     &           & & best  & average  & worst \\

                        \midrule[1pt] 
\multicolumn{1}{c|}{\multirow{4}{*}{  \begin{tabular}[c]{@{}c@{}}  \eqref{eq: attack_general_uncons} \end{tabular}  }}
 &ZO-SCD & $78\%$      & 240  &  %45.15 ,
 57.88 
 % & 4.81 ,6.17 
 & %44.86   ,
 57.51   % & 139.16     
 \\

\multicolumn{1}{c|}{} &ZO-SGD  & $78\%$   & \textbf{159}  & 
%29.92 ,
38.36 
%& 1.13   
& 
%29.52, 
37.85
%& 138.06   
\\
                        
\multicolumn{1}{c|}{} &ZO-signSGD  & $74\%$  & 179   & 
%19.21    , 
\textbf{23.00}
%&5.00
& %20.87 ,
28.52
%& 36.90 
\\
\multicolumn{1}{c|}{} &ZO-AdaMM  & {$\mathbf {81\%}$}    & 173   & %23.15  ,
28.58 
% & 1.56
& %23.10 ,
\textbf{28.20}
%& 49.90
\\

\midrule[1pt] 
                        
\multicolumn{1}{c|}{} &ZO-NES  & {$\mathbf {82\%} $}   & 229     & %67.88    ,
82.78
%& 12.80  
& %69.22, 
84.41
%& 185.26
\\

\multicolumn{1}{c|}{\multirow{4}{*}{ \begin{tabular}[c]{@{}c@{}}   \eqref{eq: attack_general} \end{tabular} 
}}
 &ZO-PSGD & $78\%$      & \textbf{112}   & %47.05  ,
 60.32
 %& 2.25 
 & %45.32, 
 58.10 % & 137.08      
 \\

\multicolumn{1}{c|}{} &ZO-SMD  & $76\%$    & 198    & 
%26.66    ,
35.08
%& 0.99 
& %26.64 ,
35.05
%& 78.42
\\

\multicolumn{1}{c|}{} &ZO-AdaMM  & $78\%$  & 197    & 
%18.54, 
\textbf{23.77}     
%& 0.93
& %18.50, 
\textbf{23.72}
%& 29.56
\\
\bottomrule[1pt]
\end{tabular}
\end{threeparttable}
\end{adjustbox}
\end{table}

\subsection{Universal black-box adversarial attack}\label{app: exp2_para}
In this experiment, we solve the constrained problem
\eqref{eq: attack_general}  for designing a universal adversarial perturbation $\boldsymbol{\delta}$, where we  
attack $M = 10$ images with  the true class label `brambling' and we set $\lambda = 10$ in \eqref{eq: attack_general}.
% In \eqref{eq: attack_general},
% we set \textcolor{Sijia_color}{[$\lambda = 10$]}
% and choose $M = 10$ images from ImageNet with the  class label `brambling', $T=20,000$.
%\textcolor{Sijia_color}{[Please elaborate on ZO-AdaMM parameter setting? $T$? and others?]}
% The other parameter settings are 
% similar to  Sec.\,\ref{sec: per_image_perturbation} except a) 
The setting of algorithmic parameters is similar to   Appendix\,\ref{app: exp1_para} except  $T = 20000$. For ZO-AdaMM, we choose 
$\alpha = 0.002$, $\beta_1 = 0.9$, and $\beta_2 = 0.3$, where 
the sensitivity  of exponential moving average parameters $(\beta_1, \beta_2)$ is shown in  Fig.\,\ref{fig: beta_per_uncons_cons_uni}-(c).
 For the other ZO algorithms, we greedily search $\alpha$
 over $[10^{-2}, 10^{-4}]$ and choose the value that achieves the best convergence accuracy as shown  in Table\,\ref{table: alpha_uni}.

% \section{Additional Experiments}
% \label{app: exp_para}

\begin{figure*}[htb]
\small
 \centering
\hspace*{-0.6in}\begin{tabular}{p{0.02in}p{0.60in}p{0.60in}p{0.60in}p{0.60in}p{0.65in}p{0.65in}p{0.65in}p{0.65in}}
 \parbox{0.5in}{\hspace*{-0.05in}\textbf{Iter = }}
 & 
 %\multicolumn{1}{c}{\hspace*{0.0in}$\mathrm{iter}=1000$} 
 \parbox{0.7in}{\centering ~ $ 1000$}  %will change to 100!
 & \parbox{0.7in}{\centering ~ $ 5000$}
 %\multicolumn{1}{c}{\hspace*{0.0in}$\mathrm{iter}=7000$}
 & %\multicolumn{1}{c}{$\mathrm{iter} = 13000$}
 \parbox{0.7in}{\centering ~ $ 10000$}
 &  %\multicolumn{1}{c}{$\mathrm{iter} = 20000$} 
 \parbox{0.7in}{\centering ~ $ 20000$}
 & %\multicolumn{4}{c}{adversarial examples when $T = 20,000$ which classified as: } 
 \multicolumn{4}{c}{\hspace*{0.11in} \textbf{\textcolor{blue}{Adv. examples} generated by universal perturbation}} 
\\
\rotatebox{90}{\parbox{0.8in}{\centering \textbf{ZO-PSGD}}}
& \includegraphics[width=.12\textwidth]{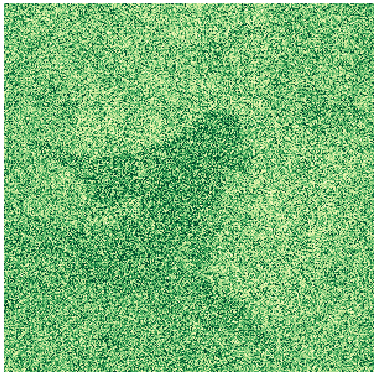}  
& \includegraphics[width=.12\textwidth]{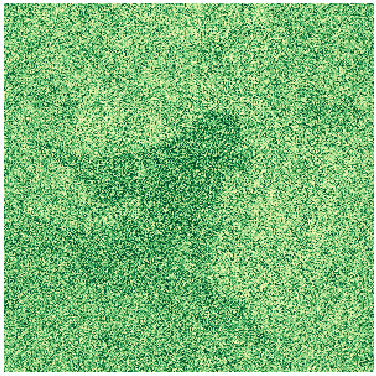}
& \includegraphics[width=.12\textwidth]{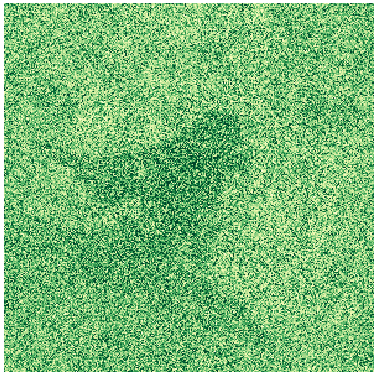}
& \includegraphics[width=.12\textwidth]{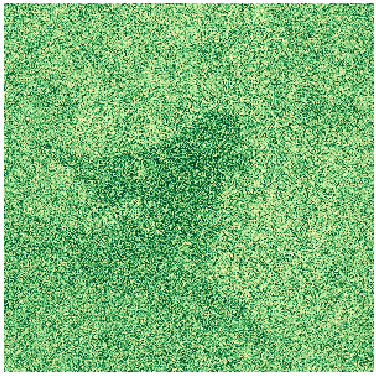}
& \hspace*{0.11in} \includegraphics[width=.12\textwidth,height=!]{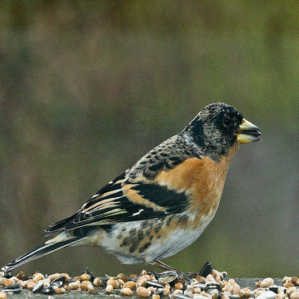}
& \hspace*{0.11in} \includegraphics[width=.12\textwidth,height=!]{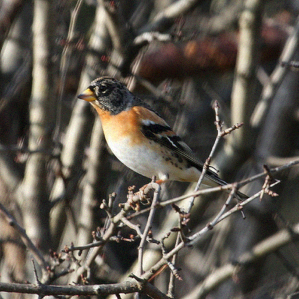}
& \hspace*{0.11in} \includegraphics[width=.12\textwidth,height=!]{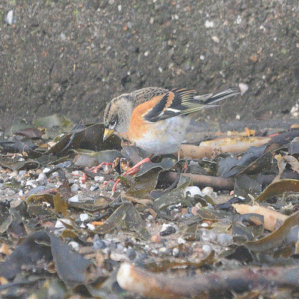}
& \hspace*{0.11in} \includegraphics[width=.12\textwidth,height=!]{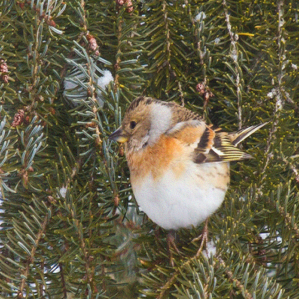} 
\vspace{-0.15in}
\\
 \parbox{0.5in}{\hspace*{-0.05in}\textbf{$\| \boldsymbol{\delta}\|_\infty $=}}
& %\multicolumn{1}{c}{0.118}
\parbox{0.7in}{\centering ~ $0.118$}
& %\multicolumn{1}{c}{0.133} 
\parbox{0.7in}{\centering ~ $0.131$}
& %\multicolumn{1}{c}{0.138}
\parbox{0.7in}{\centering ~ $0.138$}
& %\multicolumn{1}{c}{0.137} 
\parbox{0.7in}{\centering ~ $0.137$}
& %\multicolumn{1}{c}{goldfinch}
\parbox{1in}{\centering \textcolor{blue}{goldfinch}}
& % \multicolumn{1}{c}{robin} 
\parbox{1in}{\centering \textcolor{blue}{robin}}
&  % \multicolumn{1}{c}{linnet} 
\parbox{1in}{\centering \textcolor{blue}{linnet}}
&  %\multicolumn{1}{c}{indigo bunting}
\parbox{1in}{\centering \hspace*{0.03in} \textcolor{blue}{indigo bunting}}
\\
\rotatebox{90}{\parbox{0.8in}{\centering \textbf{ZO-SMD}}}   
& \includegraphics[width=.12\textwidth]{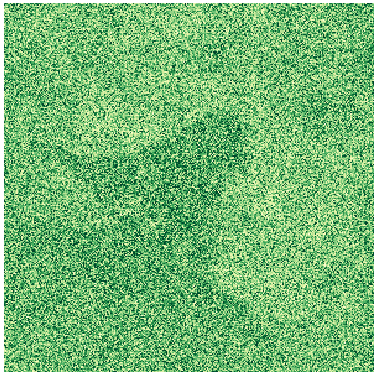}
& \includegraphics[width=.12\textwidth]{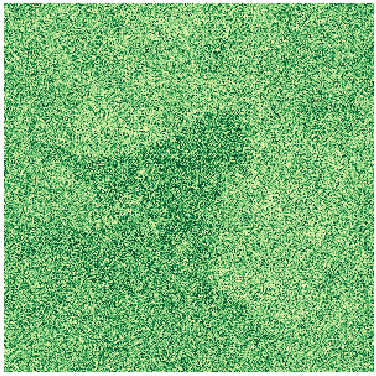}
& \includegraphics[width=.12\textwidth]{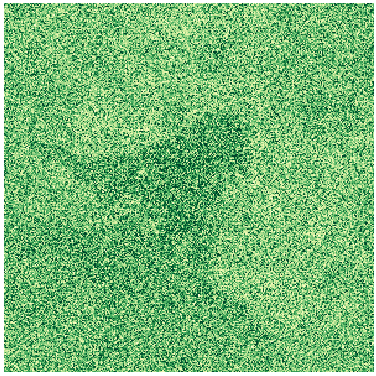}
& \includegraphics[width=.12\textwidth]{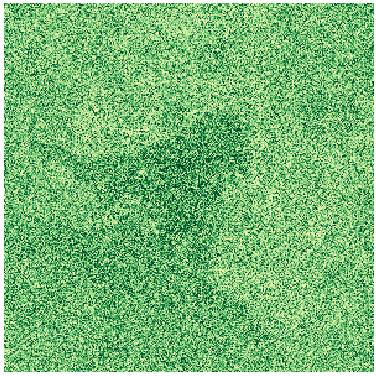}
& \hspace*{0.11in} \includegraphics[width=.12\textwidth,height=!]{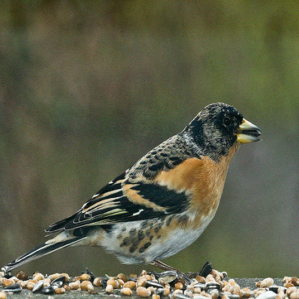}
& \hspace*{0.11in} \includegraphics[width=.12\textwidth,height=!]{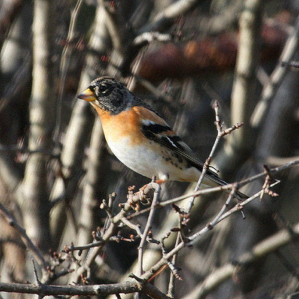}
& \hspace*{0.11in} \includegraphics[width=.12\textwidth,height=!]{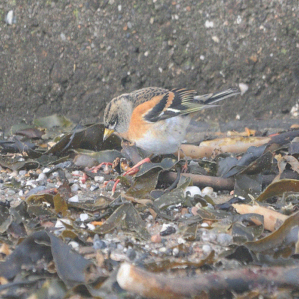}
& \hspace*{0.11in} \includegraphics[width=.12\textwidth,height=!]{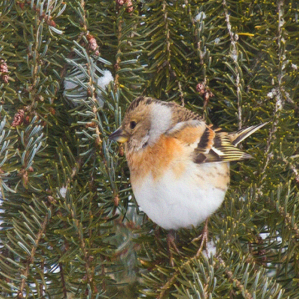}
\vspace{-0.15in}
\\
 \parbox{0.5in}{\hspace*{-0.05in}\textbf{$\| \boldsymbol{\delta}\|_\infty $=}}
& %\multicolumn{1}{c}{0.096} 
\parbox{0.7in}{\centering ~ $0.096$}
& %\multicolumn{1}{c}{0.096}
\parbox{0.7in}{\centering ~ $0.096$}
& %\multicolumn{1}{c}{0.099} 
\parbox{0.7in}{\centering ~ $0.099$}
& %\multicolumn{1}{c}{0.099} 
\parbox{0.7in}{\centering ~ $0.100$}
& %\multicolumn{1}{c}{goldfinch}
\parbox{1in}{\centering \textcolor{blue}{goldfinch}}
&  %\multicolumn{1}{c}{robin}
\parbox{1in}{\centering \textcolor{blue}{robin}}
&  %\multicolumn{1}{c}{ruddy turnstone} 
\parbox{1in}{\centering \hspace*{0.05in}
\textcolor{blue}{ruddy turnstone}}
&  %\multicolumn{1}{c}{indigo bunting}
\parbox{1in}{\centering \hspace*{0.03in} \textcolor{blue}{indigo bunting}}
\\
\rotatebox{90}{\parbox{0.8in}{\centering \textbf{ZO-NES}}}   
& \includegraphics[width=.12\textwidth]{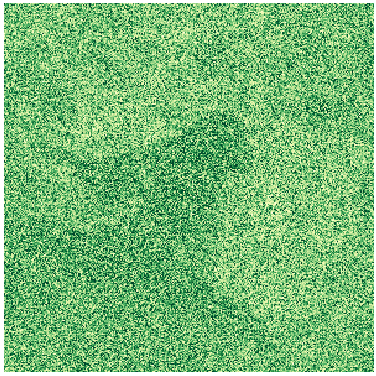}
& \includegraphics[width=.12\textwidth]{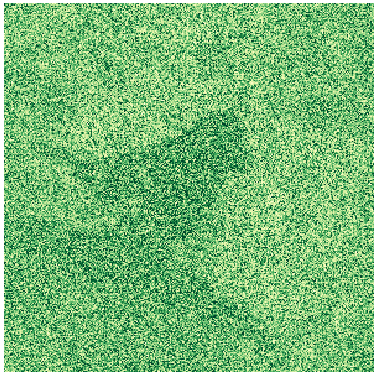}
& \includegraphics[width=.12\textwidth]{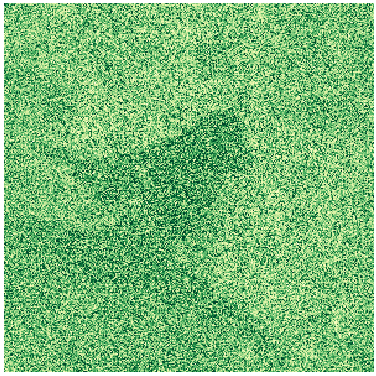}
& \includegraphics[width=.12\textwidth]{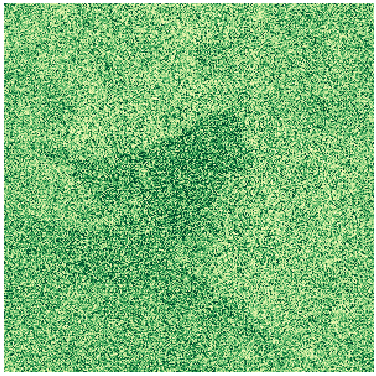}
& \hspace*{0.11in} \includegraphics[width=.12\textwidth,height=!]{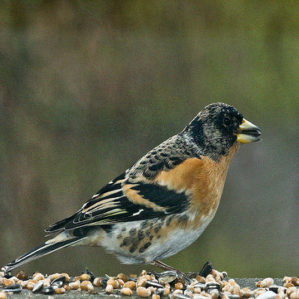}
& \hspace*{0.11in} \includegraphics[width=.12\textwidth,height=!]{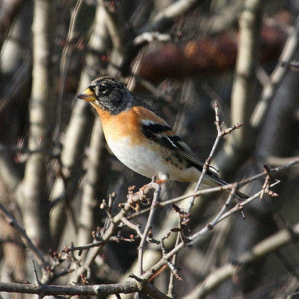}
& \hspace*{0.11in} \includegraphics[width=.12\textwidth,height=!]{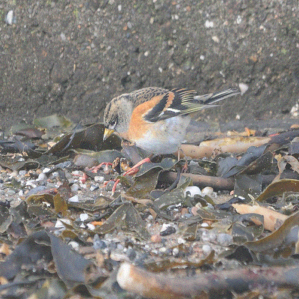}
& \hspace*{0.11in} \includegraphics[width=.12\textwidth,height=!]{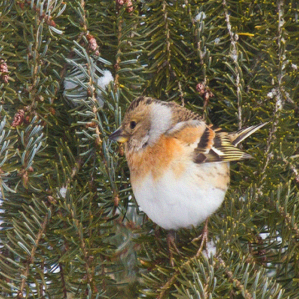}
\\
 \parbox{0.5in}{\hspace*{-0.05in}\textbf{$\| \boldsymbol{\delta}\|_\infty $=}}
& %\multicolumn{1}{c}{0.103} 
\parbox{0.7in}{\centering ~ $0.103$}
& %\multicolumn{1}{c}{0.116} 
\parbox{0.7in}{\centering ~ $0.107$}
& %\multicolumn{1}{c}{0.124} 
\parbox{0.7in}{\centering ~ $0.121$}
& %\multicolumn{1}{c}{0.124}
\parbox{0.7in}{\centering ~ $0.125$}
& \parbox{1in}{\centering \textcolor{blue}{goldfinch}}
&  %\multicolumn{1}{c}{robin}
\parbox{1in}{\centering \textcolor{blue}{robin}}
&  %\multicolumn{1}{c}{ruddy turnstone} 
\parbox{1in}{\centering \hspace*{0.05in}
\textcolor{blue}{ruddy turnstone}}
&  %\multicolumn{1}{c}{indigo bunting}
\parbox{1in}{\centering \hspace*{0.03in} \textcolor{blue}{indigo bunting}}
\\
\rotatebox{90}{\parbox{0.8in}{\centering \textbf{ZO-AdaMM}}}   
& \includegraphics[width=.12\textwidth]{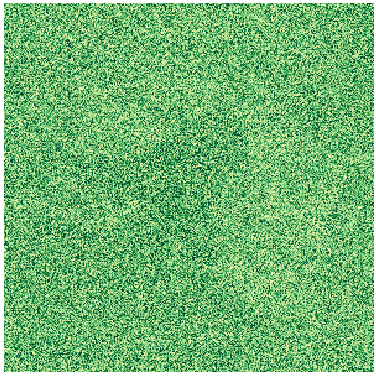}
& \includegraphics[width=.12\textwidth]{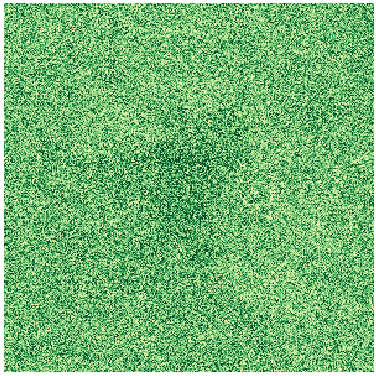}
& \includegraphics[width=.12\textwidth]{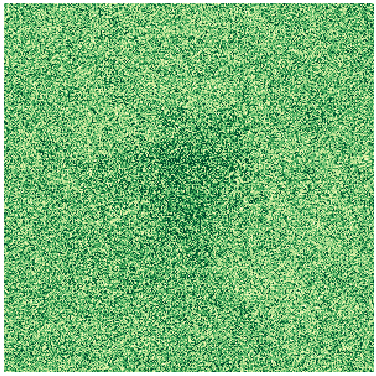}
& \includegraphics[width=.12\textwidth]{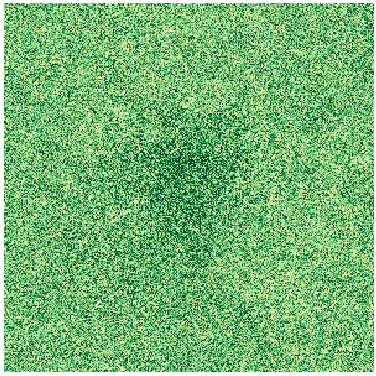}
& \hspace*{0.11in} \includegraphics[width=.12\textwidth,height=!]{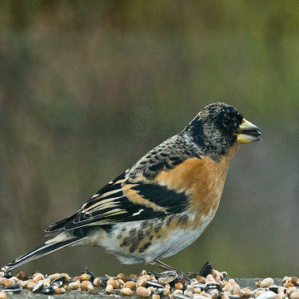}
& \hspace*{0.11in} \includegraphics[width=.12\textwidth,height=!]{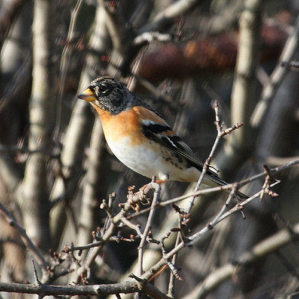}
& \hspace*{0.11in} \includegraphics[width=.12\textwidth,height=!]{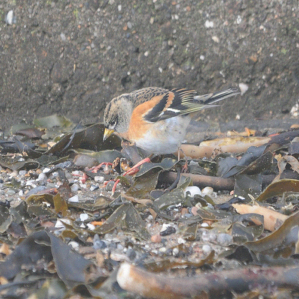}
& \hspace*{0.11in} \includegraphics[width=.12\textwidth,height=!]{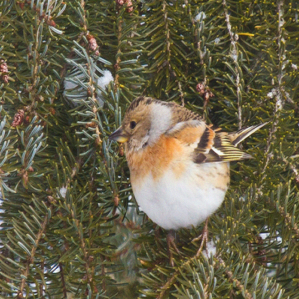}
\\
 \parbox{0.5in}{\hspace*{-0.05in}\textbf{$\| \boldsymbol{\delta}\|_\infty $=}}
& %\multicolumn{1}{c}{0.055}
\parbox{0.7in}{\centering ~ $0.055$}
& %\multicolumn{1}{c}{0.077} 
\parbox{0.7in}{\centering ~ $0.071$}
& %\multicolumn{1}{c}{0.084}
\parbox{0.7in}{\centering ~ $0.081$}
& %\multicolumn{1}{c}{0.079} 
\parbox{0.7in}{\centering ~ $0.079$}
& \parbox{1in}{\centering \textcolor{blue}{goldfinch}}
&  %\multicolumn{1}{c}{robin}
\parbox{1in}{\centering \textcolor{blue}{robin}}
&  %\multicolumn{1}{c}{ruddy turnstone} 
\parbox{1in}{\centering \hspace*{0.05in}
\textcolor{blue}{ruddy turnstone}}
&  %\multicolumn{1}{c}{indigo bunting}
\parbox{1in}{\centering \hspace*{0.03in} \textcolor{blue}{indigo bunting}}
\\

%\bottomrule[1pt]
\end{tabular}
\caption{\footnotesize{Visualization of universal perturbation versus different iterations and the eventually generated adversarial examples. 
Left four columns present universal perturbations found by different ZO algorithms at the iteration number $1000$, $5000$, $10000$ and $20000$, where the depth of the color corresponds to the strength of the perturbation, and the maximum distortion $\| \boldsymbol{\delta}\|_\infty $ (with deepest green) is given at the bottom of each subplot. 
The right four columns are $4$ of $10$ adversarial examples that lead to missclassfication from the original label `brambling' to an incorrectly predicted label given at the bottom of each subplot. }}
  \label{table: universal_pattern}
\end{figure*}

\textcolor{black}{In Fig.\,\ref{table: universal_pattern}, we visualize the pattern of universal adversarial perturbation obtained from different  methods. As we can see, the resulting universal perturbation pattern identifies the most discriminative  image  regions corresponding to the true label `brambling'. We also observe that although
each  method successfully generates the black-box adversarial example, ZO-AdaMM  yields the strongest attack that requires the least distortion strength. 
}

% \newpage
% \appendix
% \section{Do \emph{not} have an appendix here}

% \textbf{\emph{Do not put content after the references.}}
% %
% Put anything that you might normally include after the references in a separate
% supplementary file.

% We recommend that you build supplementary material in a separate document.
% If you must create one PDF and cut it up, please be careful to use a tool that
% doesn't alter the margins, and that doesn't aggressively rewrite the PDF file.
% pdftk usually works fine. 

% \textbf{Please do not use Apple's preview to cut off supplementary material.} In
% previous years it has altered margins, and created headaches at the camera-ready
% stage. 
%%%%%%%%%%%%%%%%%%%%%%%%%%%%%%%%%%%%%%%%%%%%%%%%%%%%%%%%%%%%%%%%%%%%%%%%%%%%%%%
%%%%%%%%%%%%%%%%%%%%%%%%%%%%%%%%%%%%%%%%%%%%%%%%%%%%%%%%%%%%%%%%%%%%%%%%%%%%%%%

\end{document}